\documentclass[final,3p,times]{elsarticle}

\usepackage[T1]{fontenc}

\usepackage{geometry}
\usepackage{graphicx} 
 
\usepackage{soul}
\usepackage{cancel}

\usepackage{commath}

\usepackage{enumitem}

\usepackage{amsthm}
\usepackage{amssymb}
\usepackage{amsmath}
\allowdisplaybreaks[1]
\usepackage{amsfonts}

\usepackage{slashbox}

\usepackage[titletoc,title]{appendix}

\newtheorem{definition}{Definition}
\newtheorem{theorem}{Theorem}
\newtheorem{example}{Example}

\usepackage{hyperref}
\hypersetup{
colorlinks=true,
linkcolor=red,
anchorcolor=blue,
citecolor=green
}

\usepackage[dvipsnames,table]{xcolor}
\usepackage{multirow}
\usepackage{subcaption}
\biboptions{sort&compress}

\usepackage{tikz}
\usetikzlibrary{shapes,arrows,matrix,positioning,patterns,through,calc,intersections}
\usepackage{subcaption}

\mathchardef\mhyphen="2D
\makeatletter
\newcommand*\bigcdot{\mathpalette\bigcdot@{.5}}
\newcommand*\bigcdot@[2]{\mathbin{\vcenter{\hbox{\scalebox{#2}{$\m@th#1\bullet$}}}}}
\makeatother

\begin{document}

\begin{frontmatter}

\title{Three-Way Decision with Incomplete Information Based on Similarity and Satisfiability}

\author[A,B]{Junfang Luo}
\ead{junfangluo@163.com}

\author[B]{Mengjun Hu\corref{cor1}}
\ead{hu258@cs.uregina.ca}

\author[A]{Keyun Qin}
\ead{keyunqin@263.net}

\address[A]{College of mathematics, Southwest Jiaotong University, Chengdu 610031, P.R. China}
\address[B]{Department of Computer Science, University of Regina, Regina SK S4S 0A2, Canada}
\cortext[cor1]{Corresponding author.}

\begin{abstract}
Three-way decision is widely applied with rough set theory to learn classification or decision rules. The approaches dealing with complete information are well established in the literature, including the two complementary computational and conceptual formulations. The computational formulation uses equivalence relations, and the conceptual formulation uses satisfiability of logic formulas. In this paper, based on a briefly review of these two formulations, we generalize both formulations into three-way decision with incomplete information that is more practical in real-world applications. For the computational formulation, we propose a new measure of similarity degree of objects as a generalization of equivalence relations. Based on it, we discuss two approaches to three-way decision using $\alpha$-similarity classes and approximability of objects, respectively. For the conceptual formulation, we propose a measure of satisfiability degree of formulas as a quantitative generalization of satisfiability with complete information. Based on it, we study two approaches to three-way decision using $\alpha$-meaning sets of formulas and confidence of formulas, respectively. While using similarity classes is a common method of analyzing incomplete information in the literature, the proposed concept of approximability and the two approaches in conceptual formulation point out new promising directions.
\end{abstract}

\begin{keyword}
three-way decision; rough set; incomplete information; similarity; satisfiability; fuzzy logic
\end{keyword}

\end{frontmatter}

\section{Introduction}
\label{sec:introduction}

Three-way decision is about thinking, processing, and computing in threes~\cite{yao2018three}. It is inspired by a common practice of human decision-making with three options, for example, acceptance, rejection, and non-commitment. The idea of three-way decisions has been investigated and applied in many areas, such as three-way classification~\cite{fang2019cost,hu2014decision,li2017cognitive,liu2016novel,maldonado2018credit,sun2018three,yang2019sequential}, three-way clustering~\cite{lingras2004interval,mitra2006rough,pedrycz2005cluster,yu2019ensemble}, three-way recommendation~\cite{azam2014game,zhang2016recommender}, and three-way concept analysis~\cite{zhi2019three,qi2016connections,ren2016attribute}. Yao investigates the interplay of three-way decisions and cognitive computing~\cite{yao2016three} and the relationship between three-way decisions and granular computing~\cite{yao2018three}, which promote the development of fundamental three-way decision models. In the most recent works, Yao~\cite{yao2019tri} studies tri-level thinking approaches to building concrete models of three-way decision. Based on the three-way granular computing, Yao~\cite{yao2020three} unifies rough-set concept analysis and formal concept analysis. Three-way decision under incomplete information extends potential applications of standard rough sets and is worthy of further investigation.

The computational and conceptual formulations of three-way decision in rough set theory are widely and well studied in the context of complete information. While the computational formulation focuses on constructive algorithms and procedures, the conceptual formulation focuses on semantics interpretation of the involved concepts. Although the two formulations are proved to be mathematically equivalent~\cite{yao2015two}, it is necessary to investigate both of them in order to gain a full understanding of three-way decision in rough set theory. Besides, the two formulations inspire different generalizations into three-way decision with incomplete information. Most existing studies on incomplete information~\cite{kryszkiewicz1998rough,lipski1981databases,orlowska1998introduction,guan2006set,grzymala2005incomplete,wang2001rough,luo2013incremental} follow a path of generalizing the computational formulation, which leads to a few efficient approaches. However, they face a common challenge in interpreting semantics and formulating decision rules. In this paper, we investigate both the computational and conceptual formulations, which results in computationally efficient and semantically sound approaches to learning three-way decision rules with incomplete information.

In order to obtain a description of a given class of objects, rough set theory constructs definable or describable sets of objects from a dataset and uses them as building blocks to gain an approximate description of the given class. As a result, the given class is approximately described through the precise descriptions of the definable building blocks. Three-way decision or classification rules are popular in formulating the approximate description of the given class. A set of acceptance rules classifies positive instances of the given class, a set of rejection rules classifies negative instances, and a set of non-commitment rules makes indefinite or delayed decisions. In the situation of complete information where the dataset shows complete or perfect knowledge of objects, the computational and conceptual formulations of rough sets mainly differ in their interpretation and construction of definable building blocks. More specifically, the computational formulation uses equivalence relations to model the indiscernibility of objects. Accordingly, a set of indiscernible objects has a uniform and precise description implied by the equivalence relation and thus, is a definable building block. In contrast, the conceptual formulation directly studies the descriptions which are usually logic formulas. A set of objects satisfying a formula is then used as a definable building block.

While the key concepts of indiscernibility in the computational formulation and satisfiability in the conceptual formulation are straightforward with complete information, they are much more complex to be defined with incomplete information due to different perspectives and interpretations of incompleteness. Many efforts~\cite{kryszkiewicz1998rough,lipski1981databases,orlowska1998introduction,stefanowski2001incomplete,guan2006set,grzymala2005incomplete,liu2016novel,luo2015rough,wang2001rough,luo2013incremental} have been made to generalize equivalence relations modelling indiscernibility into similarity or tolerance relations. This direction leads to the computational formulation with incomplete information. Kryszkiewicz~\cite{kryszkiewicz1998rough} presents a definition of tolerance relation to learn decision rules with incomplete information. Stefanowski and Tsouki{\`a}s~\cite{stefanowski2001incomplete} generalize Kryszkiewicz's tolerance relation and present a valued tolerance relation. Liu, Liang, and Wang~\cite{liu2016novel} define another relation to describe the similarity degree between objects. As an improvement of their work, Luo and Qin~\cite{luo2015rough} propose a graded indiscernibility relation. Although there are many similarity or tolerance relations proposed for constructing definable building blocks with incomplete information, little work discusses the learning and interpretation of decision rules due to a lack of descriptions of the definable building blocks.

In this paper, we propose a new measure of similarity degree of objects based on an in-depth examination of different semantics of incomplete information. Based on the proposed similarity degree, we further study two approaches to the computational formulation of three-way decision with incomplete information. The first approach applies a threshold $\alpha$ on the similarity degree and constructs $\alpha$-similarity classes as definable building blocks. Moreover, we present a way to describe and interpret the semantics of $\alpha$-similarity classes so that three-way decision rules can be easily formulated. The second approach uses the similarity degree to calculate positive and negative approximability of objects which is a concept inspired by the ``approximability'' proposed by Stefanowski and Tsouki{\`a}s~\cite{stefanowski2001incomplete}. Three-way decision rules are derived by applying a threshold on the approximability.

Compared with the computational formulation, the conceptual formulation with incomplete information is not widely studied by researchers despite its superiority in interpreting the semantics of the involved concepts. Grzyma{\l}a-Busse and Hu~\cite{grzymala2000comparison,grzymala2004characteristic} propose the notion of a block of a description (more specifically, an attribute-value pair) which is a set of objects satisfying the description in the context of incomplete information. Hu and Yao~\cite{hu2018structured} study the conceptual formulation by constructing an interval set~\cite{yao2009interval} that satisfies a formula as a definable building block. In this paper, we consider a quantitative generalization of measuring the satisfiability degree of formulas by objects. Based on it, we propose two approaches to the conceptual formulation of three-way decision with incomplete information. The first approach applies a threshold $\alpha$ on the satisfiability degree and constructs $\alpha$-meaning sets of formulas as definable building blocks. The $\alpha$-meaning set of a formula includes objects that satisfy the formula to a degree of at least $\alpha$, and it is considered to be defined or described by the formula. The second approach calculates acceptance and rejection confidence of formulas through the satisfiability degree, which respectively indicates the degrees to which a formula can induce an acceptance or a rejection decision rule. Three-way decision rules can be conveniently formulated by applying a threshold on the confidence of formulas.

The remainder of this paper is organized as follows. Section~\ref{sec:complete} summarizes the computational and conceptual formulations of three-way decision in rough sets with complete information. Based on it, the following two sections investigate the generalizations of these two formulations into incomplete information. Section~\ref{sec:computational_incomplete} studies the computational formulation based on the similarity degree of objects and Section~\ref{sec:conceptual_incomplete} studies the conceptual formulation based on the satisfiability degree of formulas by objects. The main results are summarized in Section \ref{sec:conclusions} along with a discussion of future work.

\section{Three-way decision in a complete table}
\label{sec:complete}

In this section, we review the computational and conceptual formulations of three-way decision in rough set with complete information. 
In addition, we propose a new definition of approximations at the end of this section, which will be used thereafter in our work with incomplete information.

Rough set analysis starts from an information table that represents a given dataset. In this section, we restrict our discussion in a complete table~\cite{pawlak1991theoretical}. Such a table can be formally represented by a tuple $T=(O\!B,AT,V,F)$ where:
\begin{enumerate}[label=(\arabic*)]
    \item $O\!B$ is a finite nonempty set of objects called the universe;
    \item $AT$ is a finite nonempty set of attributes;
    \item $V=\bigcup_{a\in AT} {V_a}$ where $V_a$ is the domain of an attribute $a$;
    \item $F=\{f_a: O\!B \rightarrow V_a \mid a \in AT\}$ where $f_a$ is an information function that maps each object $x \in O\!B$ to a unique value $f_a(x)\in V_a$ on an attribute $a \in AT$.
\end{enumerate}

\begin{figure}[ht!]
\centering
\scalebox{0.8}{
\begin{tikzpicture}[auto, >=stealth', node distance=1em,
block_multilines1/.style ={rectangle, draw=black, thick, fill=black!20, text width=8em, text centered, minimum height=2em},
block_multilines2/.style ={rectangle, draw=black, thick, fill=white,text width=11em, text centered, minimum height=3em},
block_multilines3/.style ={rectangle, draw=white, fill=white, text width=9em, text centered, minimum height=1em},
block_multilines4/.style ={rectangle, dashed, draw=black, thick, fill=white, text width=8em, text centered, minimum height=2em},
noborder_center/.style ={rectangle, draw=white, fill=white, text width=10em,text centered, minimum height=2em}]
\node [block_multilines2](nCOM){A complete \\ table};
\node [block_multilines2, below left=4.5em and -1em of nCOM](nIND){An equivalence relation};
\node [block_multilines2, below=2em of nIND] (nINDx) {Equivalence classes};
\node [block_multilines2, below right=4.5em and -1em of nCOM](n1){A description language: \\ conjunctive formulas};
\node [block_multilines2, below=2em of n1] (n2) {Conjunctively definable sets};
\node [block_multilines3, above left=2em and -4.5 of nIND](n5){{\large\textcircled{\small 1}} Computational};
\node [block_multilines3, above right=2em and -4.5 of n1](n6){{\large\textcircled{\small 2}} Conceptual};
\node [block_multilines2, below=16.5em of nCOM](nLUIND) {Three-way rough set approximations};
\node [block_multilines2, below=2em of nLUIND] (nRIND) {Three-way decision\\ rules};
\draw [dashed,thick](-17.5em,-5em) rectangle (17.5em,-15em);
\node [block_multilines3, below left=7.75em and 12em of nCOM](n7){${\bullet}$\textbf{Definibility}};
\node [block_multilines3, left =11em of nLUIND](n8){${\bullet}$\textbf{Approximations}};
\node [block_multilines3, left =10.2em of nRIND](n9){${\bullet}$\textbf{Classification rules}};
\draw [->,thick] (nCOM.south) -- (n1.north);
\draw [->,thick] (n1.south) -- (n2.north);
\draw [->,thick] (nCOM.south) -- (nIND.north);
\draw [->,thick] (nIND.south) -- (nINDx.north);
\draw [->,thick] (n2.south) -- (nLUIND.north);
\draw [->,thick] (nINDx.south) -- (nLUIND.north);
\draw [->,thick] (nLUIND.south) -- (nRIND.north);
\end{tikzpicture}}
\caption{The computational and conceptual formulations of three-way decision with complete information}
\label{fig:computational_conceptual_complete}
\end{figure}
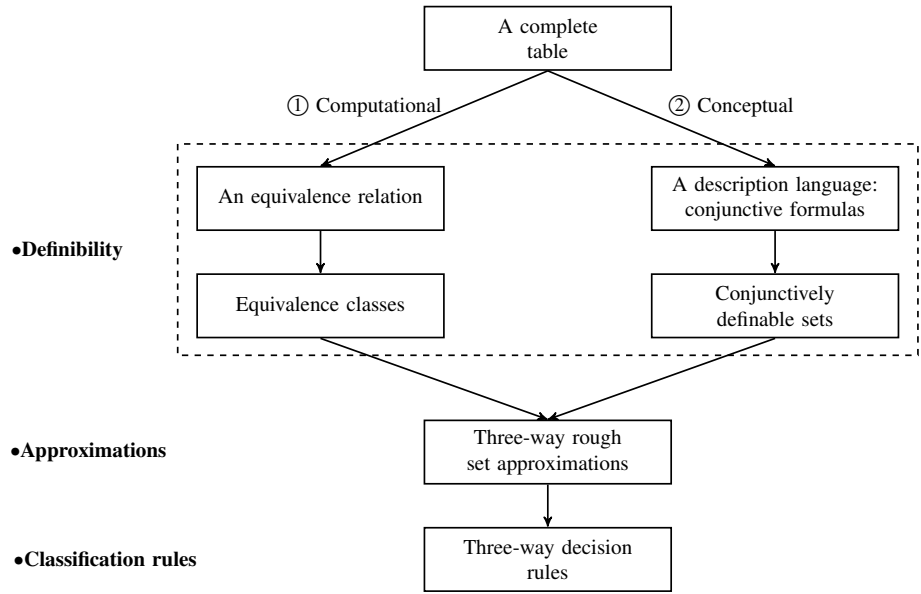

There are commonly two formulations of three-way decision with a complete table, namely, the computational and conceptual formulations~\cite{yao2015two}. Their frameworks are summarized in Figure~\ref{fig:computational_conceptual_complete} to gain insights on how three-way decision rules are obtained from a complete table through these two formulations. These two formulations mainly differ in the interpretation of the notion of definability and the construction of definable sets. Though the two formulations are mathematically equivalent, they offer different perspectives of the semantics of definability and inspire different generalizations into incomplete tables.

\subsection{Computational formulation}

The computational formulation interprets definability and constructs definable sets based on equivalence relations that are used to model the indiscernibility of objects. In this section, we summarize the approaches to learning three-way decision rules in the computational formulation based on Pawlak's work~\cite{pawlak1991theoretical}, Yao's work~\cite{yao2015two}, and some others~\cite{bryniarski1989calculus,hu2018structured}.

\begin{definition}
For a set of attributes $A\subseteq AT$, an equivalence relation ${E}_A$ on $O\!B$ is defined as:
\begin{equation}
{E}_A=\{(x,y)\in O\!B\times O\!B\mid \forall\, a\in A~(f_a(x)=f_a(y))\}.
\end{equation}
\end{definition}

Two objects are indiscernible or equivalent with respect to $A$ if they have the same values on all the attributes in $A$. In other words, the two objects cannot be distinguished under these attributes. On the other hand, two objects are discernible with respect to $A$ if they have different values on at least one attribute in $A$. One can easily verify that ${E}_A=\bigcap_{a\in A}{E_{\{a\}}}$~\cite{pawlak1991theoretical}.

Based on the equivalence relation $E_A$, we can group the indiscernible objects, which results in the equivalence classes. For an object $x \in O\!B$, its equivalence class induced by $E_A$ is:
\begin{equation}
[x]_A=\{y\in O\!B\mid (x,y)\in{E_A}\}.
\end{equation}
The family of all equivalence classes induced by $E_A$ forms a partition of $O\!B$, denoted as $O\!B/E_A=\{[x]_A\mid x\in O\!B\}$. That is, the different equivalence classes in $O\!B/E_A$ are nonempty pair-wise disjoint subsets of $O\!B$ whose union is $O\!B$. For a sequence of sets of attributes $A_1 \subseteq A_2 \subseteq \cdots \subseteq A_n \subseteq AT$, one can easily verify that: for any $x \in O\!B$,
\begin{equation}
[x]_{AT}\subseteq [x]_{A_n} \subseteq \cdots \subseteq [x]_{A_2} \subseteq [x]_{A_1}.
\end{equation}
The equivalence class $[x]_{AT}$ with respect to all attributes $AT$ is the smallest equivalence class containing $x$. For a set of attributes $A \subseteq AT$, an equivalence class $[x]_A$ is a definable set in the sense that it can be defined or described as:
\begin{equation}
{\rm Des}([x]_A) = \bigwedge_{a \in A} \langle a,f_a(x)\rangle,
\end{equation}
where $\langle a,f_a(x)\rangle$ describes that a value $f_a(x)$ is taken on an attribute $a$ and $\bigwedge$ is the classical logic conjunction.

Though a precise description is always desired for a set of objects, an arbitrary set $X \subseteq O\!B$ may not be definable. In order to describe an indefinable set which usually represents a given class, one can approximate it by using a family of definable sets as the building blocks. There are two popular formulations of rough set approximations in the literature, namely, a pair of upper and lower approximations~\cite{pawlak1991theoretical} and three-way approximations~\cite{yao2018three} consisting of three regions. We consider the latter for its superiority in rule induction.

\begin{definition}
\label{def:three_regions_complete_computational}
Given a complete table $T=(O\!B,AT,V,F)$, for a set of attributes $A \subseteq AT$, the positive, negative, and boundary regions regarding a given class $X\subseteq O\!B$ can be defined as:
\begin{eqnarray}
\label{equa:three_regions_complete_computational}
{\rm POS}^1_A(X)&=&\{[x]_A \in O\!B/E_A \mid [x]_A \subseteq X\},\nonumber\\
{\rm NEG}^1_A(X)&=&\{[x]_A \in O\!B/E_A \mid [x]_A \subseteq X^c\},\nonumber\\
{\rm BND}^1_A(X)&=&\{[x]_A \in O\!B/E_A \mid \neg([x]_A \subseteq X) \wedge \neg([x]_A \subseteq X^c)\}\nonumber\\
&=& ({\rm POS}^1_A(X) \cup {\rm NEG}^1_A(X))^c,
\end{eqnarray}
where $^c$ denotes the set-complement.
\end{definition}

The three regions given in Definition~\ref{def:three_regions_complete_computational} are referred to as structured approximations~\cite{bryniarski1989calculus,ma2017structured,hu2018structured} in the sense that the internal structure is explicitly represented by preserving individual definable building blocks. By taking union over equivalence classes in each individual region, one can get the unstructured approximations~\cite{pawlak1991theoretical} that are sets of objects instead of sets of equivalence classes. Though the structured and unstructured approximations are mathematically equivalent, the structured approximations are superior in deriving and interpreting classification rules since one definable building block explicitly corresponds to one classification rule and vice versa, as argued in~\cite{hu2018structured}. For this reason, we focus on structured approximations in this paper.

Three types of decision rules can be easily induced based on the descriptions of equivalence classes in the three structured regions.

\begin{definition}
\label{def:rules_complete_computational}
Given a set of objects $X \subseteq O\!B$ and a set of attributes $A \subseteq AT$, the three types of acceptance {\rm (A)}, rejection {\rm (R)}, and non-commitment {\rm (N)} decision rules regarding $X$ are defined as: for any $y\in O\!B$,
\begin{eqnarray*}
{\rm (A)} && {\rm Des}([x_a]_A)\longrightarrow_{\rm A} X: \textit{~if~} y\models {\rm Des}([x_a]_A) \textit{~for~} [x_a]_A\in {\rm POS}^1_A(X), \textit{~then~accept~} y,\\
{\rm (R)} && {\rm Des}([x_r]_A)\longrightarrow_{\rm R} X: \textit{~if~} y\models {\rm Des}([x_r]_A) \textit{~for~} [x_r]_A\in {\rm NEG}^1_A(X), \textit{~then~reject~} y,\\
{\rm (N)} && {\rm Des}([x_n]_A)\longrightarrow_{\rm N} X: \textit{~if~} y\models {\rm Des}([x_n]_A) \textit{~for~} [x_n]_A\in {\rm BND}^1_A(X), \textit{~then~neither~accept~nor~reject~} y,
\end{eqnarray*}
where $y\models {\rm Des}([x_{\cdot}]_A)$ means that $y$ can also be described by ${\rm Des}([x_{\cdot}]_A)$.
\end{definition}

For an equivalence class $[x_a]_A$ in the positive region, its description ${\rm Des}([x_a]_A)$ induces an acceptance rule, which is used to accept positive instances of the given class $X$. For an equivalence class $[x_r]_A$ in the negative region, its description ${\rm Des}([x_r]_A)$ induces a rejection rule for rejecting an object to be a positive instance, or equivalently, accepting it to be a negative instance. For an equivalence class $[x_n]_A$ in the boundary region, its description ${\rm Des}([x_n]_A)$ induces a non-commitment rule which cannot be used to make a definite decision. In existing studies, the boundary region is widely defined as the complement of the union of the positive and negative regions, as given in Definition \ref{def:three_regions_complete_computational}. On the other hand, from the viewpoint of deriving rules from approximations, the boundary region leads to a non-commitment decision which can be given as a complement of the acceptance and rejection decisions. In other words, a non-commitment decision is made whenever neither an acceptance nor a rejection decision can be made:
\begin{eqnarray*}
{\rm (N)} && \textit{If~none~of~the~rules~in~{\rm(A)}~and~{\rm(R)}~applies,~then~neither~accept~nor~reject~} y.
\end{eqnarray*}
For the above reasons, in the following discussion, we will only explicitly define the positive and negative regions  and assume that the boundary region is defined as the complement of their union.

The computational formulation provides an efficient way to construct a family of definable building blocks (i.e., the equivalence classes) which is used to construct the approximations and induce the three-way classification rules. Though the definition of an equivalence relation indicates the descriptions of equivalence classes (i.e., $\bigwedge_{a \in A} \langle a,f_a(x)\rangle$ for $[x]_A$), these descriptions are not explicitly explained or used throughout the construction of equivalence classes and approximations. This leads to inconvenience in deriving rules from approximations where these descriptions are necessary. The computational formulation focuses more on efficiently constructing a family of definable building blocks rather than explaining their semantics.

\begin{example}
\label{example:complete_computational}

We illustrate the computational formulation with a complete table $T$ given in Table~\ref{tab:example_complete_table}. Specifically, we have $O\!B = \{x_1,x_2,x_3,x_4,x_5,x_6\}$ and $AT = \{a_1,a_2,a_3\}$. The domains of the three attributes are $V_{a_1} = \{0,1\}$, $V_{a_2} = \{1,2\}$, and $V_{a_3} = \{1,3\}$. 

\begin{table}[ht!]
\centering
\caption{A complete table $T$}
\label{tab:example_complete_table}
\setlength{\tabcolsep}{1em}
\renewcommand{\arraystretch}{1.2}
\begin{tabular}{c|ccc}
\hline
& $a_1$ & $a_2$& $a_3$   \\
\hline
$x_1$ &1&2&3\\
$x_2$ &1&2&3\\
$x_3$ &0&1&1\\
$x_4$ &0&2&1\\
$x_5$ &0&2&1\\
$x_6$ &1&1&1\\
\hline
\end{tabular}
\end{table}

Suppose we consider all attributes in $AT$ and get a family of equivalence classes:
\begin{equation}
O\!B/E_{AT} = \big\{ \{x_1,x_2\}, \{x_3\}, \{x_4,x_5\}, \{x_6\} \big\}.
\end{equation}
For a given class $X=\{x_1,x_2,x_3,x_4\}$, its structured three-way approximations are:
\begin{eqnarray}
{\rm POS}^1_{AT}(X)&=&\big\{\{x_1,x_2\}, \{x_3\}\big\},\nonumber\\
{\rm NEG}^1_{AT}(X)&=&\big\{\{x_6\}\big\},\nonumber\\
{\rm BND}^1_{AT}(X)&=&\big\{\{x_4,x_5\}\big\}.
\end{eqnarray}
Correspondingly, we get the three types of decision rules as follows:
\begin{eqnarray*}
{\rm (A)} && \langle a_1,1\rangle\wedge\langle a_2,2\rangle\wedge\langle a_3,3\rangle \longrightarrow_{\rm A} X,\\
&& \langle a_1,0\rangle\wedge\langle a_2,1\rangle\wedge\langle a_3,1\rangle \longrightarrow_{\rm A} X;\\
{\rm (R)} && \langle a_1,1\rangle\wedge\langle a_2,1\rangle\wedge\langle a_3,1\rangle \longrightarrow_{\rm R} X; \\
{\rm (N)}  && \textit{Otherwise,~neither~accept~nor~reject~} y.
\end{eqnarray*}

It should be noted that different choices of $A \subseteq AT$ may lead to equivalence classes consisting of the same set of objects but with different descriptions. For example, for a set of attributes $\{a_1,a_2\}$, we have $[x_1]_{\{a_1,a_2\}} = \{x_1,x_2\}$ which is equal to $[x_1]_{AT}$. However, $[x_1]_{\{a_1,a_2\}}$ is described as $\langle a_1,1\rangle\wedge\langle a_2,2\rangle$ and $[x_1]_{AT}$ is described as $\langle a_1,1\rangle\wedge\langle a_2,2\rangle\wedge\langle a_3,3\rangle$. We list all the partitions induced by all possible equivalence relations as follows where the equivalence classes in the corresponding positive and negative regions are underlined:
\begin{eqnarray}
\label{equa:example_complete_all_equivalence_classes}
O\!B/\emptyset &=& \big\{ \{x_1,x_2,x_3,x_4,x_5,x_6\} \big\},\nonumber\\
O\!B/\{a_1\} &=& \big\{ \{x_1,x_2,x_6\}, \{x_3,x_4,x_5\} \big\},\nonumber\\
O\!B/\{a_2\} &=& \big\{ \{x_1,x_2,x_4,x_5\}, \{x_3,x_6\} \big\},\nonumber\\
O\!B/\{a_3\} &=& \big\{ \underline{\{x_1,x_2\}}, \{x_3,x_4,x_5,x_6\} \big\},\nonumber\\
O\!B/\{a_1,a_2\} &=& \big\{ \underline{\{x_1,x_2\}}, \underline{\{x_3\}}, \{x_4,x_5\}, \underline{\{x_6\}}\big\},\nonumber\\
O\!B/\{a_1,a_3\} &=& \big\{ \underline{\{x_1,x_2\}}, \{x_3,x_4,x_5\}, \underline{\{x_6\}} \big\},\nonumber\\
O\!B/\{a_2,a_3\} &=& \big\{ \underline{\{x_1,x_2\}}, \{x_3,x_6\}, \{x_4,x_5\} \big\},\nonumber\\
O\!B/AT &=& \big\{ \underline{\{x_1,x_2\}}, \underline{\{x_3\}}, \{x_4,x_5\}, \underline{\{x_6\}}  \big\}.
\end{eqnarray}
Thus, we get totally five different descriptions of $\{x_1,x_2\}$ and two different descriptions of $\{x_3\}$ in the positive regions and three different descriptions of $\{x_6\}$ in the negative regions. As a result, we can derive the following rules as a combination of all the rules induced by the eight equivalence relations.
\begin{eqnarray*}
{\rm (A)} &&\langle a_3,3\rangle \longrightarrow_{\rm A} X,\nonumber\\
&& \langle a_1,1\rangle\wedge\langle a_2,2\rangle \longrightarrow_{\rm A} X,\nonumber\\
&& \langle a_1,1\rangle\wedge\langle a_3,3\rangle \longrightarrow_{\rm A} X,\nonumber\\
&& \langle a_2,2\rangle\wedge\langle a_3,3\rangle \longrightarrow_{\rm A} X,\nonumber\\
&& \langle a_1,1\rangle\wedge\langle a_2,2\rangle\wedge\langle a_3,3\rangle \longrightarrow_{\rm A} X, \nonumber\\
&& \langle a_1,0\rangle\wedge\langle a_2,1\rangle \longrightarrow_{\rm A} X, \nonumber\\
&& \langle a_1,0\rangle\wedge\langle a_2,1\rangle\wedge\langle a_3,1\rangle \longrightarrow_{\rm A} X;\nonumber\\
{\rm (R)} && \langle a_1,1\rangle\wedge\langle a_2,1\rangle \longrightarrow_{\rm R} X, \nonumber\\
&& \langle a_1,1\rangle\wedge\langle a_3,1\rangle \longrightarrow_{\rm R} X, \nonumber\\
&& \langle a_1,1\rangle\wedge\langle a_2,1\rangle\wedge\langle a_3,1\rangle \longrightarrow_{\rm R} X; \nonumber\\
{\rm (N)}  && \textit{Otherwise,~neither~accept~nor~reject~} y.
\end{eqnarray*}

\end{example}

\subsection{Conceptual formulation}
\label{sec:complete_conceptual}

The conceptual formulation explains definability and constructs definable sets based on a description language. Hu and Yao~\cite{hu2018structured} argue that, from the viewpoint of learning three-way decision rules, it is sufficient to consider a conjunctive description language ${\rm CDL}_A$ that contains only conjunctive logic formulas. Based on their work, we summarize the conceptual formulation of learning three-way decision rules from a complete table. The language ${\rm CDL}_A$ is a sublanguage of the description language used by Pawlak~\cite{pawlak1991theoretical} and Marek and Pawlak~\cite{marek1976information}.

\begin{definition}
\label{def:CDL}
For a set of attributes $A\subseteq AT$, the conjunctive description language ${\rm CDL}_A$ consists of a set of conjunctive formulas that are recursively defined in the following two cases~\cite{hu2018structured}:
\begin{enumerate}[label=(\arabic*)]
\item Atomic formulas (attribute-value pair): $\langle a,v\rangle\in {\rm CDL}_A$, where $a\in A$, $v\in V_a$,\label{item: formula}
\item Composite formulas: $p \wedge q\in {\rm CDL}_A$, where $p, q \in {\rm CDL}_A$, $p$ and $q$ do not share any attribute.
\end{enumerate}
\end{definition}

With respect to a set of attributes $A\subseteq AT$, a given object $x\in O\!B$ can be described as $\bigwedge_{a\in A}\langle a,f_a(x)\rangle$. An object satisfies a formula if it takes all the designated values on the corresponding attributes as given by the formula.

\begin{definition}
The satisfiability of a formula by an object $x\in O\!B$, denoted by $\models$, is defined as:
\begin{eqnarray}
(1) & &x\models \langle a,v\rangle ~{\rm iff}~ f_a(x)=v, \nonumber\\
(2) & &x\models p\wedge q ~{\rm iff}~(x\models p)~{\rm and}~ (x\models q),
\end{eqnarray}
where $a\in A, v\in V_a$, and $p,q,p \wedge q \in {\rm CDL}_A$.
\end{definition}

The formulas are used to describe objects. In turn, the set of objects satisfying a formula can be used to demonstrate the meaning of a formula~\cite{hu2018structured}.
\begin{definition}
\label{def:meaning_set_complete}
Given a formula $p\in {\rm CDL}_A$, the set of objects satisfying $p$ is called the meaning set of $p$, which is formally defined as:
\begin{equation}
m(p)=\{x\in O\!B\mid x\models p\}.
\end{equation}
\end{definition}
One can interpret logic conjunction in a formula through set-intersection by means of meaning sets:
\begin{eqnarray}
\label{equa:conjunctive_intersection}
(1) &&m(\langle a,v\rangle)=\{x\in O\!B\mid f_a(x)=v\},\nonumber\\
(2) &&m(p\wedge q)=m(p)\cap m(q),
\end{eqnarray}
where $a\in A, v\in V_a$, and $p,q,p \wedge q \in {\rm CDL}_A$.
Accordingly, one can compute the meaning set of a composite formula through the meaning sets of atomic formulas.

The objects in a meaning set $m(p)$ can be uniformly described by the conjunctive formula $p$. In this sense, $m(p)$ is called a conjunctively definable set~\cite{hu2018structured}.

\begin{definition}
A set of objects $Y\subseteq O\!B$ is a conjunctively definable set regarding a set of attributes $A \subseteq AT$ if there exists a formula $p\in {\rm CDL}_A$ such that $Y=m(p)$.
\end{definition}

The family of conjunctively definable sets with respect to $A \subseteq AT$ in an information table $T$ is denoted as:
\begin{equation}
{\rm CDEF}_A(T)=\{m(p) \mid  p \in {\rm CDL}_A\}.
\end{equation}
This family is used as building blocks to construct the three-way approximations.

\begin{definition}
\label{def:three_regions_complete_conceptual}
Given a complete table $T=(O\!B,AT,V,F)$, for a set of attributes $A \subseteq AT$, the positive and negative regions regarding a given class $X\subseteq O\!B$ can be defined as:
\begin{eqnarray}
\label{equa:three_regions_complete_conceptual}
{\rm POS}^2_A(X)&=&\{ m(p) \in {\rm CDEF}_A(T) \mid m(p) \ne \emptyset, m(p) \subseteq X\},\nonumber\\
{\rm NEG}^2_A(X)&=&\{ m(p) \in {\rm CDEF}_A(T) \mid m(p) \ne \emptyset, m(p) \subseteq X^c\}.
\end{eqnarray}
\end{definition}

Three types of decision rules can be easily induced by using the conjunctive formulas associated with the structured regions.

\begin{definition}
\label{def:rules_complete_conceptual}
Given a set of objects $X \subseteq O\!B$ and a set of attributes $A \subseteq AT$, the three types of acceptance, rejection, and non-commitment decision rules regarding $X$ are defined as: for any $y\in O\!B$,
\begin{eqnarray*}
{\rm (A)} && p_a \longrightarrow_{\rm A} X: \textit{~if~} y\models p_a \textit{~for~} m(p_a)\in {\rm POS}_A^2(X), \textit{~then~accept~} y,\\
{\rm (R)} && p_r \longrightarrow_{\rm R} X: \textit{~if~} y\models p_r \textit{~for~} m(p_r)\in {\rm NEG}_A^2(X), \textit{~then~reject~} y,\\
{\rm (N)} && \textit{If~none~of~the~rules~in~{\rm(A)}~and~{\rm(R)}~applies,~then~neither~accept~nor~reject~} y.
\end{eqnarray*}
\end{definition}

It should be noted that a conjunctively definable set may be defined by more than one formula in ${\rm CDL}_A$. Accordingly, one conjunctively definable set in a region may result in more than one decision rule.

Compared with the computational formulation, the conceptual formulation derives a family of definable building blocks (i.e., ${\rm CDEF}_A(T)$) based on the available descriptions (i.e., formulas in ${\rm CDL}_A$). In this sense, the semantics of definable building blocks is explicitly given and used in the formulation, which benefits the rule induction from approximations. The conceptual formulation focuses more on interpreting the definable building blocks rather than efficiently constructing them.

\begin{example}
\label{example:complete_conceptual}

We illustrate the conceptual formulation with Table~\ref{tab:example_complete_table}. Suppose we consider all attributes in $AT$. Table \ref{tab:example_complete_conceptual_sets} lists all the conjunctive formulas in ${\rm CDL}_{AT}$ and the corresponding meaning sets that compose the family ${\rm CDEF}_{AT}(T)$. For the convenience of our discussion, we list the family ${\rm CDEF}_{AT}(T)$:
\begin{eqnarray}
\label{equa:example_complete_CDEF}
{\rm CDEF}_{AT}(T) &=& \big\{ \emptyset, \{x_3\}, \{x_6\}, \{x_1,x_2\}, \{x_3,x_6\}, \{x_4,x_5\}, \{x_1,x_2,x_6\},\nonumber\\
&&~~ \{x_3,x_4,x_5\}, \{x_1,x_2,x_4,x_5\}, \{x_3,x_4,x_5,x_6\} \big\}.
\end{eqnarray}

\begin{table}[ht!]
\centering
\caption{{Conjunctive formulas in ${\rm CDL}_{AT}$ and conjunctively definable sets in ${\rm CDEF}_{AT}(T)$ for Table~\ref{tab:example_complete_table}}}
\label{tab:example_complete_conceptual_sets}
\renewcommand{\arraystretch}{1.2}
\scalebox{0.8}{
\begin{tabular}{|l|l||l|l|}
    \hline
   Formulas in ${\rm CDL}_{AT}$ & Meaning sets in ${\rm CDEF}_{AT}(T)$ & Formulas in ${\rm CDL}_{AT}$ & Meaning sets in ${\rm CDEF}_{AT}(T)$ \\
    \hline
$p_1=\langle a_1,0\rangle$ & $\{x_3,x_4,x_5\}$&
$p_{14}=\langle a_1,1\rangle\wedge \langle a_3,1\rangle$&$\{x_6\}$\\
$p_2=\langle a_1,1\rangle$ &$\{x_1,x_2,x_6\}$&
$p_{15}=\langle a_2,1\rangle\wedge \langle a_3,3\rangle$ &$\emptyset$\\
$p_3=\langle a_2,1\rangle$ &$\{x_3,x_6\}$&
$p_{16}=\langle a_2,1\rangle\wedge \langle a_3,1\rangle $ &$\{x_3,x_6\}$\\
$p_4=\langle a_2,2\rangle$ &$\{x_1,x_2,x_4,x_5\}$&
$p_{17}=\langle a_2,2\rangle\wedge \langle a_3,3\rangle$&$\{x_1,x_2\}$\\
$p_5=\langle a_3,3\rangle$&$\{x_1,x_2\}$&
$p_{18}=\langle a_2,2\rangle\wedge \langle a_3,1\rangle$&$\{x_4,x_5\}$\\
$p_6=\langle a_3,1\rangle$ &$\{x_3,x_4,x_5,x_6\}$&
$p_{19}=\langle a_1,0\rangle\wedge \langle a_2,1\rangle\wedge \langle a_3,3\rangle$&$\emptyset$\\
$p_7=\langle a_1,0\rangle\wedge \langle a_2,1\rangle$& $\{x_3\}$ &
$p_{20}=\langle a_1,0\rangle\wedge \langle a_2,1\rangle\wedge \langle a_3,1\rangle$&$\{x_3\}$\\
$p_8=\langle a_1,0\rangle\wedge \langle a_2,2\rangle$ &$\{x_4,x_5\}$&
$p_{21}=\langle a_1,0\rangle\wedge \langle a_2,2\rangle\wedge \langle a_3,3\rangle$ &$\emptyset$\\
$p_9=\langle a_1,0\rangle\wedge \langle a_3,3\rangle$  &$\emptyset$&
$p_{22}=\langle a_1,0\rangle\wedge \langle a_2,2\rangle\wedge \langle a_3,1\rangle$& $\{x_4,x_5\}$\\
$p_{10}=\langle a_1,0\rangle\wedge \langle a_3,1\rangle$&$\{x_3,x_4,x_5\}$&
$p_{23}=\langle a_1,1\rangle\wedge \langle a_2,1\rangle\wedge \langle a_3,3\rangle$&$\emptyset$\\
$p_{11}=\langle a_1,1\rangle\wedge \langle a_2,1\rangle$&$\{x_6\}$&
$p_{24}=\langle a_1,1\rangle\wedge \langle a_2,1\rangle\wedge \langle a_3,1\rangle$&$\{x_6\}$\\
$p_{12}=\langle a_1,1\rangle\wedge \langle a_2,2\rangle$&$\{x_1,x_2\}$&
$p_{25}=\langle a_1,1\rangle\wedge \langle a_2,2\rangle\wedge \langle a_3,3\rangle $& $\{x_1,x_2\}$\\
$p_{13}=\langle a_1,1\rangle\wedge \langle a_3,3\rangle$&$\{x_1,x_2\}$&
$p_{26}=\langle a_1,1\rangle\wedge \langle a_2,2\rangle\wedge \langle a_3,1\rangle $&$\emptyset$\\
\hline
\end{tabular}
}
\end{table}

For a given class $X = \{x_1,x_2,x_3,x_4\}$, its positive and negative regions are:
\begin{eqnarray}
{\rm POS}^2_{AT}(X)&=&\big\{ \{x_1,x_2\}, \{x_3\} \big\},\nonumber\\
{\rm NEG}^2_{AT}(X)&=&\big\{ \{x_6\} \big\}.
\end{eqnarray}
Correspondingly, we get the three types of decision rules by looking for all the descriptions of the conjunctively definable sets in ${\rm POS}^2_{AT}(X)$ and ${\rm NEG}^2_{AT}(X)$:
\begin{eqnarray*}
{\rm (A)} &&\langle a_3,3\rangle \longrightarrow_{\rm A} X,\nonumber\\
&& \langle a_1,1\rangle\wedge\langle a_2,2\rangle \longrightarrow_{\rm A} X,\nonumber\\
&& \langle a_1,1\rangle\wedge\langle a_3,3\rangle \longrightarrow_{\rm A} X,\nonumber\\
&& \langle a_2,2\rangle\wedge\langle a_3,3\rangle \longrightarrow_{\rm A} X,\nonumber\\
&& \langle a_1,1\rangle\wedge\langle a_2,2\rangle\wedge\langle a_3,3\rangle \longrightarrow_{\rm A} X, \nonumber\\
&& \langle a_1,0\rangle\wedge\langle a_2,1\rangle \longrightarrow_{\rm A} X, \nonumber\\
&& \langle a_1,0\rangle\wedge\langle a_2,1\rangle\wedge\langle a_3,1\rangle \longrightarrow_{\rm A} X;\nonumber\\
{\rm (R)} && \langle a_1,1\rangle\wedge\langle a_2,1\rangle \longrightarrow_{\rm R} X, \nonumber\\
&& \langle a_1,1\rangle\wedge\langle a_3,1\rangle \longrightarrow_{\rm R} X, \nonumber\\
&& \langle a_1,1\rangle\wedge\langle a_2,1\rangle\wedge\langle a_3,1\rangle \longrightarrow_{\rm R} X; \nonumber\\
{\rm (N)}  && \textit{Otherwise,~neither~accept~nor~reject~} y.
\end{eqnarray*}

We get the same three-way decision rules as the combination of rules induced from all possible equivalence relations in Example \ref{example:complete_computational}. While the computational formulation is more efficient in constructing definable building blocks and approximations (assuming that only one specific $A \subseteq AT$ is considered), the conceptual formulation displays more explicit semantics of definable building blocks and is more convenient for learning rules from approximations.
\end{example}

\subsection{Mathematical equivalence of the two formulations}

Although the computational and conceptual formulations provide different perspectives of interpreting definability, they can induce the same family of definable sets in a more general sense and thus, may be considered mathematically equivalent, as argued by Yao in~\cite{yao2015two}.

For a set of attributes $A\subseteq AT$, the computational formulation gives a family of equivalence classes $O\!B/E_A$ as definable sets and the conceptual formulation gives a family of conjunctively definable sets ${\rm CDEF}_A(T)$. Recall that an equivalence class $[x]_{A}$ can be described by $\bigwedge_{a \in A} \langle a,f_a(x)\rangle$ which is actually a conjunctive formula in ${\rm CDL}_{A}$. This indicates that an equivalence class in $O\!B/E_A$ is also a conjunctively definable set in ${\rm CDEF}_A(T)$. However, in the reverse direction, a conjunctively definable set is not necessarily an equivalence class since a conjunctive formula does not necessarily use every attribute in $A$. It should be noted that in a special case of $A = \emptyset$, we get only one equivalence class $O\!B$ which can be defined by a trivially correct conjunctive formula $p = {\rm True}$ not included in ${\rm CDL}_A$ according to Definition \ref{def:CDL}. Instead of modifying the description language ${\rm CDL}_A$ to include $p={\rm True}$, we assume $A$ to be nonempty in the following discussion since the case of $A = \emptyset$ and the equivalence class including exactly the universe $O\!B$ are not useful from the view of learning classification rules.

\begin{theorem}
\label{theo:subset_relationship}
For a set of attributes $A\subseteq AT$, the following subset relationship holds:
\begin{equation}
O\!B/E_A \subseteq {\rm CDEF}_A(T).
\end{equation}
\end{theorem}

By noticing the above subset relationship, the next question is whether there exists a stronger relationship between the two families. Both families use only logic conjunction when describing objects. A consideration of other logic operators will increase the expressive power of descriptions and accordingly, make more sets definable. Specifically, we may consider both logic disjunction and conjunction. As a result, a union of an arbitrary number of sets in $O\!B/E_A$ (or ${\rm CDEF}_A(T)$) will also be definable. The union can be described by a disjunction of the descriptions of the sets from $O\!B/E_A$ (or ${\rm CDEF}_A(T)$). This idea actually constructs two Boolean algebras based on $O\!B/E_A$ and ${\rm CDEF}_A(T)$, respectively. By taking union of an arbitrary number of equivalence classes in $O\!B/E_A$, one can form an atomic Boolean algebra $B(O\!B/E_A)$ with $O\!B/E_A$ as the set of atoms~\cite{yao2015two}:
\begin{equation}
B(O\!B/E_A)=\{\bigcup Z\mid Z \subseteq O\!B/E_A\}.
\end{equation}
Similarly, by taking union of an arbitrary number of conjunctively definable sets in ${\rm CDEF}_A(T)$, one can form the following Boolean algebra:
\begin{equation}
B({\rm CDEF}_A(T))=\{\bigcup Z\mid Z \subseteq {\rm CDEF}_A(T)\}.
\end{equation}
Although the two Boolean algebras are constructed from two different families $O\!B/E_A$ and ${\rm CDEF}_A(T)$ induced by two different formulations, they are actually mathematically equivalent, which has been discussed by Yao in~\cite{yao2015two}.

\begin{theorem}
\label{theo:equivalent_Boolean_algebras}
In a complete table $T=(O\!B,AT,V,F)$, for a set of attributes $A \subseteq AT$, the two Boolean algebras induced by the partition $O\!B/E_A$ and the family of conjunctively definable sets ${\rm CDEF}_A(T)$ are mathematically equivalent~\cite{yao2015two}, that is,
\begin{equation}
\label{equa:equivalent_Boolean_algebras}
B(O\!B/E_A)=B({\rm CDEF}_A(T)).
\end{equation}
\end{theorem}

The proof of Theorem \ref{theo:equivalent_Boolean_algebras} can be found in \ref{appendix:equivalent_Boolean_algebras}. By considering both logic disjunction and conjunction, the two Boolean algebras introduce the same family of definable sets ${\rm DEF}_A(T)=B(O\!B/E_A)=B({\rm CDEF}_A(T))$ which can be used to give a more general definition of structured approximations.

\begin{definition}
\label{def:three_regions_complete_general}
Given a complete table $T=(O\!B,AT,V,F)$ and a set of attributes $A \subseteq AT$, the positive and negative regions regarding a given class $X\subseteq O\!B$ are defined as:
\begin{eqnarray}
\label{equa:three_regions_complete_computational}
{\rm POS}_A(X)&=&\{Y\in {\rm DEF}_A(T)\mid Y \ne \emptyset, Y\subseteq X\},\nonumber\\
{\rm NEG}_A(X)&=&\{Y\in {\rm DEF}_A(T)\mid Y \ne \emptyset, Y\subseteq X^c\}.
\end{eqnarray}
\end{definition}
The decision rules can be accordingly induced by using the description of each set $Y$ in a specific region.

\begin{example}
\label{example:complete_equivalence}

We illustrate the equivalence of the two Boolean algebras and the general definition of structured regions with Table~\ref{tab:example_complete_table}. Table \ref{tab:example_complete_relationships} extends Table \ref{tab:example_complete_conceptual_sets} by adding the equivalence classes in $O\!B/E_{AT}$ that are equivalent with the corresponding conjunctively definable sets.

\begin{table}[ht!]
\centering
\caption{Relationships between conjunctively definable sets and equivalence classes for Table \ref{tab:example_complete_table}}
\label{tab:example_complete_relationships}
\renewcommand{\arraystretch}{1.2}
\scalebox{0.68}{
\begin{tabular}{|l|l|l||l|l|l|}
    \hline
    \multirow{2}{*}{Formulas in ${\rm CDL}_{AT}$} & Meaning sets  & Equivalence classes & \multirow{2}{*}{Formulas in ${\rm CDL}_{AT}$} & Meaning sets & Equivalence classes \\
    & in ${\rm CDEF}_{AT}(T)$ & in $O\!B/E_{AT}$ & & in ${\rm CDEF}_{AT}(T)$ & in $O\!B/E_{AT}$ \\
    \hline
$p_1=\langle a_1,0\rangle$ & $\{x_3,x_4,x_5\}$&&
$p_{14}=\langle a_1,1\rangle\wedge \langle a_3,1\rangle$&$\{x_6\}$&$[x_6]_{AT}$\\
$p_2=\langle a_1,1\rangle$ &$\{x_1,x_2,x_6\}$& &
$p_{15}=\langle a_2,1\rangle\wedge \langle a_3,3\rangle$ &$\emptyset$&\\
$p_3=\langle a_2,1\rangle$ &$\{x_3,x_6\}$&&
$p_{16}=\langle a_2,1\rangle\wedge \langle a_3,1\rangle $ &$\{x_3,x_6\}$&\\
$p_4=\langle a_2,2\rangle$ &$\{x_1,x_2,x_4,x_5\}$&&
$p_{17}=\langle a_2,2\rangle\wedge \langle a_3,3\rangle$&$\{x_1,x_2\}$&$[x_1]_{AT}{=[x_2]_{AT}}$\\
$p_5=\langle a_3,3\rangle$&$\{x_1,x_2\}$&$[x_1]_{AT}{=[x_2]_{AT}}$ &
$p_{18}=\langle a_2,2\rangle\wedge \langle a_3,1\rangle$&$\{x_4,x_5\}$&$[x_4]_{AT}{=[x_5]_{AT}}$\\
$p_6=\langle a_3,1\rangle$ &$\{x_3,x_4,x_5,x_6\}$& &
$p_{19}=\langle a_1,0\rangle\wedge \langle a_2,1\rangle\wedge \langle a_3,3\rangle$&$\emptyset$&\\
$p_7=\langle a_1,0\rangle\wedge \langle a_2,1\rangle$&$\{x_3\}$&$[x_3]_{AT}$&
$p_{20}=\langle a_1,0\rangle\wedge \langle a_2,1\rangle\wedge \langle a_3,1\rangle$&$\{x_3\}$& $[x_3]_{AT}$\\
$p_8=\langle a_1,0\rangle\wedge \langle a_2,2\rangle$ &$\{x_4,x_5\}$&$[x_4]_{AT}{=[x_5]_{AT}}$&
$p_{21}=\langle a_1,0\rangle\wedge \langle a_2,2\rangle\wedge \langle a_3,3\rangle$ &$\emptyset$&\\
$p_9=\langle a_1,0\rangle\wedge \langle a_3,3\rangle$  &$\emptyset$&&
$p_{22}=\langle a_1,0\rangle\wedge \langle a_2,2\rangle\wedge \langle a_3,1\rangle$& $\{x_4,x_5\}$& $[x_4]_{AT}{=[x_5]_{AT}}$\\
$p_{10}=\langle a_1,0\rangle\wedge \langle a_3,1\rangle$&$\{x_3,x_4,x_5\}$&&
$p_{23}=\langle a_1,1\rangle\wedge \langle a_2,1\rangle\wedge \langle a_3,3\rangle$&$\emptyset$&\\
$p_{11}=\langle a_1,1\rangle\wedge \langle a_2,1\rangle$&$\{x_6\}$&$[x_6]_{AT}$&
$p_{24}=\langle a_1,1\rangle\wedge \langle a_2,1\rangle\wedge \langle a_3,1\rangle$&$\{x_6\}$&$[x_6]_{AT}$ \\
$p_{12}=\langle a_1,1\rangle\wedge \langle a_2,2\rangle$&$\{x_1,x_2\}$&$ [x_1]_{AT}{=[x_2]_{AT}}$&
$p_{25}=\langle a_1,1\rangle\wedge \langle a_2,2\rangle\wedge \langle a_3,3\rangle $& $\{x_1,x_2\}$&$[x_1]_{AT}{=[x_2]_{AT}}$\\
$p_{13}=\langle a_1,1\rangle\wedge \langle a_3,3\rangle$&$\{x_1,x_2\}$& $[x_1]_{AT}{=[x_2]_{AT}}$&
$p_{26}=\langle a_1,1\rangle\wedge \langle a_2,2\rangle\wedge \langle a_3,1\rangle $&$\emptyset$&
\\
\hline
\end{tabular}
}
\end{table}

Figure~\ref{fig:example_complete_Boolean_algebras} shows a hierachical structure of the two equivalent Boolean algebras $B(O\!B/E_{AT})$ and $B({\rm CDEF}_{AT}(T))$ built from the computational and conceptual formulations. In the computational formulation, the Boolean algebra is built by taking unions of equivalence classes in $O\!B/E_{AT}=\{\{x_1,x_2\},\{x_3\},\{x_4,x_5\},\{x_6\}\}$ which are given in the lowest nonempty level. The hierarchy includes all equivalence classes given in Equation (\ref{equa:example_complete_all_equivalence_classes}). This indicates that any equivalent class $[x]_A$ induced by $A\subseteq AT$ can be expressed as a union of some equivalent classes in $O\!B/E_{AT}$ following a path in Figure~\ref{fig:example_complete_Boolean_algebras}. However, in the reverse direction, a union of some equivalent classes in $O\!B/E_{AT}$ may not be an equivalent class that can be induced by a set $A\subseteq AT$, such as the set $\{x_1,x_2,x_3\}$ in Figure~\ref{fig:example_complete_Boolean_algebras}. Nevertheless, the set $\{x_1,x_2,x_3\}$ is definable and it can be described by a disjunction of the descriptions of the two equivalence classes $\{x_1,x_2\}, \{x_3\} \in O\!B/E_{AT}$, that is, $(\langle a_1,1\rangle\wedge\langle a_2,2\rangle\wedge\langle a_3,3\rangle)\vee (\langle a_1,0\rangle\wedge\langle a_2,1\rangle\wedge\langle a_3,1\rangle)$.

\begin{figure}[!ht]
\centering
\scalebox{0.9}{
\begin{tikzpicture}[auto, >=stealth', node distance=1em]
\node  (OB) {$O\!B$};

\node [below left = 4em and 7em of OB] (12345) {$\{x_1,x_2,x_3,x_4,x_5\}$};
\node [right = 1em of 12345] (1236) {$\{x_1,x_2,x_3,x_6\}$};
\node [right = 1em of 1236] (12456) {$\{x_1,x_2,x_4,x_5,x_6\}$};
\node [right = 1em of 12456] (3456) {$\{x_3,x_4,x_5,x_6\}$};

\node [below left = 4em and -3em of 12345] (123) {$\{x_1,x_2,x_3\}$};
\node [right = 1em of 123] (1245) {$\{x_1,x_2,x_4,x_5\}$};
\node [right = 1em of 1245] (126) {$\{x_1,x_2,x_6\}$};
\node [right = 1em of 126] (345) {$\{x_3,x_4,x_5\}$};
\node [right = 1em of 345] (36) {$\{x_3,x_6\}$};
\node [right = 1em of 36] (456) {$\{x_4,x_5,x_6\}$};

\node [below right = 4em and 4em of 123] (12) {$\{x_1,x_2\}$};
\node [right = 2em of 12] (3) {$\{x_3\}$};
\node [right = 2em of 3] (45) {$\{x_4,x_5\}$};
\node [right = 2em of 45] (6) {$\{x_6\}$};

\node [below = 21em of OB] (emptyset) {$\emptyset$};

\path (OB.south) edge (12345.north)
				   edge (1236.north)
				   edge (12456.north)
				   edge (3456.north)
		 (12345.south) edge (123.north)
		 			edge (1245.north)
					edge (345.north)
		(1236.south) edge (123.north)
				    edge (126.north)
				    edge (36.north)
		(12456.south) edge (1245.north)
					edge (126.north)
					edge (456.north)
		(3456.south) edge (345.north)
				    edge (36.north)
				    edge (456.north)
		(123.south) edge (12.north)
				  edge (3.north)
		(1245.south) edge (12.north)
				   edge (45.north)
		(126.south) edge (12.north)
				  edge (6.north)
		(345.south) edge (3.north)
				  edge (45.north)
		(36.south) edge (3.north)
				edge (6.north)
		(456.south) edge (45.north)
				edge (6.north)
		(emptyset.north) edge (12.south)
					 edge(3.south)
					 edge (45.south)
					 edge (6.south);
\end{tikzpicture}
}
\caption{The Boolean algebras $B(O\!B/E_{AT})$ and $B({\rm CDEF}_{AT}(T))$}
\label{fig:example_complete_Boolean_algebras}
\end{figure}
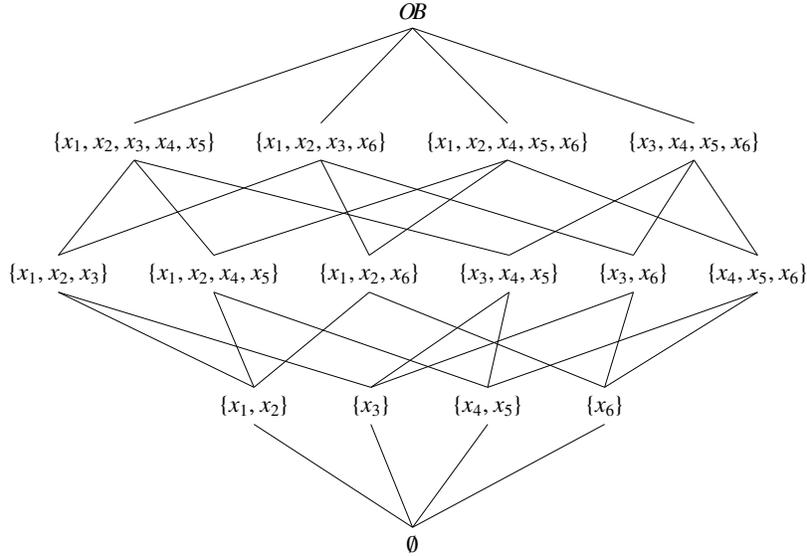

From the view of conceptual formulation, the Boolean algebra $B({\rm CDEF}_{AT}(T))$ is built from conjunctively definable sets in ${\rm CDEF}_{AT}(T)$ given in Equation (\ref{equa:example_complete_CDEF}) which spread in different levels in Figure~\ref{fig:example_complete_Boolean_algebras}. A union of some conjunctively definable sets may not be conjunctively definable, but it can be described by a disjunction of the corresponding conjunctive formulas. For example, the set $\{x_1,x_2,x_3\}$ is a union of $\{x_1,x_2\}$ and $\{x_3\}$, and it could be described by a disjunction of any conjunctive formula describing $\{x_1,x_2\}$ and any conjunctive formula describing $\{x_3\}$, such as $p_5 \vee p_7=\langle a_3,3\rangle\vee(\langle a_1,0\rangle \wedge \langle a_2,1\rangle)$ and $p_{20} \vee p_{25}=(\langle a_1,0\rangle\wedge\langle a_2,1\rangle\wedge\langle a_3,1\rangle) \vee (\langle a_1,1\rangle\wedge\langle a_2,2\rangle\wedge\langle a_3,3\rangle)$.

The two Boolean algebras give two different perspectives of interpreting and constructing the family of definable sets:
\begin{eqnarray}
{\rm DEF}_{AT}(T) &=& \big\{ \, \emptyset, \{x_1,x_2\}, \{x_3\}, \{x_4,x_5\}, \{x_6\}, \{x_1,x_2,x_3\}, \{x_1,x_2,x_4,x_5\}, \nonumber\\
&&~~\{x_1,x_2,x_6\}, \{x_3,x_4,x_5\}, \{x_3,x_6\}, \{x_4,x_5,x_6\}, \{x_1,x_2,x_3,x_4,x_5\}, \nonumber\\
&&~~\{x_1,x_2,x_3,x_6\}, \{x_1,x_2,x_4,x_5,x_6\}, \{x_3,x_4,x_5,x_6\}, O\!B \,\big\}.
\end{eqnarray}
This family can be used to construct the positive and negative regions of a given class $X = \{x_1,x_2,x_3,x_4\}$ as follows:
\begin{eqnarray}
{\rm POS}_{AT}(X)&=&\big\{ \{x_1,x_2\}, \{x_3\}, \{x_1,x_2,x_3\}\big\},\nonumber\\
{\rm NEG}_{AT}(X)&=&\big\{ \{x_6\} \big\}.
\end{eqnarray}
The definable set $\{x_1,x_2,x_3\}$ is not included in the positive regions constructed in Examples \ref{example:complete_computational} and \ref{example:complete_conceptual} of the computational and conceptual formulations. Accordingly, the descriptions of $\{x_1,x_2,x_3\}$ lead to new acceptance rules. For example, we may get the following two new acceptance rules by using the two descriptions of $\{x_1,x_2,x_3\}$ discussed above:
\begin{eqnarray*}
&&\langle a_3,3\rangle\vee(\langle a_1,0\rangle \wedge \langle a_2,1\rangle) \longrightarrow_{\rm A} X, \\
&&(\langle a_1,0\rangle\wedge\langle a_2,1\rangle\wedge\langle a_3,1\rangle) \vee (\langle a_1,1\rangle\wedge\langle a_2,2\rangle\wedge\langle a_3,3\rangle) \longrightarrow_{\rm A} X.
\end{eqnarray*}
By breaking their left-hand-sides at disjunction, these two rules can be equivalently expressed by the following four rules:
\begin{eqnarray*}
&&\langle a_3,3\rangle \longrightarrow_{\rm A} X, \\
&&\langle a_1,0\rangle \wedge \langle a_2,1\rangle \longrightarrow_{\rm A} X, \\
&&\langle a_1,0\rangle\wedge\langle a_2,1\rangle\wedge\langle a_3,1\rangle \longrightarrow_{\rm A} X, \\
&&\langle a_1,1\rangle\wedge\langle a_2,2\rangle\wedge\langle a_3,3\rangle \longrightarrow_{\rm A} X,
\end{eqnarray*}
which already exist in the rules derived in Examples \ref{example:complete_computational} and \ref{example:complete_conceptual}. This illustrates our assumption at the beginning of Section \ref{sec:complete_conceptual} that it is sufficient to consider only logic conjunction from the perspective of rule learning~\cite{hu2018structured}.
\end{example}

\subsection{Three-way decision based on approximations of description}

For the purpose of deriving and interpreting classification rules, we apply the idea of three-way approximations with descriptions and propose the concept of description regions as approximations of the actual description of a given class. Instead of using definable sets of objects to approximate the given class of objects, we use the set of all available descriptions to approximate the actual description.

\begin{definition}
\label{def:three_regions_complete_descriptions}
Given a complete table $T=(O\!B,AT,V,F)$, and a description space ${\rm DES}$ that includes all available descriptions, the positive and negative description regions regarding a given class $X$ are defined as:
\begin{eqnarray}
\label{equa:three_regions_complete_descriptions}
{\rm DPOS}(X)&=&\{p \in {\rm DES} \mid \, \exists\, Y\in {\rm DEF}(T) \, (Y \ne \emptyset \wedge Y \subseteq X \wedge {\rm Des}(Y) = p)\}, \nonumber\\
{\rm DNEG}(X)&=&\{p \in {\rm DES} \mid \, \exists\, Y\in {\rm DEF}(T) \, (Y \ne \emptyset \wedge Y \subseteq X^c \wedge {\rm Des}(Y) = p)\},
\end{eqnarray}
where ${\rm Des}(Y)$ is a description of a definable set $Y$.
\end{definition}

The description space ${\rm DES}$ depends on the set $A \subseteq AT$, the domains of attributes in $A$, and the computational or conceptual formulation under consideration. It is independent of specific values taken by objects on $A$. In the computational formulation, $\rm DES$ is the set of all possible descriptions of equivalence classes. More specifically, it is the set of all conjunctive formulas where each attribute in $A$ appears exactly once, though some conjunctive formulas may not actually describe an existing equivalence class in a specific information table. In terms of the conceptual formulation with a consideration of $A\subseteq AT$, ${\rm DES}$ is the description language ${\rm CDL}_A$. The boundary description region is implicitly defined as the complement of the union of ${\rm DPOS}(X)$ and ${\rm DNEG}(X)$, that is:
\begin{eqnarray}
\label{equa:implicit_boundary_description_region}
{\rm DBND}(X) &=& ({\rm DPOS}(X) \cup {\rm DNEG}(X))^c\nonumber\\
&=& {\rm DES} - {\rm DPOS}(X) - {\rm DNEG}(X).
\end{eqnarray}
The three types of decision rules can be easily induced from the description regions.

\begin{definition}
Given a set of objects $X \subseteq O\!B$, the acceptance, rejection, and non-commitment decision rules regarding $X$ are defined as: for any $y\in O\!B$,
\begin{eqnarray*}
{\rm (A)} && p_a\longrightarrow_{\rm A} X: \textit{~if~} y\models p_a \textit{~for~} p_a\in {\rm DPOS}(X), \textit{~then~accept~} y,\\
{\rm (R)} && p_r\longrightarrow_{\rm R} X: \textit{~if~} y\models p_r \textit{~for~} p_r\in {\rm DNEG}(X), \textit{~then~reject~} y,\\
{\rm (N)} && \textit{If~neither~{\rm(A)}~nor~{\rm(R)}~applies,~then~neither~accept~nor~reject~} y.
\end{eqnarray*}
\end{definition}

These three-way decision rules make the same decisions as those given in Definitions~\ref{def:rules_complete_computational} and \ref{def:rules_complete_conceptual}.

\begin{example}
\label{example:complete_rules}
We continue with Example \ref{example:complete_equivalence} to illustrate the construction of description regions and the derivation of rules. Suppose we are given a class $X=\{x_1,x_2,x_3,x_4\}$. We can construct the positive and negative description regions according to Table \ref{tab:example_complete_relationships}. Suppose we consider all attributes in $AT$. Then we get the following positive and negative description regions:
\begin{eqnarray}
{\rm DPOS}(X)&=&\{\langle a_3,3\rangle, \langle a_1,1\rangle\wedge\langle a_2,2\rangle, \langle a_1,1\rangle\wedge\langle a_3,3\rangle, \langle a_2,2\rangle\wedge\langle a_3,3\rangle, \nonumber\\
&& ~\langle a_1,1\rangle\wedge\langle a_2,2\rangle\wedge\langle a_3,3\rangle, \langle a_1,0\rangle\wedge\langle a_2,1\rangle, \langle a_1,0\rangle\wedge\langle a_2,1\rangle\wedge\langle a_3,1\rangle\}, \nonumber\\
{\rm DNEG}(X)&=&\{\langle a_1,1\rangle\wedge\langle a_2,1\rangle, \langle a_1,1\rangle\wedge\langle a_3,1\rangle, \langle a_1,1\rangle\wedge\langle a_2,1\rangle\wedge\langle a_3,1\rangle\}.
\end{eqnarray}
The corresponding decision rules are:
\begin{eqnarray*}
{\rm (A)} &&\langle a_3,3\rangle \longrightarrow_{\rm A} X,\nonumber\\
&& \langle a_1,1\rangle\wedge\langle a_2,2\rangle \longrightarrow_{\rm A} X,\nonumber\\
&& \langle a_1,1\rangle\wedge\langle a_3,3\rangle \longrightarrow_{\rm A} X,\nonumber\\
&& \langle a_2,2\rangle\wedge\langle a_3,3\rangle \longrightarrow_{\rm A} X,\nonumber\\
&& \langle a_1,1\rangle\wedge\langle a_2,2\rangle\wedge\langle a_3,3\rangle \longrightarrow_{\rm A} X, \nonumber\\
&& \langle a_1,0\rangle\wedge\langle a_2,1\rangle \longrightarrow_{\rm A} X, \nonumber\\
&& \langle a_1,0\rangle\wedge\langle a_2,1\rangle\wedge\langle a_3,1\rangle \longrightarrow_{\rm A} X;\nonumber\\
{\rm (R)} && \langle a_1,1\rangle\wedge\langle a_2,1\rangle \longrightarrow_{\rm R} X, \nonumber\\
&& \langle a_1,1\rangle\wedge\langle a_3,1\rangle \longrightarrow_{\rm R} X, \nonumber\\
&& \langle a_1,1\rangle\wedge\langle a_2,1\rangle\wedge\langle a_3,1\rangle \longrightarrow_{\rm R} X; \nonumber\\
{\rm (N)}  && \textit{Otherwise,~neither~accept~nor~reject~} y.
\end{eqnarray*}
which are the same as those derived in Examples \ref{example:complete_computational} and \ref{example:complete_conceptual}.
\end{example}

It is easy to observe that there is redundancy in the approximations and decision rules. A consideration of removing the redundancy leads to the topics of attribute reduction and rule reduction~\cite{hu2018structured,ma2018three,li2016approximate,lang2018related,lang2019related,yang2019sequential,benitez2018bireducts,benitez2018fca,cornelis2010attribute}, which may be a direction of our future work.

\section{Similarity-based three-way decision in an incomplete table - A computational formulation}
\label{sec:computational_incomplete}

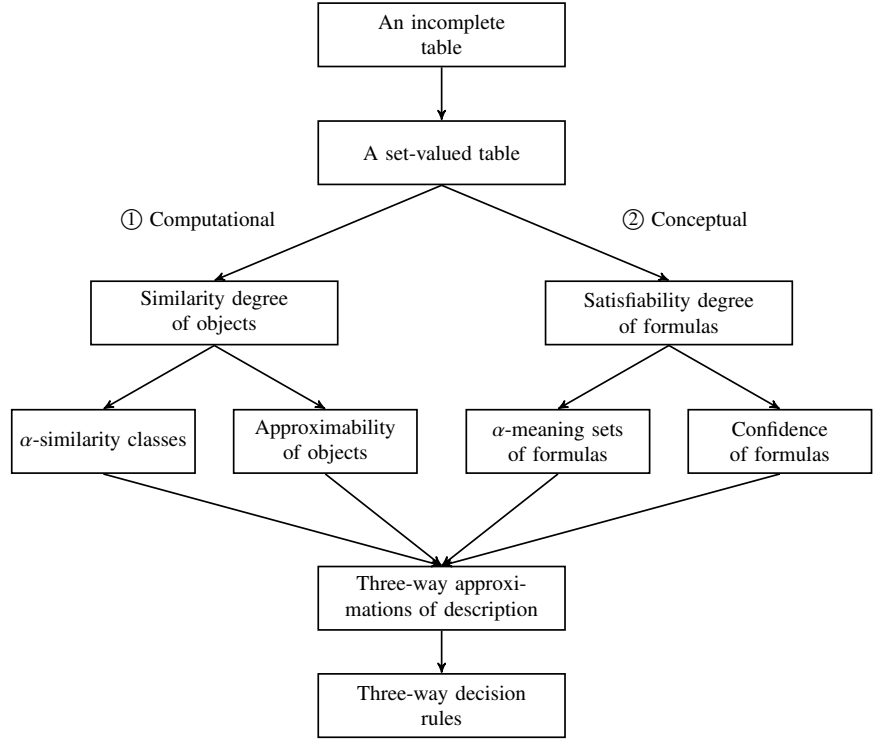
\begin{figure}[!ht]
\centering
\scalebox{0.8}{
\begin{tikzpicture}[auto, >=stealth', node distance=1em,
block_multilines1/.style ={rectangle, draw=black, thick, fill=white, text width=8em, text centered, minimum height=3em},
block_multilines2/.style ={rectangle, draw=black, thick, fill=white,text width=11em, text centered, minimum height=3em},
block_multilines3/.style ={rectangle, draw=white, fill=white, text width=9em, text centered, minimum height=1em},
block_multilines4/.style ={rectangle, dashed, draw=black, thick, fill=white, text width=8em, text centered, minimum height=2em},
noborder_center/.style ={rectangle, draw=white, fill=white, text width=10em,text centered, minimum height=2em}]
\node [block_multilines2](nCOM){An incomplete \\ table};
\node [block_multilines2, below = 2.5em of nCOM](nSET){A set-valued table};
\node [block_multilines2, below left=4.5em and -1em of nSET](nIND){Similarity degree\\ of objects};
\node [block_multilines1, below left=3em and -5em of nIND] (nINDx) {$\alpha$-similarity classes};
\node [block_multilines1, below right=3em and -5em of nIND] (nA) {Approximability of objects};
\node [block_multilines2, below right=4.5em and -1em of nSET](n1){Satisfiability degree\\ of formulas};
\node [block_multilines1, below left=3em and -5em of n1] (n2) {$\alpha$-meaning sets of formulas};
\node [block_multilines1, below right=3em and -5em of n1] (n3) {Confidence of formulas};
\node [block_multilines3, above left=2em and -3.5 of nIND](n5){{\large\textcircled{\small 1}} Computational};
\node [block_multilines3, above right=2em and -3.5 of n1](n6){{\large\textcircled{\small 2}} Conceptual};
\node [block_multilines2, below=18em of nSET](nLUIND) {Three-way approximations of description};
\node [block_multilines2, below=2em of nLUIND] (nRIND) {Three-way decision\\ rules};
\draw [->,thick] (nCOM.south) -- (nSET.north);
\draw [->,thick] (nSET.south) -- (n1.north);
\draw [->,thick] (n1.south) -- (n2.north);
\draw [->,thick] (n1.south) -- (n3.north);
\draw [->,thick] (nSET.south) -- (nIND.north);
\draw [->,thick] (nIND.south) -- (nA.north);
\draw [->,thick] (nA.south) -- (nLUIND.north);
\draw [->,thick] (nIND.south) -- (nINDx.north);
\draw [->,thick] (n2.south) -- (nLUIND.north);
\draw [->,thick] (n3.south) -- (nLUIND.north);
\draw [->,thick] (nINDx.south) -- (nLUIND.north);
\draw [->,thick] (nLUIND.south) -- (nRIND.north);
\end{tikzpicture}}
\caption{The computational and conceptual formulations of three-way decision with incomplete information}
\label{fig:computational_conceptual_incomplete}
\end{figure}

Based on the above review of three-way decision with a complete table, we generalize the results of both computational and conceptual formulations into an incomplete table. Firstly, we equivalently interpret an incomplete table by a set-valued table~\cite{orlowska1998introduction,lipski1979semantic,lipski1981databases}. Secondly, we generalize the results with a complete table into a set-valued table in order to analyze the incomplete information. This leads to a computational formulation and a conceptual formulation with incomplete information as shown in Figure~\ref{fig:computational_conceptual_incomplete}. In this section, we investigate the computational formulation based on a proposed measure of similarity degree of objects.

\subsection{Interpreting an incomplete table by a set-valued table}
\label{sec:incomplete_set-valued}

In a complete table, the value of each object on each attribute is exactly known. In real-world situations, we often have incomplete information about the actual values. In this case, an incomplete table can be used to represent the dataset. There are various formal representations and definitions of an incomplete table proposed in the literature~\cite{stefanowski2001incomplete,lipski1979semantic,guan2012generalized,kryszkiewicz1998rough,lipski1981databases,wang2001rough,grzymala2000comparison,grzymala2005incomplete,grzymala2004characteristic,leung2008dependence} that consider different semantics and categories of incomplete information. In this paper, we consider the following definition proposed in~\cite{luo2019on}.

\begin{definition}
\label{def:incomplete_table}
An incomplete table can be represented by a tuple $IT=(O\!B,AT,V^{\prime},F^{\prime})$, where
\begin{enumerate}[label=(\arabic*)]
    \item $O\!B$ is a finite nonempty set of objects called the universe;
    \item $AT$ is a finite nonempty set of attributes;
\item $V'=\bigcup_{a \in AT} {V'_a}$, where $V'_a=V_a\cup \{*\} \cup \{\dag\}\cup  \{{\triangledown}\}\cup \{{\rm NA}\}$, $V_a$ is the domain of an attribute $a$ and $*$, $\dag$, ${\triangledown}$, and ${\rm NA}$ are four special symbols representing a do-not-care value {\rm(D)}, a partially-known value {\rm(P)}, a class-specific value {\rm(C)}, and a non-applicable value {\rm(N)}, respectively, which will be explained below;
    \item $F^{\prime}=\{f_a^{\prime}:O\!B\longrightarrow V^{\prime}_a \mid a\in AT\}$, where $f_a^{\prime}$ is an information function that maps each object $x \in O\!B$ to a value $f_a^{\prime}(x)\in V_a^\prime$ on an attribute $a$.
\end{enumerate}
The meanings of the four special values $*$, $\dag_a^x$, $\triangledown^x_{(a,b)}$, and ${\rm NA}$ are:
\begin{itemize}
\item[\rm(D)]
``Do-not-care values'' denoted by $*$: If $f_a^{\prime}(x)=*$, the actual value of an object $x$ on an attribute $a$ is missing, but we do not care what the actual value is. In other words, the actual value does not affect the analysis or decision-making on the object. One can replace $*$ by any value $v$ in $V_a$ and obtain the same result. That is, our result is independent of the choice of $v$.

\item [\rm(P)]
``Partially-known values'' denoted by $\dag$: If $f_a^{\prime}(x)=\dag_a^x$, the actual value of an object $x$ on an attribute $a$ is missing, but we know that the actual value must be in a subset $P_a(x)\subseteq V_a$, where $|P_a(x)|>1$.

\item[\rm(C)]
``Class-specific values'' denoted by $\triangledown$: If $f_a^{\prime}(x)=\triangledown^x_{(a,b)}$, the actual value of an object $x$ on an attribute $a$ is missing, but we know that it depends on the value of $x$ on a special attribute $b\in AT$. The actual value must be one of those known values on attribute $a$ taken by the objects that have the same value on attribute $b$ with $x$. Formally, $\triangledown^x_{(a,b)}$ represents a value in the set $V_a^{\triangledown_{b}}(x)=\{f_a^{\prime}(y)\mid f_b^{\prime}(x)=f_b^{\prime}(y) \wedge f_a^\prime(y) \in V_a\}$.

\item [\rm(N)]
``Non-applicable values'' denoted by {\rm NA}: If $f_a^{\prime}(x)={\rm NA}$,
the actual value of an object $x$ on an attribute $a$ does not exist.
A ``non-applicable value'' represents incomplete information since complete information requires a unique value taken from the domain $V_a$. Nevertheless, we consider {\rm NA} as a special known value in the sense that we know the attribute is not applicable to the object, that is, no value in $V_a$ will be taken by the object. Thus, we treat {\rm NA} in the same way as values in $V_a$.
\end{itemize}
\end{definition}

According to Definition \ref{def:incomplete_table}, given an object $x \in O\!B$ and an attribute $a \in AT$, there are five cases for the value $f^\prime_a(x)$, namely, $f^\prime_a(x)\in V_a$, $f^\prime_a(x)=*$, $f^\prime_a(x)=\dag_a^x$, $f^\prime_a(x)=\triangledown^x_{(a,b)}$, and $f^\prime_a(x)={\rm NA}$. In the case of $f^\prime_a\in V_a$, $x$ takes a unique value on the attribute $a$ which is the actual value. Thus, in this case, we have complete information about the actual value of $x$ on $a$. In the other four cases of $f^\prime_a\in \{*,\dag_a^x,\triangledown^x_{(a,b)}, {\rm NA}\}$, we have incomplete information in the sense that we do not know the actual value of $x$ on $a$. The four values $*$, $\dag_a^x$, $\triangledown^x_{(a,b)}$, and NA represent four different interpretations and cases of such incompleteness.

There are many approaches proposed to analyzing an incomplete table~\cite{stefanowski2001incomplete,lipski1979semantic,guan2012generalized,kryszkiewicz1998rough,lipski1981databases,wang2001rough,grzymala2000comparison,grzymala2005incomplete,grzymala2004characteristic,leung2008dependence,luo2019on}. Lipski~\cite{lipski1981databases} presents a possible-world semantics that replaces an incompletely known value on an attribute with a subset of values in the domain. Though we do not know the actual value, it must be one of the values in this subset. Each value in this subset partially represents one possible-world of the reality, that is, the value is taken in this cell in one possibility of the actual table. Based on this possible-world semantics, an incomplete table can be equivalently represented by a set-valued table~\cite{lipski1979semantic,lipski1981databases,guan2006set,luo2019on}.

\begin{definition}
A set-valued table can be presented by a tuple $S\!T=(O\!B,AT,V \cup \{{\rm NA}\},S)$ where
\begin{enumerate}[label=(\arabic*)]
    \item $O\!B$ is a finite nonempty set of objects called the universe;
    \item $AT$ is a finite nonempty set of attributes;
    \item $V=\bigcup_{a \in AT} {V_a}$, where $V_a$ is the domain of an attribute $a$;
    \item $S=\{s_a: O\!B \longrightarrow 2^{V_a \cup \{{\rm NA}\}} -\{\emptyset\} \mid a\in AT\}$, where $s_a$ is a set-valued information function that maps each object $x\in O\!B$ to a nonempty set of values $s_a(x)\subseteq V_a \cup \{{\rm NA}\}$ on an attribute $a \in AT$. $2^{V_a \cup \{{\rm NA}\}}$ denotes the power set of $V_a \cup \{{\rm NA}\}$.
    \end{enumerate}
\end{definition}

An incomplete table can be easily transformed into a set-valued table by considering the five cases of knowingness of the actual values:
\begin{itemize}
\item[\rm(K)]
Known value: if $f_a^{\prime} (x) \in V_a$, then $s_a(x)=\{f_a^{\prime}(x)\}$;
\item[\rm(D)]
Do-not-care value: if $f_a^{\prime} (x)=\ast$, then $s_a(x)=V_a$;
\item[\rm(P)]
Partially-known value: if $f_a^{\prime} (x)=\dag_a^x$, then $s_a(x)=P_a(x)$;
\item[\rm(C)]
Class-specific value: if $f_a^{\prime} (x) =\triangledown^x_{(a,b)}$, then $s_a(x)=V_a^{\triangledown_b}(x)$;
\item[\rm(N)]
Non-applicable value: if $f_a^{\prime} (x)={\rm NA}$, then $s_a(x)=\{{\rm NA}\}$.
\end{itemize}
Based on the above transformation, we study an incomplete table through its equivalent set-valued table.

\subsection{The similarity degree of objects}

In order to generalize the results in computational formulation with a complete table into a set-valued table, we propose a measure of similarity degree between objects which can be considered as a generalization of the equivalence relation. In a complete table, two objects are indiscernible with respect to an attribute $a\in AT$ if they have the same value on $a$. The indiscernibility can also be interpreted in terms of similarity. If two objects are indiscernible, they are considered to be totally similar to each other, that is, exactly the same. In a set-valued table, the actual value of an object on an attribute may not be known. In this case, we measure the similarity degree of two objects regarding a specific attribute by looking at the probability of having the same actual value for them, that is, the probability of the two objects being indiscernible.

\begin{definition}
In a set-valued table $S\!T=(O\!B,AT,V\cup \{{\rm NA}\},S)$, the similarity degree of objects on an attribute $a\in AT$ is given by a mapping $G_a:~O\!B\times O\!B\longrightarrow [0,1]$ defined as:

\begin{equation}
\label{equa:similarity_degree_single_attribute}
G_a(x,y)= \left \{
\begin{array}{lcl}
\dfrac{ \abs{\{(v,v)\,\mid\, v\in s_a(x), v \in s_a(y)\}}} { \abs{\{(v,w)\,\mid\, v\in s_a(x), w\in s_a(y)\}} }, & & x \ne y,\\
&&\\
1, & & x = y,\\
\end{array}
\right.
\end{equation}
where $|\cdot|$ denotes the cardinality of a set.
\end{definition}

The similarity degree for the case $x \ne y$ can be equivalently computed as:
\begin{equation}
G_a(x,y)=\frac{|s_a(x)\cap s_a(y)|}{| s_a(x) \times s_a(y)|},
\end{equation}
where $\times$ denotes the Cartesian product. In the special case where we have complete information, each cell in a set-valued table is a singleton set and $G_a(x,y)$ is either 0 or 1 for any two objects $x$ and $y$. Another special case that should be clarified is when two objects $x$ and $y$ take the same non-singleton set of values on $a$, that is, $s_a(x) = s_a(y)$ and $|s_a(x)| = |s_a(y)| > 1$. Since each value in the set represents a possibility of the actual value, it may happen that $x$ and $y$ take different actual values. In this sense, the similarity degree of $x$ and $y$ is not 1 which is the value indicating indiscernibility or total similarity. Instead, the similarity degree of $x$ and $y$ is the probability that they have the same actual value and are indiscernible, which is $\frac{1}{|s_a(x)|} = \frac{1}{|s_a(y)|}$.

Equation~\eqref{equa:similarity_degree_single_attribute} measures the similarity degree of two objects with respect to a single attribute. In practical situations, we usually consider a set of attributes. In a complete table, two objects $x,y\in O\!B$ are indiscernible regarding a set of attributes $A\subseteq AT$ (i.e. $(x,y)\in E_A$), if and only if they are indiscernible on each attribute $a\in A$ (i.e. $(x,y)\in E_A\Leftrightarrow \forall\,a\in A,  f_a(x)=f_a(y)$). A complete table is a special case of an incomplete table. Correspondingly, indiscernibility regarding a set of attributes in a complete table should be a special case of similarity regarding a set of attributes in an incomplete table. Following this idea, in an incomplete table, two objects are similar regarding $A\subseteq AT$ if and only if they are similar on each attribute in $A$. In other words, the similarity degree of two objects regarding $A$ can be expressed as a ``conjunction'' of their similarity degrees regarding each single attribute in $A$. Such a ``conjunction'' can be achieved through the notion of Triangle-norm (T-norm for short)~\cite{stefanowski2001incomplete,hajek1998metamathematics,rosen2011discrete,cornelis2014multi}, which is a generalization of the classical logic conjunction and is also referred to as the fuzzy logic conjunction. Formally, a T-norm is a mapping T $:[0,1]^2\rightarrow[0,1]$ which satisfies the following conditions: for any real number $u_i \in [0,1]$,
\begin{enumerate}[label=(\arabic*)]
\item Commutativity: ${\rm T}(u_1,u_2)={\rm T}(u_2,u_1)$,
\item Associativity: ${\rm T}(u_1,{\rm T}(u_2,u_3))={\rm T}({\rm T}(u_1,u_2),u_3)$,
\item Monotonicity: ${\rm T}(u_1,u_2)\leq{\rm T}(u_1,u_3)$, if $u_2\leq u_3$,
\item Boundary condition: ${\rm T}(u_1,1)=u_1$.
\end{enumerate}
According to the associativity, Wang and Li~\cite{wang1996fuzzy} generalize the above T-norm into
$n$-dimensional T-norm which is a mapping T: $[0,1]^n\longrightarrow [0,1]$. Specifically, ${\rm T}(u_1,u_2,\cdots,u_n)$ is recursively calculated as:
\begin{eqnarray}
  {\rm T}(u_1,u_2,\cdots,u_n)&=&{\rm T}(u_1,{\rm T}(u_2,u_3,\cdots,u_n)),\nonumber\\
  {\rm T}(u_2,u_3,\cdots,u_n)&=&{\rm T}(u_2,{\rm T}(u_3,u_4,\cdots,u_n)),\nonumber\\
  &\vdots&\nonumber\\
  {\rm T}(u_{n-2},u_{n-1},u_n)&=&{\rm T}(u_{n-2},{\rm T}(u_{n-1},u_n)).
\end{eqnarray}
Based on the $n$-dimensional T-norm, we propose the following definition of similarity degree regarding a set of attributes.

\begin{definition}
\label{def:similarity_degree_attribute_set}
In a set-valued table $S\!T=(O\!B,AT,V\cup \{{\rm NA}\},S)$, the similarity degree of objects on attributes $A=\{a_1,a_2,\cdots,a_m\}\subseteq AT$ is given by a mapping $G_A$: $O\!B\times O\!B\longrightarrow [0,1]$ defined as:
\begin{equation}
\label{equa:similarity_degree_attribute_set}
G_A(x,y)= {\rm T}(G_{a_1}(x,y),G_{a_2}(x,y),\cdots,G_{a_m}(x,y))= {\rm T}_{1 \le i \le m}G_{a_i}(x,y).
\end{equation}
\end{definition}

According to the boundary condition for T-norm and the fact $G_a(x,x) = 1$ for any attribute $a$, the similarity degree $G_A(x,x)$ is always 1 for any set $A$ and any mapping T. One can simply apply a specific definition of the T-norm in Equation~\eqref{equa:similarity_degree_attribute_set} to calculate $G_A(x,y)$. In this paper, we consider two popular T-norm~\cite{menger1942statistical,bellman1973analytic,dubois1980fuzzy,giles1976lukasiewicz,dubois1982class}:
\begin{enumerate}[label=(\arabic*)]
  \item The standard minimum T-norm: ${\rm T}^{\min}(x,y)=\min(x,y)$;
  \item The algebraic-product T-norm: ${\rm T}^{\prod}(x,y)=x\cdot y$.
\end{enumerate}
Based on these two T-norms, we get the corresponding two similarity degrees regarding attributes $A\subseteq AT$ as follows:
\begin{enumerate}[label=(\arabic*)]
\item Minimum similarity degree: $G_A^{\min}(x,y)={\min}_{a\in A} G_a(x,y)$;
\item Algebraic-product similarity degree: $G_A^{\prod}(x,y)=\prod_{a\in A}{G_a(x,y)}$.
\end{enumerate}

For simplicity, in the following discussion, we will use a matrix to represent the similarity degree $G_A$ as follows:
\begin{equation}
M_{n\times n}=
\bordermatrix{
       &x_1          & x_2          & \cdots &x_n          \cr
 x_1   &G_A(x_1,x_1) & G_A(x_1,x_2) & \cdots & G_A(x_1,x_n) \cr
 x_2   &G_A(x_2,x_1) & G_A(x_2,x_2) & \cdots & G_A(x_2,x_n) \cr
\vdots &\vdots       & \vdots       & \ddots & \vdots       \cr
 x_n   &G_A(x_n,x_1) & G_A(x_n,x_2) & \cdots & G_A(x_n,x_n)
}
\end{equation}
where $n = |O\!B|$. Moreover, we will use a subscript $A$ to generally represent both the case of multiple attributes and the case of a single attribute with $A$ as a singleton set.

\begin{example}
\label{example:incomplete_computational_similarity_degree}

We illustrate the computation of the similarity degree with an incomplete table $IT$ given in Table~\ref{tab:example_incomplete_tables}(a). Specifically, we have $O\!B=\{{x_1},{x_2},{x_3},{x_4},{x_5},{x_6},{x_7},{x_8}\}$ and $AT=\{{a_1},{a_2},{a_3}\}$. The domains of the three attributes are $V_{a_1}=\{0,1\}$, $V_{a_2}=\{1,2,3\}$, and $V_{a_3}=\{0,1,3\}$. According to our discussion in Section~\ref{sec:incomplete_set-valued}, $IT$ can be equivalently interpreted by a set-valued table $S\!T$ given in Table~\ref{tab:example_incomplete_tables}(b) where we simply use a value to represent a singleton set for a known value. The value range of $\dag_{a_3}^{x_6}$ is $\{1,3\}$. To find the set of values for replacing $\triangledown^{x_2}_{(a_2,a_3)}$, we look for objects that have the same actual value on attribute $a_3$ with object $x_2$, that is, the two objects $x_1$ and $x_3$. They take values 2 and 1, respectively, on attribute $a_2$. Thus, $\triangledown^{x_2}_{(a_2,a_3)}$ is replaced by \{1,2\}.

\begin{table}[!ht]
\centering
\caption{An incomplete table $IT$ and its equivalent set-valued table $S\!T$}
\label{tab:example_incomplete_tables}
\begin{subtable}{.45\linewidth}
\centering
\caption{$IT$}
\setlength{\tabcolsep}{1em}
\renewcommand{\arraystretch}{1.2}
\begin{tabular}{c|ccc}
\hline
           & $a_1$& $a_2$               & $a_3$  \\ \hline
  $x_1$    &1     &2                    &3 \\
  $x_2$    &1     &$\triangledown^{x_2}_{(a_2,a_3)}$ &3 \\
  $x_3$    &1     &1                    &3 \\
  $x_4$    &0     &3                    &$\ast$ \\
  $x_5$    &$\ast$&3                    &1 \\
  $x_6$    &$\ast$&3                    &$\dag_{a_3}^{x_6}$ \\
  $x_7$    &NA    &1                    &0 \\
  $x_8$    &NA    &$\ast$               &0\\
\hline
\end{tabular}
\end{subtable}
~
\begin{subtable}{.45\linewidth}
\centering
\caption{$S\!T$}
\setlength{\tabcolsep}{1em}
\renewcommand{\arraystretch}{1.2}
\begin{tabular}{c|ccc}
\hline
           & $a_1$   & $a_2$         & $a_3$  \\ \hline
  $x_1$    &1        &2              &3 \\
  $x_2$    &1        &$\{1,2\}$      &3 \\
  $x_3$    &1        &1              &3 \\
  $x_4$    &0        &3              &$\{0,1,3\}$ \\
  $x_5$    &$\{0,1\}$&3              &1 \\
  $x_6$    &$\{0,1\}$&3              &$\{1,3\}$ \\
  $x_7$    &NA       &1              &0 \\
  $x_8$    &NA       &$\{1,2,3\}$    &0\\
\hline
\end{tabular}
\end{subtable}
\end{table}

When considering all the three attributes in $AT$, we get the minimum similarity degree:
\begin{equation}
\label{equa:example_incomplete_Matrix_min}
M_{8\times 8}^{\min}=
\bordermatrix{
          &x_1  &x_2    &x_3   &x_4   &x_5   &x_6   &x_7   &x_8\cr
    x_1   &1    & 0.5   & 0    & 0    & 0    & 0    & 0    & 0   \cr
    x_2   &0.5  & 1     & 0.5  & 0    & 0    & 0    & 0    & 0  \cr
    x_3   &0    & 0.5   & 1    & 0    & 0    & 0    & 0    & 0 \cr
    x_4   &0    & 0     & 0    & 1    & 0.333& 0.333& 0    & 0\cr
    x_5   &0    & 0     & 0    & 0.333& 1    & 0.5  & 0    & 0 \cr
    x_6   &0    & 0     & 0    & 0.333& 0.5  & 1    & 0    & 0 \cr
    x_7   &0    & 0     & 0    & 0    & 0    & 0    & 1    & 0.333 \cr
    x_8   &0    & 0     & 0    & 0    & 0    & 0    & 0.333& 1 }
\end{equation}
and the algebraic-product similarity degree:
\begin{equation}
\label{equa:example_incomplete_Matrix_prod}
M_{8\times 8}^{\prod}=
\bordermatrix{
          &x_1  &x_2    &x_3   &x_4   &x_5   &x_6   &x_7   &x_8   \cr
    x_1   &1    & 0.5   & 0    & 0    & 0    & 0    & 0    & 0    \cr
    x_2   &0.5  & 1     & 0.5  & 0    & 0    & 0    & 0    & 0    \cr
    x_3   &0    & 0.5   & 1    & 0    & 0    & 0    & 0    & 0    \cr
    x_4   &0    & 0     & 0    & 1    & 0.167& 0.167& 0    & 0    \cr
    x_5   &0    & 0     & 0    & 0.167& 1    & 0.25 & 0    & 0    \cr
    x_6   &0    & 0     & 0    & 0.167& 0.25 & 1    & 0    & 0    \cr
    x_7   &0    & 0     & 0    & 0    & 0    & 0    & 1    & 0.333\cr
    x_8   &0    & 0     & 0    & 0    & 0    & 0    & 0.333& 1    }
\end{equation}

We illustrate the calculation of $G_{AT}^{\min}(x_4,x_6)$ and $G_{AT}^{\prod}(x_4,x_6)$. We first calculate their similarity degree regarding each single attribute:
\begin{eqnarray}
G_{a_1}(x_4,x_6) &=& \frac{|\{0\} \cap \{0,1\}|}{|\{0\} \times \{0,1\}|} = \frac{1}{2},\nonumber\\
G_{a_2}(x_4,x_6) &=& \frac{|\{3\} \cap \{3\}|}{|\{3\} \times \{3\}|} = 1,\nonumber\\
G_{a_3}(x_4,x_6) &=& \frac{|\{0,1,3\} \cap \{1,3\}|}{|\{0,1,3\} \times \{1,3\}|} = \frac{1}{3}.
\end{eqnarray}
Then we use the two T-norms to get their similarity degree regarding $AT$:
\begin{eqnarray}
G_{AT}^{\min}(x_4,x_6) &=& \min(\frac{1}{2}, 1, \frac{1}{3}) = \frac{1}{3} \approx 0.333,\nonumber\\
G_{AT}^{\prod}(x_4,x_6) &=& \frac{1}{2} \cdot 1 \cdot \frac{1}{3} = \frac{1}{6} \approx 0.167.
\end{eqnarray}
Similarly, one can calculate the similarity degrees given in Equations (\ref{equa:example_incomplete_Matrix_min}) and (\ref{equa:example_incomplete_Matrix_prod}).
\end{example}

\subsection{Three-way decision based on $\alpha$-similarity classes}

Based on the similarity degree, we propose two methods to learn three-way classification rules respectively in this and the following sections. In this section, we focus on a method based on similarity classes derived by applying a threshold $\alpha\in [0,1]$ on the similarity degree.

\begin{definition}
In a set-valued table $S\!T=(O\!B,AT,V\cup \{{\rm NA}\},S)$, for a set of attributes $A\subseteq AT$ and a threshold $\alpha\in [0,1]$, the $\alpha$-similarity class of an object $x\in O\!B$ is defined as:
\begin{equation}
(x)_{G_A}^\alpha=\{y\in O\!B\mid G_A(x,y)\geq\alpha\}.
\end{equation}
\end{definition}

An object $y$ is considered to be similar to $x$, that is, $y \in (x)_{G_A}^\alpha$, if $y$ is similar to $x$ in at least an $\alpha$-level. Since there may be overlap between two similarity classes, the family of similarity classes $\{(x)_{G_A}^\alpha\mid x\in O\!B\}$ is a covering of $O\!B$. In other words, it is a family of nonempty subset of $O\!B$ whose union is $O\!B$. In this paper, we assume that the threshold $\alpha$ is given. The determination of $\alpha$ may be further investigated in our future work. Intuitively, a threshold $\alpha \ge 0.5$ is desirable. However, in fact, the similarity degree of two objects is more likely to be a small number below 0.5, especially when we have incomplete information for both objects. Thus, a threshold $\alpha < 0.5$ is more practical.

In the computational formulation with a complete table, we use equivalence classes to construct the positive and negative description regions regarding a given class $X\subseteq O\!B$, as given in Equation~\eqref{equa:three_regions_complete_descriptions}. In an incomplete table, the equivalence classes are generalized into similarity classes. In order to use similarity classes to construct the description regions, we need to find the description of a similarity class. In this paper, we adopt the description of an object $x$ in the context of incomplete information proposed by Luo et al.~\cite{luo2019on} and use it as the description of the similarity class $(x)_{G_A}^\alpha$.

\begin{definition}
In a set-valued table $S\!T=(O\!B,AT,V\cup \{{\rm NA}\},S)$, given a set of attributes $A = \{a_1,a_2,...,a_m\} \subseteq AT$, the (conjunctive) description of a similarity class $(x)_{G_A}^\alpha$ for $x\in O\!B$ is defined as:
\begin{equation}
\label{equa:description_CDes}
{\rm CDes}((x)_{G_A}^\alpha)= {\rm CDes}_A(x) = \{\bigwedge_{1 \le i \le m}\langle a_i,v_i\rangle\mid v_i\in s_{a_i}(x)\},
\end{equation}
where ${\rm CDes}_A(x)$ denotes a conjunctive description of $x$ regarding $A$.
\end{definition}

Although ${\rm CDes}((x)_{G_A}^\alpha)$ and ${\rm CDes}_A(x)$ are the same in our definition, their meanings or semantics are different. ${\rm CDes}((x)_{G_A}^\alpha)$ denotes the description of a similarity class and ${\rm CDes}_A(x)$ denotes the description of an object. These two notations are interchangeably used in this paper according to different contexts. One can easily verify that ${\rm CDes}((x)_{G_A}^\alpha) \subseteq {\rm CDL}_A$. Intuitively, ${\rm CDes}((x)_{G_A}^\alpha)$ is the set of conjunctive formulas in ${\rm CDL}_A$ that satisfy the following two conditions:
\begin{enumerate}[label=(\arabic*)]
\item Each attribute in $A$ appears exactly once in the formula.
\item The formula is possibly satisfied by $x$ once the information is complete.
\end{enumerate}
By using the conjunctive descriptions of {\rm similarity classes}, we can generalize the description regions defined in Definition~\ref{def:three_regions_complete_descriptions} into a set-valued table.
\begin{definition}
\label{def:incomplete_approximation_1}
In a set-valued table $S\!T=(O\!B,AT,V\cup \{{\rm NA}\},S)$, given a set of attributes $A\subseteq AT$ and a threshold $\alpha \in [0,1]$, the positive and negative description regions regarding a given class $X \subseteq O\!B$ are defined as:
\begin{eqnarray}
\label{equa:incomplete_approximation_1}
{\rm DPOS}_{(A,{\alpha})}^1(X)&=&\bigcup \{ {\rm CDes}((x)_{G_A}^\alpha) \mid (x)_{G_A}^\alpha \subseteq X\},\nonumber\\
{\rm DNEG}_{(A,{\alpha})}^1(X)&=&\bigcup \{ {\rm CDes}((x)_{G_A}^\alpha) \mid (x)_{G_A}^\alpha \subseteq X^c\}.
\end{eqnarray}
\end{definition}

Since the similarity degree $G_A(x,x)$ is always 1, we have $x \in  (x)_{G_A}^\alpha$ for any threshold $\alpha$. Thus, we omit the condition $ (x)_{G_A}^\alpha \ne \emptyset$ in Equation (\ref{equa:incomplete_approximation_1}). An object may be described by more than one conjunctive formula, that is, $|{\rm CDes}((x)_{G_A}^\alpha)|$ may be greater than 1. On the other hand, one conjunctive formula may be used to describe more than one similarity class, that is, ${\rm CDes}((x)_{G_A}^\alpha) \cap {\rm CDes}((x^\prime)_{G_A}^\alpha)$ may not be empty for $x\neq x^\prime$. This indicates that a particular conjunctive formula may be included in both ${\rm DPOS}_{(A,{\alpha})}^1(X)$ and ${\rm DNEG}_{(A,{\alpha})}^1(X)$. Due to such conflict, we derive a non-commitment rule with this particular formula.

\begin{definition}
\label{def:incomplete_rule_1}
In a set-valued table $S\!T=(O\!B,AT,V\cup \{{\rm NA}\},S)$, for a set of attributes $A\subseteq AT$, we get the acceptance, rejection, and non-commitment rules as follows: for any $y\in O\!B$,
\begin{eqnarray*}
{\rm (A)} && p_a\longrightarrow_{\rm A} X: \textit{~if~} y\models p_a \textit{~for~} p_a\in {\rm DPOS}_{(A,{\alpha})}^1(X) - {\rm DNEG}_{(A,{\alpha})}^1(X), \textit{~then~accept~} y,\\
{\rm (R)} && p_r\longrightarrow_{\rm R} X: \textit{~if~} y\models p_r \textit{~for~} p_r\in {\rm DNEG}_{(A,{\alpha})}^1(X) - {\rm DPOS}_{(A,{\alpha})}^1(X), \textit{~then~reject~} y,\\
{\rm (N)} && \textit{If~neither~{\rm(A)}~nor~{\rm(R)}~applies,~then~neither~accept~nor~reject~} y.
\end{eqnarray*}
\end{definition}

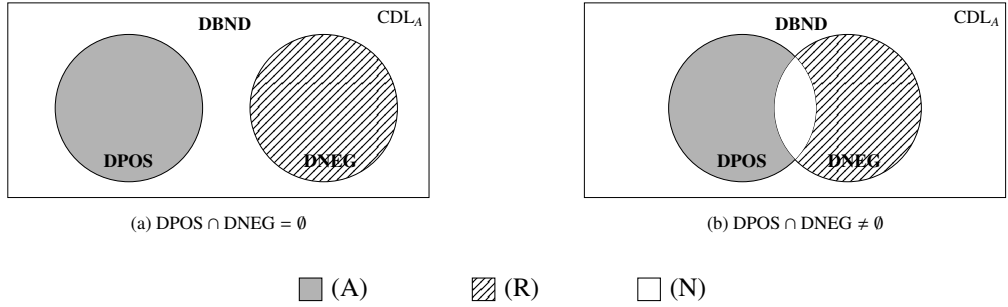
\begin{figure}[!ht]
\begin{center}
\begin{subfigure}[b]{0.45\textwidth}
\centering
\scalebox{0.7}{
\begin{tikzpicture}
\centering
\draw (0,3.7) rectangle (8,0);
\draw[pattern=north east lines] (6,1.7) circle (1.4);
\draw[fill=black!30] (2.3,1.7) circle (1.4);

\node at (3.5,3.3) [right] {\textbf{DBND}};
\node at (5.5,0.7) [right] {\textbf{DNEG}};
\node at (1.7,0.7) [right] {\textbf{DPOS}};
\node at (6.9,3.4) [right] {${\rm CDL}_A$};
\end{tikzpicture}
}
\caption{${\rm DPOS} \cap {\rm DNEG} = \emptyset$}
\end{subfigure}
~
\begin{subfigure}[b]{0.45\textwidth}
\centering
\scalebox{0.7}{
\begin{tikzpicture}
\centering
\draw (0,3.7) rectangle (8,0);
\draw[pattern=north east lines] (5,1.7) circle (1.4);
\draw[fill=black!30] (3,1.7) circle (1.4);

\node at (3.5,3.3) [right] {\textbf{DBND}};
\node at (4.5,0.7) [right] {\textbf{DNEG}};
\node at (2.4,0.7) [right] {\textbf{DPOS}};
\node at (6.9,3.4) [right] {${\rm CDL}_A$};

\begin{scope}
	\clip (5,1.7) circle (1.4);
	\clip (3,1.7) circle (1.4);
	\draw[fill=white] (5,1.7) circle (1.4);
\end{scope}
\end{tikzpicture}
}
\caption{${\rm DPOS} \cap {\rm DNEG} \ne \emptyset$}
\end{subfigure}
\par\bigskip
\begin{subfigure}[b]{1.0\textwidth}
\centering
\begin{tikzpicture}
\def\myscale{0.1}
\def\startxpos{0.4*10.7*\myscale + 1}
\def\starty{0.4*2.35*\myscale}
\draw[fill=black!30] (\startxpos,\starty) rectangle (\startxpos+0.3,\starty+0.3);
\node at (\startxpos+0.3, \starty+0.15) [right] {(A)};
\draw[pattern=north east lines] (\startxpos + 2.3,\starty) rectangle (\startxpos+2.6,\starty+0.3);
\node at (\startxpos+2.6, \starty+0.15) [right] {(R)};
\draw[] (\startxpos + 4.5,\starty) rectangle (\startxpos+4.8,\starty+0.3);
\node at (\startxpos+4.8, \starty+0.15) [right] {(N)};

\end{tikzpicture}
\end{subfigure}

\end{center}
\caption{Two possibilities of relationships among the three description regions and the corresponding three-way decision rules}
\label{fig:two_possibilities}
\end{figure}

The boundary description region is implicitly defined by:
\begin{eqnarray}
\label{equa:boundary_description_region_1}
{\rm DBND}_{(A,{\alpha})}^1(X)&=& ({\rm DPOS}_{(A,{\alpha})}^1(X) \cup {\rm DNEG}_{(A,{\alpha})}^1(X))^c,
\end{eqnarray}
as assumed in Equation (\ref{equa:implicit_boundary_description_region}). It should be noted that ${\rm DBND}_{(A,{\alpha})}^1(X)$ may not be equivalent with the following set:
\begin{eqnarray}
\bigcup \{ {\rm CDes}((x)_{G_A}^\alpha) \mid \neg((x)_{G_A}^\alpha \subseteq X)\wedge \neg((x)_{G_A}^\alpha \subseteq X^c)\}.
\end{eqnarray}
This is because the description of a similarity class $(x)_{G_A}^\alpha$ satisfying the condition $\neg((x)_{G_A}^\alpha \subseteq X)\wedge \neg((x)_{G_A}^\alpha \subseteq X^c)$ may share a formula with the description of another similarity class $(x^\prime)_{G_A}^\alpha$ satisfying the condition $((x^\prime)_{G_A}^\alpha \subseteq X)$ (or $((x^\prime)_{G_A}^\alpha \subseteq X^c)$). By defining the boundary description region as Equation (\ref{equa:boundary_description_region_1}), we eliminate the possibility of overlaps between the boundary region and the positive/negative region. As a result, Figure~\ref{fig:two_possibilities} shows the only two possibilities of relationships among the three description regions and the corresponding three-way decision rules derived. The two circles represent the positive and negative description regions, respectively, and the remaining region represents the boundary description region. The three types of rules derived are indicated as given by the legend. Thus, the non-commitment rules in (N) given in Definition \ref{def:incomplete_rule_1} include both the possible conflict between ${\rm DPOS}_{(A,{\alpha})}^1(X)$ and ${\rm DNEG}_{(A,{\alpha})}^1(X)$ and those rules induced from an implicitly defined boundary description region. The same result applies to other description regions proposed in this work. Similar discussions will be omitted in our following discussions.

\begin{example}
We illustrate the above approach to inducing three-way decision rules based on $\alpha$-similarity classes by continuing with Example~\ref{example:incomplete_computational_similarity_degree}. Suppose we choose a threshold $\alpha=0.3$. Then we get the 0.3-similarity classes based on the minimum similarity degree given by $M^{\min}$ in Equation (\ref{equa:example_incomplete_Matrix_min}) as follows:
\begin{alignat}{5}
(x_1)_{G_{AT}^{\min}}^{0.3}&=\{x_1,x_2\},&\qquad&(x_2)_{G_{AT}^{\min}}^{0.3}=\{x_1,x_2,x_3\},\nonumber\\
(x_3)_{G_{AT}^{\min}}^{0.3}&=\{x_2,x_3\},&&(x_4)_{G_{AT}^{\min}}^{0.3}=\{x_4,x_5,x_6\},\nonumber\\
(x_5)_{G_{AT}^{\min}}^{0.3}&=\{x_4,x_5,x_6\},&&(x_6)_{G_{AT}^{\min}}^{0.3}=\{x_4,x_5,x_6\},\nonumber\\
(x_7)_{G_{AT}^{\min}}^{0.3}&=\{x_7,x_8\},&&(x_8)_{G_{AT}^{\min}}^{0.3}=\{x_7,x_8\}.
\end{alignat}
Given a class $X=\{x_1,x_2,x_3,x_4\}$, the positive and negative description regions are constructed as follows:
\begin{eqnarray}
{\rm DPOS}_{(AT,0.3)}^{1~\min}(X)&=&\bigcup \{{\rm CDes}((x)_{G_{AT}^{\min}}^{0.3})\mid (x)_{G_{AT}^{\min}}^{0.3}\subseteq X\}\nonumber\\
&=&{\rm CDes}((x_1)_{G_{AT}^{\min}}^{0.3}) \cup {\rm CDes}((x_2)_{G_{AT}^{\min}}^{0.3}) \cup {\rm CDes}((x_3)_{G_{AT}^{\min}}^{0.3})\nonumber\\
&=&\{\langle a_1,1\rangle\wedge\langle a_2,2\rangle\wedge\langle a_3, 3\rangle\} \nonumber\\
&& \cup \{\langle a_1,1\rangle\wedge\langle a_2,1\rangle\wedge\langle a_3, 3\rangle,\langle a_1,1\rangle\wedge\langle a_2,2\rangle\wedge\langle a_3,3\rangle\} \nonumber\\
&&\cup \{ \langle a_1,1\rangle\wedge\langle a_2,1\rangle\wedge\langle a_3,3\rangle\}\nonumber\\
&=&\{\langle a_1,1\rangle\wedge\langle a_2,2\rangle\wedge\langle a_3, 3\rangle,\langle a_1,1\rangle\wedge\langle a_2,1\rangle\wedge\langle a_3,3\rangle\}, \nonumber\\
{\rm DNEG}_{(AT,0.3)}^{1~\min}(X)&=&\bigcup \{{\rm CDes}((x)_{G_{AT}^{\min}}^{0.3})\mid (x)_{G_{AT}^{\min}}^{0.3}\subseteq X^c\}\nonumber\\
&=&{\rm CDes}((x_7)_{G_{AT}^{\min}}^{0.3}) \cup {\rm CDes}((x_8)_{G_{AT}^{\min}}^{0.3}) \nonumber\\
&=&\{\langle a_1,{\rm NA}\rangle\wedge\langle a_2,1\rangle\wedge\langle a_3,0\rangle\} \nonumber\\
&& \cup \{\langle a_1,{\rm NA}\rangle\wedge\langle a_2,1\rangle\wedge\langle a_3,0\rangle,
\langle a_1,{\rm NA}\rangle\wedge\langle a_2,2\rangle\wedge\langle a_3,0\rangle,\nonumber\\
&&\langle a_1,{\rm NA}\rangle\wedge\langle a_2,3\rangle\wedge\langle a_3,0\rangle\}\nonumber\\
&=&\{\langle a_1,{\rm NA}\rangle\wedge\langle a_2,1\rangle\wedge\langle a_3,0\rangle,
\langle a_1,{\rm NA}\rangle\wedge\langle a_2,2\rangle\wedge\langle a_3,0\rangle,\nonumber\\
&&\langle a_1,{\rm NA}\rangle\wedge\langle a_2,3\rangle\wedge\langle a_3,0\rangle\}.
\end{eqnarray}
Accordingly, we get the three-way decision rules as follows:
\begin{eqnarray*}
{\rm (A)}  && \langle a_1,1\rangle\wedge\langle a_2,2\rangle\wedge\langle a_3,3\rangle \longrightarrow_{\rm A} X,\\
&& \langle a_1,1\rangle\wedge\langle a_2,1\rangle\wedge\langle a_3,3\rangle \longrightarrow_{\rm A} X;\\
{\rm (R)}  && \langle a_1,{\rm NA}\rangle\wedge\langle a_2,1\rangle \wedge\langle a_3,0\rangle \longrightarrow_{\rm R} X,\\
&& \langle a_1,{\rm NA}\rangle\wedge\langle a_2,2\rangle \wedge\langle a_3,0\rangle \longrightarrow_{\rm R} X,\\
&& \langle a_1,{\rm NA}\rangle\wedge\langle a_2,3\rangle \wedge\langle a_3,0\rangle \longrightarrow_{\rm R} X;\\
{\rm (N)}  && \textit{Otherwise,~neither~accept~nor~reject~} y.
\end{eqnarray*}

With the same threshold $\alpha=0.3$, we can get the 0.3-similarity classes based on the algebraic-product similarity degree given by matrix $M^{\prod}$ in Equation (\ref{equa:example_incomplete_Matrix_prod}) as follows:
\begin{alignat}{5}
(x_1)_{G_{AT}^{\prod}}^{0.3}&=\{x_1,x_2\},&\qquad& (x_2)_{G_{AT}^{\prod}}^{0.3}=\{x_1,x_2,x_3\},\nonumber\\
(x_3)_{G_{AT}^{\prod}}^{0.3}&=\{x_2,x_3\},&& (x_4)_{G_{AT}^{\prod}}^{0.3}=\{x_4\},\nonumber\\
(x_5)_{G_{AT}^{\prod}}^{0.3}&=\{x_5\},&& (x_6)_{G_{AT}^{\prod}}^{0.3}=\{x_6\},\nonumber\\
(x_7)_{G_{AT}^{\prod}}^{0.3}&=\{x_7,x_8\},&& (x_8)_{G_{AT}^{\prod}}^{0.3}=\{x_7,x_8\}.
\end{alignat}
Regarding the given class $X=\{x_1,x_2,x_3,x_4\}$, the positive and negative description regions are constructed as follows:
\begin{eqnarray}
{\rm DPOS}_{(A,0.3)}^{1~\prod}(X)&=&\bigcup \{{\rm CDes}((x)_{G_{AT}^{\prod}}^{0.3})\mid (x)_{G_{AT}^{\prod}}^{0.3}\subseteq X\}\nonumber\\
&=&{\rm CDes}((x_1)_{G_{AT}^{\prod}}^{0.3}) \cup {\rm CDes}((x_2)_{G_{AT}^{\prod}}^{0.3}) \cup {\rm CDes}((x_3)_{G_{AT}^{\prod}}^{0.3})\cup {\rm CDes}((x_4)_{G_{AT}^{\prod}}^{0.3})\nonumber\\
&=&\{\langle a_1,1\rangle\wedge\langle a_2,2\rangle\wedge\langle a_3, 3\rangle\} \nonumber\\
&& \cup \{\langle a_1,1\rangle\wedge\langle a_2,1\rangle\wedge\langle a_3, 3\rangle,\langle a_1,1\rangle\wedge\langle a_2,2\rangle\wedge\langle a_3,3\rangle\} \nonumber\\
&&\cup \{ \langle a_1,1\rangle\wedge\langle a_2,1\rangle\wedge\langle a_3,3\rangle\}\nonumber\\
&&\cup \{ \langle a_1,0\rangle\wedge\langle a_2,3\rangle\wedge\langle a_3,0\rangle,
\langle a_1,0\rangle\wedge\langle a_2,3\rangle\wedge\langle a_3,1\rangle,\nonumber\\
&&\langle a_1,0\rangle\wedge\langle a_2,3\rangle\wedge\langle a_3,3\rangle\}\nonumber\\
&=&\{\langle a_1,1\rangle\wedge\langle a_2,2\rangle\wedge\langle a_3, 3\rangle,\langle a_1,1\rangle\wedge\langle a_2,1\rangle\wedge\langle a_3,3\rangle\nonumber\\
&&\langle a_1,0\rangle\wedge\langle a_2,3\rangle\wedge\langle a_3,0\rangle,
\langle a_1,0\rangle\wedge\langle a_2,3\rangle\wedge\langle a_3,1\rangle,\nonumber\\
&&\langle a_1,0\rangle\wedge\langle a_2,3\rangle\wedge\langle a_3,3\rangle\}, \nonumber\\
{\rm DNEG}_{(A,0.3)}^{1~\prod}(X)&=&\bigcup \{{\rm CDes}((x)_{G_{AT}^{\prod}}^{0.3})\mid (x)_{G_{AT}^{\prod}}^{0.3}\subseteq X^c\}\nonumber\\
&=&{\rm CDes}((x_5)_{G_{AT}^{\prod}}^{0.3}) \cup{\rm CDes}((x_6)_{G_{AT}^{\prod}}^{0.3}) \cup {\rm CDes}((x_7)_{G_{AT}^{\prod}}^{0.3}) \cup {\rm CDes}((x_8)_{G_{AT}^{\prod}}^{0.3}) \nonumber\\
&=& \{\langle a_1,0\rangle\wedge\langle a_2,3\rangle\wedge\langle a_3,1\rangle,
\langle a_1,1\rangle\wedge\langle a_2,3\rangle\wedge\langle a_3,1\rangle\}\nonumber\\
&& \cup \{\langle a_1,0\rangle\wedge\langle a_2,3\rangle\wedge\langle a_3,1\rangle,
\langle a_1,0\rangle\wedge\langle a_2,3\rangle\wedge\langle a_3,3\rangle,\nonumber\\
&&\langle a_1,1\rangle\wedge\langle a_2,3\rangle\wedge\langle a_3,1\rangle,
\langle a_1,1\rangle\wedge\langle a_2,3\rangle\wedge\langle a_3,3\rangle\}\nonumber\\
&& \cup \{\langle a_1,{\rm NA}\rangle\wedge\langle a_2,1\rangle\wedge\langle a_3,0\rangle\} \nonumber\\
&& \cup \{\langle a_1,{\rm NA}\rangle\wedge\langle a_2,1\rangle\wedge\langle a_3,0\rangle,
\langle a_1,{\rm NA}\rangle\wedge\langle a_2,2\rangle\wedge\langle a_3,0\rangle,\nonumber\\
&&\langle a_1,{\rm NA}\rangle\wedge\langle a_2,3\rangle\wedge\langle a_3,0\rangle\}\nonumber\\
&=&\{\langle a_1,0\rangle\wedge\langle a_2,3\rangle\wedge\langle a_3,1\rangle,
\langle a_1,0\rangle\wedge\langle a_2,3\rangle\wedge\langle a_3,3\rangle,\nonumber\\
&&\langle a_1,1\rangle\wedge\langle a_2,3\rangle\wedge\langle a_3,1\rangle,
\langle a_1,1\rangle\wedge\langle a_2,3\rangle\wedge\langle a_3,3\rangle, \nonumber\\
&&\langle a_1,{\rm NA}\rangle\wedge\langle a_2,1\rangle\wedge\langle a_3,0\rangle,
\langle a_1,{\rm NA}\rangle\wedge\langle a_2,2\rangle\wedge\langle a_3,0\rangle,\nonumber\\
&&\langle a_1,{\rm NA}\rangle\wedge\langle a_2,3\rangle\wedge\langle a_3,0\rangle\}.
\end{eqnarray}
These two description regions have a nonempty overlap:
\begin{equation}
{\rm DPOS}_{(A,0.3)}^{1~\prod}(X) \cap {\rm DNEG}_{(A,0.3)}^{1~\prod}(X) = \{\langle a_1,0\rangle\wedge\langle a_2,3\rangle\wedge\langle a_3,1\rangle,\langle a_1,0\rangle\wedge\langle a_2,3\rangle\wedge\langle a_3,3\rangle\}.
\end{equation}
Accordingly, we get the three-way decision rules as follows:
\begin{eqnarray*}
{\rm (A)}  && \langle a_1,1\rangle\wedge\langle a_2,2\rangle\wedge\langle a_3,3\rangle \longrightarrow_{\rm A} X,\\
&& \langle a_1,1\rangle\wedge\langle a_2,1\rangle\wedge\langle a_3,3\rangle \longrightarrow_{\rm A} X,\\
&& \langle a_1,0\rangle\wedge\langle a_2,3\rangle\wedge\langle a_3,0\rangle \longrightarrow_{\rm A} X;\\
{\rm (R)}  && \langle a_1,1\rangle\wedge\langle a_2,3\rangle\wedge\langle a_3,1\rangle \longrightarrow_{\rm R} X,\\
&&\langle a_1,1\rangle\wedge\langle a_2,3\rangle\wedge\langle a_3,3\rangle \longrightarrow_{\rm R} X,\\
&&\langle a_1,{\rm NA}\rangle\wedge\langle a_2,1\rangle \wedge\langle a_3,0\rangle \longrightarrow_{\rm R} X,\\
&& \langle a_1,{\rm NA}\rangle\wedge\langle a_2,2\rangle \wedge\langle a_3,0\rangle \longrightarrow_{\rm R} X,\\
&& \langle a_1,{\rm NA}\rangle\wedge\langle a_2,3\rangle \wedge\langle a_3,0\rangle \longrightarrow_{\rm R} X;\\
{\rm (N)}  && \textit{Otherwise,~neither~accept~nor~reject~} y.
\end{eqnarray*}
\end{example}

\subsection{Three-way decision based on approximability of objects}
\label{sec:approximability}

Instead of constructing similarity classes and using their descriptions, we can directly consider and evaluate the descriptions of individual objects. The description of an individual object is used as the left-hand-side of a classification rule. Then our task is to determine the right-hand-side to be an acceptance, a rejection, or a non-commitment decision. In our work, we measure the degree to which a description of an object implies an acceptance or a rejection decision by using fuzzy logic operators. The two degrees are defined as the positive and negative approximability of the object and are used to construct positive and negative description regions.

In the computational formulation with a complete table, we use the subset relationship $[x]_A \subseteq X$ for defining the positive region in Definition \ref{def:three_regions_complete_descriptions}, which can be equivalently expressed through classical logic operators as:
\begin{eqnarray}
\label{equa:logic_implication_POS}
[x]_A \subseteq X &\Longleftrightarrow& \forall\, y \in O\!B, y \in [x]_A \longrightarrow y\in X,\nonumber\\
&\Longleftrightarrow& \bigwedge_{y \in O\!B} (y \in [x]_A \longrightarrow y\in X).
\end{eqnarray}
This indicates at least two directions of generalizing a positive region into the context of an incomplete table. The first direction is to generalize the subset relationship $\subseteq$ into a subsethood measure. There are many studies on subsethood measures~\cite{Hu2019on,bustince2006definition,dimuro2017ql,fan1999subsethood,sinha1993fuzzification,young1996fuzzy,zhang2009hybrid} which can be directly used. In our work, we consider a second direction where we generalize the classical logic conjunction and implication into fuzzy conjunction and implication.

As discussed when defining the similarity degree $G_A$, a T-norm can be used as a fuzzy conjunction. A fuzzy implication is a function ${\rm I}:[0,1]^2\longrightarrow [0,1]$ that satisfies the following two properties:
\begin{eqnarray*}
(1) &&{\rm I}(1,0)=0,\\
(2) &&{\rm I}(1,1)={\rm I}(0,1)={\rm I}(0,0)=1.
\end{eqnarray*}
It can be considered as a quantitative generalization of the truth values in classical logic implication. Intuitively, for two real numbers $u_1,u_2 \in [0,1]$, the value ${\rm I}(u_1,u_2)$ quantifies the degree to which a truth value of $u_1$ implies a truth value of $u_2$. By requiring the above two properties, the classical logic implication is a special case of ${\rm I}$.

In order to generalize the classical logic implication in Equation (\ref{equa:logic_implication_POS}) by a fuzzy implication, we need to consider how to generalize the premise $y \in [x]_A$ and the conclusion $y \in X$ into two real numbers in the context of incomplete information. The premise $y \in [x]_A$ can be interpreted as $y$ and $x$ are equivalent. In the context of incomplete information, this can be generalized into how $y$ is similar to $x$ regarding $A$, which is measured by $G_A(x,y)$. The conclusion $y \in X$ can be generalized into an indicator or a characteristic function. More specifically, the indicator function $1_X:~O\!B\longrightarrow \{0,1\}$ indicates the membership of objects from $O\!B$ in a set $X\subseteq O\!B$, which is formally defined as:
\begin{align}
1_X(y)=
\begin{cases}
  1, & y\in X, \\
  0, & y\notin X.
\end{cases}
\end{align}
Thus, the implication $y \in [x]_A \longrightarrow y\in X$ in Equation (\ref{equa:logic_implication_POS}) can be generalized into the following value with respect to incomplete information:
\begin{equation}
{\rm I}(G_A(x,y), 1_X(y)).
\end{equation}
By using a T-norm to generalize the classical logic conjunction, the entire condition in Equation (\ref{equa:logic_implication_POS}) is generalized with respect to incomplete information as:
\begin{equation}
\label{equa:fuzzy_condition}
{\rm T}_{y\in O\!B}{\rm I}(G_A(x,y), 1_X(y)).
\end{equation}

Based on the above discussion, we can use Equation \eqref{equa:fuzzy_condition} as a measure of the degree to which $x$ implies a positive instance of a given class $X\subseteq O\!B$. Similarly, by using $X^c$ instead of $X$, we immediately get a measure of the degree to which $x$ implies a negative instance of $X$. We define these two measures as the positive and negative approximability of an object $x$ regarding $X$.

\begin{definition}
\label{def:approximability}
In a set-valued table $S\!T=(O\!B,AT,V\cup \{{\rm NA}\},S)$, for a set of attributes $A\subseteq AT$, the positive and negative approximability of objects regarding a given class $X\subseteq O\!B$ are respectively given by two mappings $papr_{G_A}^X: O\!B \longrightarrow [0,1]$ and $napr_{G_A}^X: O\!B \longrightarrow [0,1]$ where:
\begin{eqnarray}
papr_{G_A}^X(x)&=&{\rm T}_{y\in O\!B} {\rm I}(G_A(x,y),1_X(y)),\nonumber\\
napr_{G_A}^X(x)&=&{\rm T}_{y\in O\!B} {\rm I}(G_A(x,y),1_{X^c}(y)).
\end{eqnarray}
\end{definition}

The specific calculation of approximability depends on the specific T-norm, fuzzy implication I, and the similarity degree $G_A$. A fuzzy implication can be obtained from either a T-norm or a combination of a Triangle-conorm (S-norm for short, which is a generalization of classical logic disjunction) and a fuzzy negation~\cite{baczynski2008fuzzy}. These two types are called R-implication and S-implication, respectively. In this work, we consider S-implication as an illustration. In our previous discussion, we use the minimum T-norm ${\rm T}^{\min}$ and the algebraic-product T-norm ${\rm T}^{\prod}$ to calculate the similarity degrees. As generalizations of classical logic conjunction and disjunction, T-norms and S-norms are dual fuzzy logic operators. Based on the dual S-norms of ${\rm T}^{\min}$ and ${\rm T}^{\prod}$, one can get the following Kleene-Dienes implication ${\rm I_{KD}}$~\cite{baczynski2008fuzzy,dienes1949on,kleene1938notation} and the Reichenbach implication ${\rm I_{RC}} $~\cite{baczynski2008fuzzy,reichenbach1935wahrscheinlichkeitslogik}, respectively: for two real numbers $u_1,u_2 \in [0,1]$,
\begin{eqnarray}
\label{equa:implication_KD_RC}
{\rm I_{KD}} (u_1,u_2) &=& \max (1-u_1, u_2),\nonumber\\
{\rm I_{RC}} (u_1,u_2) &=&  1 - u_1+ u_1 \cdot u_2.
\end{eqnarray}
Thus, we get the approximability based on the minimum T-norm ${\rm T}^{\min}$ and the corresponding fuzzy implication ${\rm I_{KD}}$ as:
\begin{eqnarray}
\label{equa:approximability_min}
papr_{G_A^{\min}}^X(x)&=&{\rm T}_{y\in O\!B}^{\min}{\rm I_{KD}}\big(G_A^{\min}(x,y),1_X(y)\big)\nonumber\\
&=&{\rm T}_{y\in O\!B}^{\min}\big(\max(1-G_A^{\min}(x,y),1_X(y))\big)\nonumber\\
&=&\min\big( \min_{y\in X}(\max(1-G_A^{\min}(x,y),1)),
\min_{y\in X^c}(\max(1-G_A^{\min}(x,y),0) )\big)\nonumber\\
&=&\min\big(1,\min_{y\in X^c}(1-G_A^{\min}(x,y)) \big)\nonumber\\
&=&\min_{y\in X^c}\big(1-G_A^{\min}(x,y)\big),\nonumber\\
napr_{G_A^{\min}}^X(x)&=&{\rm T}^{\min}_{y\in O\!B} {\rm I_{KD}}\big(G_A^{\min}(x,y),1_{X^c}(y)\big)\nonumber\\
&=&{\rm T}_{y\in O\!B}^{\min}\big(\max(1-G_A^{\min}(x,y),1_{X^c}(y))\big)\nonumber\\
&=&\min\big( \min_{y\in X}(\max(1-G_A^{\min}(x,y),0)),
\min_{y\in X^c}(\max(1-G_A^{\min}(x,y),1))\big)\nonumber\\
&=&\min\big( \min_{y\in X}(1-G_A^{\min}(x,y)),1 \big)\nonumber\\
&=&\min_{y\in X}\big(1-G_A^{\min}(x,y)\big).
\end{eqnarray}
Similarly, the approximability based on the algebraic-product T-norm ${\rm T}^{\prod}$ and the corresponding fuzzy implication ${\rm I_{RC}}$ is:
\begin{eqnarray}
\label{equa:approximability_prod}
papr_{G_A^{\prod}}^X(x)&=&{\rm T}^{\prod}_{y\in O\!B} {\rm I_{RC}}\big(G_A^{\prod}(x,y),1_X(y)\big)
\nonumber\\
&=&{\rm T}_{y\in O\!B}^{\prod}\big(1-G_A^{\prod}(x,y)+G_A^{\prod}(x,y)
\cdot 1_X(y)\big)\nonumber\\
&=&\prod_{y\in X}\big(1-G_A^{\prod}(x,y)+G_A^{\prod}(x,y)\big)\cdot
\prod_{y\in X^c}\big(1-G_A^{\prod}(x,y)\big)\nonumber\\
&=&\prod_{y\in X^c}\big(1-G_A^{\prod}(x,y)\big),\nonumber\\
napr_{G_A^{\prod}}^X(x)&=&{\rm T}^{\prod}_{y\in O\!B} {\rm I_{RC}}\big(G_A^{\prod}(x,y),1_{X^c}(y)\big)\nonumber\\
&=&{\rm T}_{y\in O\!B}^{\prod}\big(1-G_A^{\prod}(x,y)+G_A^{\prod}(x,y)\cdot 1_{X^c}(y)\big)\nonumber\\
&=&\prod_{y\in X}\big(1-G_A^{\prod}(x,y)\big)\cdot
\prod_{y\in X^c}\big(1-G_A^{\prod}(x,y)+G_A^{\prod}(x,y)\big)\nonumber\\
&=&\prod_{y\in X}\big(1-G_A^{\prod}(x,y)\big).
\end{eqnarray}

By applying a threshold on the approximability of objects, we can construct the positive and negative description regions.

\begin{definition}
In a set-valued table $S\!T=(O\!B,AT,V\cup \{{\rm NA}\},S)$, for a set of attributes $A\subseteq AT$ and a threshold $\alpha\in [0,1]$, the positive and negative description regions regarding a given class $X \subseteq O\!B$ are defined as:
\begin{eqnarray}
{\rm DPOS}_{(A,{\alpha})}^2(X)&=&\bigcup \{{\rm CDes}_A(x)\mid papr_{G_A}^X(x) \geq \alpha\},\nonumber\\
{\rm DNEG}_{(A,{\alpha})}^2(X)&=&\bigcup \{{\rm CDes}_A(x)\mid napr_{G_A}^X(x) \geq \alpha\},
\end{eqnarray}
where ${\rm CDes}_A(x)$ is the description of $x$ defined in Equation (\ref{equa:description_CDes}).
\end{definition}

Due to the selection of the threshold $\alpha$, the two regions ${\rm DPOS}_{(A,{\alpha})}^2(X)$ and ${\rm DNEG}_{(A,{\alpha})}^2(X)$ may have overlap. In this case, we induce non-commitment rules.

\begin{definition}
In a set-valued table $S\!T=(O\!B,AT,V\cup \{{\rm NA}\},S)$, for a set of attributes $A\subseteq AT$, the acceptance, rejection, and non-commitment rules are induced as follows:
for any $y\in O\!B$,
\begin{eqnarray*}
{\rm (A)}  &&p_a \longrightarrow_{\rm A} X:~\textit{ if}~y\models p_a~\textit{ for}~p_a\in {\rm DPOS}_{(A,{\alpha})}^2(X)-{\rm DNEG}_{(A,{\alpha})}^2(X), \textit{ then~accept~}y,\\
{\rm (R)}  &&p_r \longrightarrow_{\rm R} X:~\textit{ if}~y\models p_r~\textit{ for}~p_r\in {\rm DNEG}_{(A,{\alpha})}^2(X)-{\rm DPOS}_{(A,{\alpha})}^2(X),\textit{ then~reject~}y,\\
{\rm (N)}  &&\textit {If~neither~{\rm(A)}~nor~{\rm(R)}~applies,~then~neither~accept~nor~reject~}y.
\end{eqnarray*}
\end{definition}

The above three-way decision rules are similar to Definition~\ref{def:incomplete_rule_1}.

\begin{example}
\label{example:incomplete_approximability}
We illustrate the above approach to inducing three-way decision rules based on approximability of objects by continuing with Example~\ref{example:incomplete_computational_similarity_degree}. Given a class $X=\{x_1,x_2,x_3,x_4\}$, we compute the approximability of objects based on ${\rm T}^{\min}$ and ${\rm T}^{\prod}$ by using Equations (\ref{equa:approximability_min}) and (\ref{equa:approximability_prod}), which is given in Table~\ref{tab:example_incomplete_approximability}.

\begin{table}[!ht]
\centering
\caption{The approximability of objects for Example \ref{example:incomplete_approximability}}
\label{tab:example_incomplete_approximability}
\setlength{\tabcolsep}{1em}
\renewcommand{\arraystretch}{1.2}
\begin{tabular}{|c|c|c||c|c|}
\hline
    & $papr_{G_{AT}^{\min}}^X$ & $napr_{G_{AT}^{\min}}^X$ & $papr_{G_{AT}^{\prod}}^X$ & $papr_{G_{AT}^{\prod}}^X$   \\
  \hline
  $x_1$ &1    &0    &1    & 0\\
  $x_2$ &1    &0    &1    &0 \\
  $x_3$ &1    &0    &1    &0\\
  $x_4$ &0.667&0    &0.694&0\\
  $x_5$ &0    &0.667&0    &0.833\\
  $x_6$ &0    &0.667&0    &0.833\\
  $x_7$ &0    &1    &0    & 1\\
  $x_8$ &0    &1    &0    & 1\\
\hline
\end{tabular}
\end{table}
We take the computation of $papr_{G_{AT}^{\min}}^X(x_4)$ as an illustration. According to Equation (\ref{equa:approximability_min}) and the matrix $M^{\min}$ in Equation (\ref{equa:example_incomplete_Matrix_min}), we have:
\begin{eqnarray}
papr_{G_{AT}^{\min}}^X(x_4)&=&\min_{y\in X^c}(1-G_{AT}^{\min}(x_4,y))\nonumber\\
&=&\min(\, 1-G_{AT}^{\min}(x_4,x_5),1-G_{AT}^{\min}(x_4,x_6),1-G_{AT}^{\min}(x_4,x_7),1-G_{AT}^{\min}(x_4,x_8) \,)\nonumber\\
&=&\min(\, 1-0.333, 1-0.333, 1-0, 1-0\,)\nonumber\\
&=&0.667.
\end{eqnarray}
Suppose we choose a threshold $\alpha=0.8$. The positive and negative description regions based on ${\rm T}^{\min}$ are:
\begin{eqnarray}
{\rm DPOS}_{(AT,0.8)}^{2~\min}(X)&=&\bigcup \{{\rm CDes}_{AT}(x)\mid papr_{G_{AT}^{\min}}^X(x) \geq 0.8\}\nonumber\\
&=&\bigcup \{{\rm CDes}_{AT}(x_1),{\rm CDes}_{AT}(x_2),{\rm CDes}_{AT}(x_3)\}\nonumber\\
&=&\{\langle a_1,1\rangle\wedge\langle a_2,2\rangle\wedge\langle a_3,3\rangle\}\nonumber\\
&& \cup \{\langle a_1,1\rangle\wedge\langle a_2,1\rangle\wedge\langle a_3,3\rangle,\langle a_1,1\rangle\wedge\langle a_2,2\rangle\wedge\langle a_3,3\rangle\}\nonumber\\
&& \cup \{\langle a_1,1\rangle\wedge\langle a_2,1\rangle\wedge\langle a_3,3\rangle\} \nonumber\\
&=&\{\langle a_1,1\rangle\wedge\langle a_2,1\rangle\wedge\langle a_3,3\rangle,\langle a_1,1\rangle\wedge\langle a_2,2\rangle\wedge\langle a_3,3\rangle\},\nonumber\\
{\rm DNEG}_{(AT,0.8)}^{2~\min}(X)&=&\bigcup \{{\rm CDes}_{AT}(x)\mid napr_{G_{AT}^{\min}}^X(x) \geq 0.8\}\nonumber\\
&=&\bigcup \{{\rm CDes}_{AT}(x_7),{\rm CDes}_{AT}(x_8)\}\nonumber\\
&=&\{\langle a_1,{\rm NA}\rangle\wedge\langle a_2,1\rangle\wedge\langle a_3,0\rangle\}\nonumber\\
&& \cup \{\langle a_1,{\rm NA}\rangle\wedge\langle a_2,1\rangle\wedge\langle a_3,0\rangle,\langle a_1,{\rm NA}\rangle\wedge\langle a_2,2\rangle\wedge\langle a_3,0\rangle,\nonumber\\
&&\langle a_1,{\rm NA}\rangle\wedge\langle a_2,3\rangle\wedge\langle a_3,0\rangle\}\nonumber\\
&=&\{\langle a_1,{\rm NA}\rangle\wedge\langle a_2,1\rangle\wedge\langle a_3,0\rangle,\langle a_1,{\rm NA}\rangle\wedge\langle a_2,2\rangle\wedge\langle a_3,0\rangle,\nonumber\\
&&\langle a_1,{\rm NA}\rangle\wedge\langle a_2,3\rangle\wedge\langle a_3,0\rangle\}.
\end{eqnarray}
Accordingly, we get the three-way decision rules as follows:
\begin{eqnarray*}
{\rm (A)}  & &\langle a_1,1\rangle\wedge\langle a_2,1\rangle\wedge\langle a_3,3\rangle \longrightarrow_{\rm A}X,\\
&&\langle a_1,1\rangle\wedge\langle a_2,2\rangle\wedge\langle a_3,3\rangle \longrightarrow_{\rm A}X;\\
{\rm (R)}  & &\langle a_1,{\rm NA}\rangle\wedge\langle a_2,1\rangle \wedge\langle a_3,0\rangle \longrightarrow_{\rm R}X,\\
&&\langle a_1,{\rm NA}\rangle\wedge\langle a_2,2\rangle \wedge\langle a_3,0\rangle \longrightarrow_{\rm R}X,\\
& &\langle a_1,{\rm NA}\rangle\wedge\langle a_2,3\rangle \wedge\langle a_3,0\rangle \longrightarrow_{\rm R}X;\\
{\rm (N)}  & & \textit{Otherwise,~neither~accept~nor~reject~} y.
\end{eqnarray*}
Similarly, one can get the positive and negative description regions based on ${\rm T}^{\prod}$:
\begin{eqnarray}
{\rm DPOS}_{(AT,0.8)}^{2~\prod}(X)&=&\bigcup \{{\rm CDes}_{AT}(x)\mid papr_{G_{AT}^{\prod}}^X(x) \geq 0.8\}\nonumber\\
&=& \bigcup \{{\rm CDes}_{AT}(x_1),{\rm CDes}_{AT}(x_2),{\rm CDes}_{AT}(x_3)\}\nonumber\\
&=&\{\langle a_1,1\rangle\wedge\langle a_2,2\rangle\wedge\langle a_3,3\rangle\}\nonumber\\
&& \cup \{\langle a_1,1\rangle\wedge\langle a_2,1\rangle\wedge\langle a_3,3\rangle,\langle a_1,1\rangle\wedge\langle a_2,2\rangle\wedge\langle a_3,3\rangle\}\nonumber\\
&& \cup \{\langle a_1,1\rangle\wedge\langle a_2,1\rangle\wedge\langle a_3,3\rangle\} \nonumber\\
&=&\{\langle a_1,1\rangle\wedge\langle a_2,1\rangle\wedge\langle a_3,3\rangle,
\langle a_1,1\rangle\wedge\langle a_2,2\rangle\wedge\langle a_3,3\rangle\},\nonumber\\
{\rm DNEG}_{(AT,0.8)}^{2~\prod}(X)&=&\bigcup \{{\rm CDes}_{AT}(x)\mid napr_{G_{AT}^{\prod}}^X(x) \geq 0.8\}\nonumber\\
&=&\bigcup \{{\rm CDes}_{AT}(x_5),{\rm CDes}_{AT}(x_6),{\rm CDes}_{AT}(x_7),{\rm CDes}_{AT}(x_8)\}\nonumber\\
&=&\{\langle a_1,0\rangle\wedge\langle a_2,3\rangle\wedge\langle a_3,1\rangle,
\langle a_1,1\rangle\wedge\langle a_2,3\rangle\wedge\langle a_3,1\rangle\}\nonumber\\
&&\cup \{\langle a_1,0\rangle\wedge\langle a_2,3\rangle\wedge\langle a_3,1\rangle,
\langle a_1,0\rangle\wedge\langle a_2,3\rangle\wedge\langle a_3,3\rangle,\nonumber\\
&&\langle a_1,1\rangle\wedge\langle a_2,3\rangle\wedge\langle a_3,1\rangle,
\langle a_1,1\rangle\wedge\langle a_2,3\rangle\wedge\langle a_3,3\rangle\}\nonumber\\
&&\cup \{\langle a_1,{\rm NA}\rangle\wedge\langle a_2,1\rangle\wedge\langle a_3,0\rangle\}\nonumber\\
&& \cup \{\langle a_1,{\rm NA}\rangle\wedge\langle a_2,1\rangle\wedge\langle a_3,0\rangle,\langle a_1,{\rm NA}\rangle\wedge\langle a_2,2\rangle\wedge\langle a_3,0\rangle,\nonumber\\
&&\langle a_1,{\rm NA}\rangle\wedge\langle a_2,3\rangle\wedge\langle a_3,0\rangle\}\nonumber\\
&=&\{\langle a_1,0\rangle\wedge\langle a_2,3\rangle\wedge\langle a_3,1\rangle,
\langle a_1,0\rangle\wedge\langle a_2,3\rangle\wedge\langle a_3,3\rangle,\nonumber\\
&&\langle a_1,1\rangle\wedge\langle a_2,3\rangle\wedge\langle a_3,1\rangle,
\langle a_1,1\rangle\wedge\langle a_2,3\rangle\wedge\langle a_3,3\rangle,\nonumber\\
&&\langle a_1,{\rm NA}\rangle\wedge\langle a_2,1\rangle\wedge\langle a_3,0\rangle,
\langle a_1,{\rm NA}\rangle\wedge\langle a_2,2\rangle\wedge\langle a_3,0\rangle,\nonumber\\
&&\langle a_1,{\rm NA}\rangle\wedge\langle a_2,3\rangle\wedge\langle a_3,0\rangle\}.
\end{eqnarray}
Accordingly, we get the three-way decision rules as follows:
\begin{eqnarray*}
{\rm (A)}  & &\langle a_1,1\rangle\wedge\langle a_2,1\rangle\wedge\langle a_3,3\rangle \longrightarrow_{\rm A}X,\\
&&\langle a_1,1\rangle\wedge\langle a_2,2\rangle\wedge\langle a_3,3\rangle \longrightarrow_{\rm A}X;\\
{\rm (R)}  &&\langle a_1,0\rangle\wedge\langle a_2,3\rangle\wedge\langle a_3,1\rangle \longrightarrow_{\rm R}X,\\
&&\langle a_1,0\rangle\wedge\langle a_2,3\rangle\wedge\langle a_3,3\rangle \longrightarrow_{\rm R}X,\\
&&\langle a_1,1\rangle\wedge\langle a_2,3\rangle\wedge\langle a_3,1\rangle \longrightarrow_{\rm R}X,\\
&&\langle a_1,1\rangle\wedge\langle a_2,3\rangle\wedge\langle a_3,3\rangle \longrightarrow_{\rm R}X,\\
&&\langle a_1,{\rm NA}\rangle\wedge\langle a_2,1\rangle \wedge\langle a_3,0\rangle \longrightarrow_{\rm R}X,\\
&&\langle a_1,{\rm NA}\rangle\wedge\langle a_2,2\rangle \wedge\langle a_3,0\rangle \longrightarrow_{\rm R}X,\\
& &\langle a_1,{\rm NA}\rangle\wedge\langle a_2,3\rangle \wedge\langle a_3,0\rangle \longrightarrow_{\rm R}X;\\
{\rm (N)}  & &\textit{Otherwise,~neither~accept~nor~reject~} y.
\end{eqnarray*}
\end{example}

In our above discussion on the computational formulation, we consider the value ${\rm NA}$ as a meaningful value and accordingly, an attribute-value pair $\langle a, {\rm NA}\rangle$ is also meaningful to appear in a decision rule. In practical situations, the meaningfulness of the value ${\rm NA}$ depends on the semantics of a specific attribute. If ${\rm NA}$ is meaningful regarding a specific attribute $a$, then a pair $\langle a, {\rm NA}\rangle$ can be included in a decision rule. Otherwise, one may remove such a pair from a decision rule induced in our above discussion. For example, if a person takes the value ${\rm NA}$ on an attribute ``pregnancy'', it may imply that the person is a male. In this case, the value ${\rm NA}$ on the attribute ``pregnancy'' contains useful information and a pair $\langle {\rm pregnancy}, {\rm NA}\rangle$ should be kept in a decision rule. If a person takes the value ${\rm NA}$ on an attribute ``X-Ray'', the X-Ray examination result is not available for this person due to some reason (e.g. the person was not asked to take the examination). The value ${\rm NA}$ in this case doesn't convey useful information about the result of X-Ray. And it is not meaningful to include a pair $\langle {\rm X \mhyphen Ray }, {\rm NA}\rangle$ in any decision rule.

Based on the similarity degree, we discussed two approaches to the computational formulation of three-way decision with incomplete information. One approach is based on $\alpha$-similarity classes derived by using the similarity degree, and another approach is based on the approximability of objects calculated through the similarity degree. In the next section, we will focus on the conceptual formulation that starts from the satisfiability degree of formulas in a description language.

\section{Satisfiability-based three-way decision in an incomplete table - A conceptual formulation}
\label{sec:conceptual_incomplete}

In this section, we generalize the conceptual formulation in a complete table into a set-valued table to analyze incomplete information. More specifically, we start from the descriptions of objects, that is, the logic formulas in a description language ${\rm CDL}_A$. Based on a proposed measure of satisfiability degree of formulas, we discuss two approaches to inducing three-way decision rules. One approach defines $\alpha$-meaning sets of formulas, and another approach measures the confidence of formulas.

\subsection{The satisfiability degree of a formula}

Although there is incomplete information in a set-valued table, the domain $V_a$ for each attribute $a\in AT$ is exactly known. Therefore, the formulas in ${\rm CDL}_A$ can be completely formed following Definition~\ref{def:CDL}. However, due to our incomplete knowledge about the actual value of an object on an attribute, the satisfiability of a formula by an object may be uncertain. Thus, we generalize the satisfiability of formulas in a complete table into a quantitative measure of satisfiability degree. More specifically, we use the probability of an object satisfying a formula.

\begin{definition}
In a set-valued table $S\!T=(O\!B,AT,V\cup{\rm NA},S)$, for a subset of attributes $A \subseteq AT$, the satisfiability degree of formulas in ${\rm CDL}_A$ by objects in $O\!B$ is given by a mapping $D: O\!B \times {\rm CDL}_A \longrightarrow [0,1]$ which is defined as:
\begin{eqnarray}
{\rm (1)} && D(x \models \langle a,v\rangle) = \frac{|s_a(x)\cap \{v\}|}{|s_a(x)|},\nonumber\\
{\rm (2)} && D(x \models p \wedge q) = {\rm T} (D(x \models p), D(x \models q)),
\end{eqnarray}
where {\rm T} is a T-norm, $s_a(x)$ is the set of values taken by $x$ on $a$, and $p,q,p \wedge q \in {\rm CDL}_A$.
\end{definition}

In the above definition, we use a notation $D(x \models p)$ to denote the degree to which an object $x \in O\!B$ satisfies a formula $p \in {\rm CDL}_A$. Compared with a more precise mathematical notation $D(x, p)$ where $(x,p) \in O\!B \times {\rm CDL}_A$, the notation $D(x \models p)$ more intuitively indicates a generalization from the qualitative satisfiability $x \models p$ defined with respect to complete information to the corresponding quantitative satisfiability degree defined with respect to incomplete information.

The satisfiability degree $D(x \models \langle a,v\rangle)$ is defined as the probability of $x$ taking an actual value $v$ on $a$. It is also the probability of $x$ actually satisfying the atomic formula $\langle a,v\rangle$. The satisfiability degree of a composite formula is calculated through the satisfiability degree of atomic formulas by using a T-norm. As a result, for a formula $p=\langle a_1,v_1\rangle \wedge \langle a_2,v_2\rangle\wedge \cdots \wedge \langle a_k,v_k\rangle\in {\rm CDL}_A$, its satisfiability degree by an object $x \in O\!B$ can be calculated as:
\begin{eqnarray}
D(x\models p) &=& {\rm T}_{1\leq i\leq k}\big(D(x \models \langle a_i,v_i\rangle)\big)\nonumber\\
&=&{\rm T}_{1\leq i\leq k}\big(\frac{|s_{a_i}(x)\cap \{v_i\}|}{|s_{a_i}(x)|}\big),
\end{eqnarray}
which represents the probability of $x$ actually satisfying $p$.

\begin{example}
\label{example:incomplete_satisfiability}
We illustrate the calculation of satisfiability degree of formulas with Table~\ref{tab:example_set-valued_table_duplicate} which is a duplicate of Table~\ref{tab:example_incomplete_tables}(b). We consider the two T-norms ${\rm T}^{\min}$ and ${\rm T}^{\prod}$ to compute the satisfiability degrees of formulas in ${\rm CDL}_{AT}$. The results are given in Table~\ref{tab:example_incomplete_meaning_sets}, where we use $\frac{D}{x_i}$ to represent a satisfiability degree $D$ of a formula by an object $x_i$. For brevity, we omit a satisfiability degree of 0. Thus, a blank cell means that we have a satisfiability degree 0 of the corresponding formula by every object.

\begin{table}[!ht]
\centering
\caption{A duplicate of Table~\ref{tab:example_incomplete_tables}(b)}
\label{tab:example_set-valued_table_duplicate}
\setlength{\tabcolsep}{1em}
\renewcommand{\arraystretch}{1.2}
\begin{tabular}{c|ccc}
\hline
           & $a_1$   & $a_2$         & $a_3$  \\ \hline
  $x_1$    &1        &2              &3 \\
  $x_2$    &1        &$\{1,2\}$      &3 \\
  $x_3$    &1        &1              &3 \\
  $x_4$    &0        &3              &$\{0,1,3\}$ \\
  $x_5$    &$\{0,1\}$&3              &1 \\
  $x_6$    &$\{0,1\}$&3              &$\{1,3\}$ \\
  $x_7$    &NA       &1              &0 \\
  $x_8$    &NA       &$\{1,2,3\}$    &0\\
\hline
\end{tabular}
\end{table}

\begin{table}[!ht]
\centering
\caption{The satisfiability degrees of formulas in ${\rm CDL}_{A}$ for Table~\ref{tab:example_set-valued_table_duplicate}}
\label{tab:example_incomplete_meaning_sets}
\setlength{\tabcolsep}{0.5em}
\renewcommand{\arraystretch}{1.2}
\scalebox{0.83}{
\begin{tabular}{|c|c|c|c||c|c|c|c|}
\hline
{\multirow{2}{*}{Label}}&{\multirow{2}{*}{Formula}}& \multicolumn{2}{c||}{Satisfiability degree}
&{\multirow{2}{*}{Label}}&{\multirow{2}{*}{Formula}}& \multicolumn{2}{c|}{Satisfiability degree}\\
\cline{3-4}\cline{7-8}
&&$D^{\min}$& $D^{\prod}$ &&&$D^{\min}$& $D^{\prod}$\\
\hline
$p_1$ & $\langle a_1,0\rangle$ & $\frac{1}{{x_4}},\frac{0.5}{{x_5}},\frac{0.5}{{x_6}}$&
$\frac{1}{{x_4}},\frac{0.5}{{x_5}},\frac{0.5}{{x_6}}$&
$p_{25}$ &  $\langle a_2,2\rangle\wedge \langle a_3,1\rangle$ && \\
$p_2$ & $\langle a_1,1\rangle$ & $\frac{1}{{x_1}},\frac{1}{{x_2}},\frac{1}{{x_3}},\frac{0.5}{{x_5}},\frac{0.5}{{x_6}}$ & $\frac{1}{{x_1}},\frac{1}{{x_2}},\frac{1}{{x_3}},\frac{0.5}{{x_5}},\frac{0.5}{{x_6}}$
&$p_{26}$ & $\langle a_2,2\rangle\wedge \langle a_3,3\rangle$ & $\frac{1}{{x_1}},\frac{0.5}{{x_2}}$&$\frac{1}{{x_1}},\frac{0.5}{{x_2}}$\\
$p_3$ & $\langle a_2,1\rangle$ & $\frac{0.5}{{x_2}},\frac{1}{{x_3}},\frac{1}{{x_7}},\frac{0.333}{{x_8}}$&
$\frac{0.5}{{x_2}},\frac{1}{{x_3}},\frac{1}{{x_7}},\frac{0.333}{{x_8}}$
& $p_{27}$ & $\langle a_2,3\rangle\wedge \langle a_3,0\rangle$ & $\frac{0.333}{{x_4}},\frac{0.333}{{x_8}}$&$\frac{0.333}{{x_4}},\frac{0.333}{{x_8}}$\\
$p_4$ & $\langle a_2,2\rangle$ & $\frac{1}{{x_1}},\frac{0.5}{{x_2}},\frac{0.333}{{x_8}}$ &$\frac{1}{{x_1}},\frac{0.5}{{x_2}},\frac{0.333}{{x_8}}$&
$p_{28}$ & $\langle a_2,3\rangle\wedge \langle a_3,1\rangle $ & $\frac{0.333}{{x_4}},\frac{1}{{x_5}},\frac{0.5}{{x_6}}$&
$\frac{0.333}{{x_4}},\frac{1}{{x_5}},\frac{0.5}{{x_6}}$\\
$p_5$ &  $\langle a_2,3\rangle$ & $\frac{1}{{x_4}},\frac{1}{{x_5}},\frac{1}{{x_6}},\frac{0.333}{{x_8}}$&
$\frac{1}{{x_4}},\frac{1}{{x_5}},\frac{1}{{x_6}},\frac{0.333}{{x_8}}$&
$p_{29}$ & $\langle a_2,3\rangle\wedge \langle a_3,3\rangle $ & $\frac{0.333}{{x_4}},\frac{0.5}{{x_6}}$&
$\frac{0.333}{{x_4}},\frac{0.5}{{x_6}}$\\
$p_6$ & $\langle a_3,0\rangle$ & $\frac{0.333}{{x_4}},\frac{1}{{x_7}},\frac{1}{{x_8}}$
&$\frac{0.333}{{x_4}},\frac{1}{{x_7}},\frac{1}{{x_8}}$&
$p_{30}$ &  $\langle a_1,0\rangle\wedge \langle a_2,1\rangle\wedge \langle a_3,0\rangle $&
 & \\
$p_7$ &  $\langle a_3,1\rangle$ & $\frac{0.333}{{x_4}},\frac{1}{{x_5}},\frac{0.5}{{x_6}}$& $\frac{0.333}{{x_4}},\frac{1}{{x_5}},\frac{0.5}{{x_6}}$&
$p_{31}$ & $\langle a_1,0\rangle\wedge \langle a_2,1\rangle\wedge \langle a_3,1\rangle $ &  & \\
 $p_8$&  $\langle a_3,3\rangle$ & $\frac{1}{{x_1}},\frac{1}{{x_2}},
 \frac{1}{{x_3}},\frac{0.333}{{x_4}},\frac{0.5}{{x_6}}$
&   $\frac{1}{{x_1}},\frac{1}{{x_2}},
 \frac{1}{{x_3}},\frac{0.333}{{x_4}},\frac{0.5}{{x_6}}$&
$p_{32}$ & $\langle a_1,0\rangle\wedge \langle a_2,1\rangle\wedge \langle a_3,3\rangle $ &  & \\
$p_{9}$ & $\langle a_1,0\rangle\wedge \langle a_2,1\rangle$ &  & &
$p_{33}$ & $\langle a_1,0\rangle\wedge \langle a_2,2\rangle\wedge \langle a_3,0\rangle $ &  & \\
$p_{10}$ & $\langle a_1,0\rangle\wedge \langle a_2,2\rangle$ &  &
&$p_{34}$ & $\langle a_1,0\rangle\wedge \langle a_2,2\rangle\wedge \langle a_3,1\rangle$ &  & \\
$p_{11}$ & $\langle a_1,0\rangle\wedge \langle a_2,3\rangle$ & $\frac{1}{{x_4}},\frac{0.5}{{x_5}},\frac{0.5}{{x_6}}$&
$\frac{1}{{x_4}},\frac{0.5}{{x_5}},\frac{0.5}{{x_6}}$
&$p_{35}$ & $\langle a_1,0\rangle\wedge \langle a_2,2\rangle\wedge \langle a_3,3\rangle$ &  & \\
$p_{12}$ &$\langle a_1,0\rangle\wedge \langle a_3,0\rangle$ & $\frac{0.333}{{x_4}}$&
$\frac{0.333}{{x_4}}$
&$p_{36}$ & $\langle a_1,0\rangle\wedge \langle a_2,3\rangle\wedge \langle a_3,0\rangle$ & $\frac{0.333}{{x_4}}$&$\frac{0.333}{{x_4}}$\\
$p_{13}$ &$\langle a_1,0\rangle\wedge \langle a_3,1\rangle$ & $\frac{0.333}{{x_4}},\frac{0.5}{{x_5}},\frac{0.5}{{x_6}}$&
$\frac{0.333}{{x_4}},\frac{0.5}{{x_5}},\frac{0.25}{{x_6}}$
&$p_{37}$ & $\langle a_1,0\rangle\wedge \langle a_2,3\rangle\wedge \langle a_3,1\rangle$ & $\frac{0.333}{{x_4}},\frac{0.5}{{x_5}},\frac{0.5}{{x_6}}$
&$\frac{0.333}{{x_4}},\frac{0.5}{{x_5}},\frac{0.25}{{x_6}}$\\
$p_{14}$ & $\langle a_1,0\rangle\wedge \langle a_3,3\rangle$ & $\frac{0.333}{{x_4}},\frac{0.5}{{x_6}}$& $\frac{0.333}{{x_4}},\frac{0.25}{{x_6}}$
&$P_{38}$ & $\langle a_1,0\rangle\wedge \langle a_2,3\rangle\wedge \langle a_3,3\rangle$ & $\frac{0.333}{{x_4}},\frac{0.5}{{x_6}}$ &$\frac{0.333}{{x_4}},\frac{0.25}{{x_6}}$\\
$p_{15}$ &$\langle a_1,1\rangle\wedge \langle a_2,1\rangle$ &
$\frac{0.5}{{x_2}},\frac{1}{{x_3}}$& $\frac{0.5}{{x_2}},\frac{1}{{x_3}}$&
$p_{39}$ & $\langle a_1,1\rangle\wedge \langle a_2,1\rangle\wedge \langle a_3,0\rangle$ &   & \\
$p_{16}$ &  $\langle a_1,1\rangle\wedge \langle a_2,2\rangle $& $\frac{1}{{x_1}},\frac{0.5}{{x_2}}$& $\frac{1}{{x_1}},\frac{0.5}{{x_2}}$
&$p_{40}$ & $\langle a_1,1\rangle\wedge \langle a_2,1\rangle\wedge \langle a_3,1\rangle $ &   & \\
$p_{17}$ & $\langle a_1,1\rangle\wedge \langle a_2,3\rangle$ & $\frac{0.5}{{x_5}},\frac{0.5}{{x_6}}$&
$\frac{0.5}{{x_5}},\frac{0.5}{{x_6}}$
&$p_{41}$ & $\langle a_1,1\rangle\wedge \langle a_2,1\rangle\wedge \langle a_3,3\rangle $ & $\frac{0.5}{{x_2}},\frac{1}{{x_3}}$& $\frac{0.5}{{x_2}},\frac{1}{{x_3}}$\\
$p_{18}$& $\langle a_1,1\rangle\wedge \langle a_3,0\rangle$ &   &
&$p_{42}$ &  $\langle a_1,1\rangle\wedge \langle a_2,2\rangle\wedge \langle a_3,0\rangle$ &   &  \\
$p_{19}$ &$\langle a_1,1\rangle\wedge \langle a_3,1\rangle$ &  $\frac{0.5}{{x_5}},\frac{0.5}{{x_6}}$&
 $\frac{0.5}{{x_5}},\frac{0.25}{{x_6}}$
& $p_{43}$ & $\langle a_1,1\rangle\wedge \langle a_2,2\rangle\wedge \langle a_3,1\rangle $ &   & \\
$p_{20}$ & $\langle a_1,1\rangle\wedge \langle a_3,3\rangle$ & $\frac{1}{{x_1}},\frac{1}{{x_2}},\frac{1}{{x_3}},\frac{0.5}{{x_6}}$&
$\frac{1}{{x_1}},\frac{1}{{x_2}},\frac{1}{{x_3}},\frac{0.25}{{x_6}}$
& $p_{44}$ & $\langle a_1,1\rangle\wedge \langle a_2,2\rangle\wedge \langle a_3,3\rangle $ & $\frac{1}{{x_1}},\frac{0.5}{{x_2}}$ &$\frac{1}{{x_1}},\frac{0.5}{{x_2}}$\\
$p_{21}$ &  $\langle a_2,1\rangle\wedge \langle a_3,0\rangle$ &
$\frac{1}{{x_7}},\frac{0.333}{{x_8}}$& $\frac{1}{{x_7}},\frac{0.333}{{x_8}}$&
$p_{45}$ &  $\langle a_1,1\rangle\wedge \langle a_2,3\rangle\wedge \langle a_3,0\rangle $ &   & \\
$p_{22}$ &  $\langle a_2,1\rangle\wedge \langle a_3,1\rangle$ &
  & &
$p_{46}$ & $\langle a_1,1\rangle\wedge \langle a_2,3\rangle\wedge \langle a_3,1\rangle $ & $\frac{0.5}{{x_5}},\frac{0.5}{{x_6}}$&$\frac{0.5}{{x_5}},\frac{0.25}{{x_6}}$\\

$p_{23}$ &  $\langle a_2,1\rangle\wedge \langle a_3,3\rangle$ &
$\frac{0.5}{{x_2}},\frac{1}{{x_3}}$& $\frac{0.5}{{x_2}},\frac{1}{{x_3}}$&
$p_{47}$ &  $\langle a_1,1\rangle\wedge \langle a_2,3\rangle\wedge \langle a_3,3\rangle $ & $\frac{0.5}{{x_6}}$ & $\frac{0.25}{{x_6}}$\\
$p_{24}$ &  $\langle a_2,2\rangle\wedge \langle a_3,0\rangle$ &
$\frac{0.333}{{x_8}}$& $\frac{0.333}{{x_8}}$&
&& &\\
\hline
\end{tabular}
}
\end{table}

We demonstrate the computation for $p_{13}=\langle a_1,0\rangle\wedge \langle a_3,1\rangle = p_1 \wedge p_{7}$. The satisfiability degree of the two atomic formulas $p_1 = \langle a_1,0\rangle$ and $p_{7} = \langle a_3,1\rangle$ is computed as:
\begin{eqnarray}
D(x_1\models p_1) &=& D(x_2\models p_1) = D(x_3\models p_1) = \frac{|\{1\} \cap \{0\}|}{|\{1\}|}= 0, \nonumber\\
D(x_4\models p_1) &=& \frac{|\{0\} \cap \{0\}|}{|\{0\}|}= 1, \nonumber\\
D(x_5\models p_1) &=& D(x_6\models p_1) =  \frac{|\{0,1\} \cap \{0\}|}{|\{0,1\}|}= 0.5, \nonumber\\
D(x_7\models p_1) &=& D(x_8\models p_1) =  \frac{|\{{\rm NA}\} \cap \{0\}|}{|\{ {\rm NA}\}|}= 0;\nonumber\\
D(x_1\models p_7) &=& D(x_2\models p_7) = D(x_3\models p_7) = \frac{|\{3\} \cap \{1\}|}{|\{3\}|}= 0, \nonumber\\
D(x_4\models p_7) &=& \frac{|\{0,1,3\} \cap \{1\}|}{|\{0,1,3\}|}= \frac{1}{3} \approx 0.333, \nonumber\\
D(x_5\models p_7) &=& \frac{|\{1\} \cap \{1\}|}{|\{1\}|}= 1, \nonumber\\
D(x_6\models p_7) &=& \frac{|\{1,3\} \cap \{1\}|}{|\{1,3\}|}= 0.5, \nonumber\\
D(x_7\models p_7) &=& D(x_8\models p_7) = \frac{|\{0\} \cap \{1\}|}{|\{0\}|}= 0.
\end{eqnarray}
Then we can compute the satisfiability degree of the composite formula $p_{13}=p_1 \wedge p_7$ by using ${\rm T}^{\min}$ as:
\begin{eqnarray}
D^{\min}(x_1\models p_{13}) &=& D^{\min}(x_2\models p_{13}) = D^{\min}(x_3\models p_{13})\nonumber\\
&=& \min (D^{\min}(x_1\models p_1), D^{\min}(x_1\models p_7))= 0, \nonumber\\
D^{\min}(x_4\models p_{13}) &=& \min (D^{\min}(x_4\models p_1), D^{\min}(x_4\models p_7))= \frac{1}{3} \approx 0.333, \nonumber\\
D^{\min}(x_5\models p_{13}) &=& \min (D^{\min}(x_5\models p_1), D^{\min}(x_5\models p_7))= 0.5, \nonumber\\
D^{\min}(x_6\models p_{13}) &=& \min (D^{\min}(x_6\models p_1), D^{\min}(x_6\models p_7))= 0.5, \nonumber\\
D^{\min}(x_7\models p_{13}) &=& D^{\min}(x_8\models p_{13})\nonumber\\
&=& \min (D^{\min}(x_7\models p_1), D^{\min}(x_7\models p_7))= 0.
\end{eqnarray}
Similarly, one can compute the satisfiability degree of $p_{13}$ by using ${\rm T}^{\prod}$.

\end{example}

\subsection{Three-way decision based on $\alpha$-meaning sets of formulas}

In the conceptual formulation with a complete table, the notion of satisfiability of formulas leads to the notion of meaning sets. In the context of an incomplete table, one can apply a threshold on the satisfiability degree to get a meaning set of a formula.

\begin{definition}
In a set-valued table $S\!T=(O\!B,AT,V\cup \{{\rm NA}\},S)$, given a set of attributes $A \subseteq AT$ and a threshold $\alpha\in [0,1]$, the $\alpha$-meaning set of a formula $p \in {\rm CDL}_A$ is defined as:
\begin{equation}
m_{\alpha}(p)=\{ x\in O\!B\mid D(x\models p)\geq \alpha\}.
\end{equation}
\end{definition}

According to the definition, the $\alpha$-meaning set of a formula $p$ includes those objects that satisfy $p$ to a degree of at least $\alpha$. These objects have a probability of at least $\alpha$ to be described by $p$. The family of $\alpha$-meaning sets is used to construct the positive and negative description regions regarding a given class.

\begin{definition}
In a set-valued table $S\!T=(O\!B,AT,V\cup \{{\rm NA}\},S)$, given a set of attributes $A\subseteq AT$ and a threshold $\alpha \in [0,1]$, the positive and negative description regions regarding a class $X \subseteq O\!B$ based on $\alpha$-meaning sets are defined as:
\begin{eqnarray}
{\rm DPOS}_{(A,{\alpha})}^3(X)&=&\{p \in {\rm CDL}_A \mid m_{\alpha}(p)\neq\emptyset, m_{\alpha}(p)\subseteq X \},\nonumber\\
{\rm DNEG}_{(A,{\alpha})}^3(X)&=&\{p \in {\rm CDL}_A \mid m_{\alpha}(p)\neq\emptyset, m_{\alpha}(p)\subseteq X^c \}.
\end{eqnarray}
\end{definition}

The three-way decision rules can be easily induced from the two description regions.

\begin{definition}
In a set-valued table $S\!T=(O\!B,AT,V\cup \{{\rm NA}\},S)$, given a set of attributes $A\subseteq AT$ and a threshold $\alpha \in [0,1]$, the acceptance, rejection, and non-commitment rules for a given class $X \subseteq O\!B$ are as follows: for any $y\in O\!B$,
\begin{eqnarray*}
{\rm (A)}  & &p_a\longrightarrow_{\rm A} X:~\textit{ if}~y\models p_a~\textit{ for}~p_a\in {\rm DPOS}_{(A,{\alpha})}^3(X), ~\textit{ then~accept~}y,\\
{\rm (R)}  & &p_r\longrightarrow_{\rm R} X:~\textit{ if}~y\models p_r~\textit{ for}~p_r\in {\rm DNEG}_{(A,{\alpha})}^3(X),~\textit{ then~reject~}y,\\
{\rm (N)}  & &\textit{ If~neither~{\rm(A)}~nor~{\rm(R)}~applies,~then~neither~accept~nor~reject~}y.
\end{eqnarray*}
\end{definition}

\begin{example}

We continue with Example \ref{example:incomplete_satisfiability} to illustrate the three-way decision based on $\alpha$-meaning sets.
We apply a threshold $\alpha=0.5$ on the satisfiability degrees in Table~\ref{tab:example_incomplete_meaning_sets} and get the $0.5$-meaning sets of formulas given in Table~\ref{tab:example_incomplete_0.5-meaning_sets}.

\begin{table}[!ht]
\centering
\caption{The 0.5-meaning sets of all formulas for Table~\ref{tab:example_set-valued_table_duplicate}}
\label{tab:example_incomplete_0.5-meaning_sets}
\renewcommand{\arraystretch}{1.3}
\scalebox{0.8}{
\begin{tabular}{|c|c|c|c||c|c|c|c|}
\hline
{\multirow{2}{*}{Label}}&{\multirow{2}{*}{Formula}}& \multicolumn{2}{c||}{0.5-meaning sets}
&{\multirow{2}{*}{Label}}&{\multirow{2}{*}{Formula}}& \multicolumn{2}{c|}{0.5-meaning sets}\\
\cline{3-4}\cline{7-8}
&&$m^{\min}_{0.5}$& $m^{\prod}_{0.5}$ &&&$m^{\min}_{0.5}$& $m^{\prod}_{0.5}$\\
\hline
$p_1$ & $\langle a_1,0\rangle$ & $\{x_4,x_5,x_6\}$&$\{x_4,x_5,x_6\}$&
$p_{25}$ &  $\langle a_2,2\rangle\wedge \langle a_3,1\rangle$ &
$\emptyset$ &$\emptyset$\\
$p_2$ & $\langle a_1,1\rangle$ & $\{x_1,x_2,x_3,x_5,x_6\}$ & $\{x_1,x_2,x_3,x_5,x_6\}$
&$p_{26}$ & $\langle a_2,2\rangle\wedge \langle a_3,3\rangle$ & $\{x_1,x_2\}$&$\{x_1,x_2\}$\\
$p_3$ & $\langle a_2,1\rangle$ & $\{x_2,x_3,x_7\}$&$\{x_2,x_3,x_7\}$&
$p_{27}$ & $\langle a_2,3\rangle\wedge \langle a_3,0\rangle$ & $\emptyset$&$\emptyset$\\
$p_4$ & $\langle a_2,2\rangle$ & $\{x_1,x_2\}$ &$\{x_1,x_2\}$&
$p_{28}$ & $\langle a_2,3\rangle\wedge \langle a_3,1\rangle $ & $\{x_5,x_6\}$&$\{x_5,x_6\}$\\
$p_5$ &  $\langle a_2,3\rangle$ & $\{x_4,x_5,x_6\}$&$\{x_4,x_5,x_6\}$&
$p_{29}$ & $\langle a_2,3\rangle\wedge \langle a_3,3\rangle $ & $\{x_6\}$&$\{x_6\}$\\
$p_6$ & $\langle a_3,0\rangle$ & $\{x_7,x_8\}$&$\{x_7,x_8\}$&
$p_{30}$ &  $\langle a_1,0\rangle\wedge \langle a_2,1\rangle\wedge \langle a_3,0\rangle $&
$\emptyset$&$\emptyset$\\
$p_7$ &  $\langle a_3,1\rangle$ & $\{x_5,x_6\}$& $\{x_5,x_6\}$&
$p_{31}$ & $\langle a_1,0\rangle\wedge \langle a_2,1\rangle\wedge \langle a_3,1\rangle $ & $\emptyset$&$\emptyset$\\
$p_8$&  $\langle a_3,3\rangle$ & $\{x_1,x_2,x_3,x_6\}$&   $\{x_1,x_2,x_3,x_6\}$&
$p_{32}$ & $\langle a_1,0\rangle\wedge \langle a_2,1\rangle\wedge \langle a_3,3\rangle $ & $\emptyset$&$\emptyset$\\
$p_{9}$ & $\langle a_1,0\rangle\wedge \langle a_2,1\rangle$ & $\emptyset$&$\emptyset$&
$p_{33}$ & $\langle a_1,0\rangle\wedge \langle a_2,2\rangle\wedge \langle a_3,0\rangle $ & $\emptyset$&$\emptyset$\\
$p_{10}$ & $\langle a_1,0\rangle\wedge \langle a_2,2\rangle$ & $\emptyset$&$\emptyset$&
$p_{34}$ & $\langle a_1,0\rangle\wedge \langle a_2,2\rangle\wedge \langle a_3,1\rangle$ & $\emptyset$&$\emptyset$\\
$p_{11}$ & $\langle a_1,0\rangle\wedge \langle a_2,3\rangle$ & $\{x_4,x_5,x_6\}$&$\{x_4,x_5,x_6\}$
&$p_{35}$ & $\langle a_1,0\rangle\wedge \langle a_2,2\rangle\wedge \langle a_3,3\rangle$ & $\emptyset$&$\emptyset$\\
$p_{12}$ &$\langle a_1,0\rangle\wedge \langle a_3,0\rangle$ & $\emptyset$&
$\emptyset$
&$p_{36}$ & $\langle a_1,0\rangle\wedge \langle a_2,3\rangle\wedge \langle a_3,0\rangle$ & $\emptyset$&$\emptyset$\\
$p_{13}$ &$\langle a_1,0\rangle\wedge \langle a_3,1\rangle$ & $\{x_5,x_6\}$&$\{x_5\}$
&$p_{37}$ & $\langle a_1,0\rangle\wedge \langle a_2,3\rangle\wedge \langle a_3,1\rangle$ & $\{x_5,x_6\}$&$\{x_5\}$\\
$p_{14}$ & $\langle a_1,0\rangle\wedge \langle a_3,3\rangle$ & $\{x_6\}$& $\emptyset$
&$p_{38}$ & $\langle a_1,0\rangle\wedge \langle a_2,3\rangle\wedge \langle a_3,3\rangle$ & $\{x_6\}$ &$\emptyset$\\
$p_{15}$ &$\langle a_1,1\rangle\wedge \langle a_2,1\rangle$ &$\{x_2,x_3\}$& $\{x_2,x_3\}$&
$p_{39}$ & $\langle a_1,1\rangle\wedge \langle a_2,1\rangle\wedge \langle a_3,0\rangle$ & $\emptyset$ &$\emptyset$\\
$p_{16}$ &  $\langle a_1,1\rangle\wedge \langle a_2,2\rangle $& $\{x_1,x_2\}$& $\{x_1,x_2\}$
&$p_{40}$ & $\langle a_1,1\rangle\wedge \langle a_2,1\rangle\wedge \langle a_3,1\rangle $ & $\emptyset$ &$\emptyset$\\
$p_{17}$ & $\langle a_1,1\rangle\wedge \langle a_2,3\rangle$ & $\{x_5,x_6\}$&$\{x_5,x_6\}$
&$p_{41}$ & $\langle a_1,1\rangle\wedge \langle a_2,1\rangle\wedge \langle a_3,3\rangle $ & $\{x_2,x_3\}$& $\{x_2,x_3\}$\\
$p_{18}$& $\langle a_1,1\rangle\wedge \langle a_3,0\rangle$ & $\emptyset$ &$\emptyset$
&$p_{42}$ &  $\langle a_1,1\rangle\wedge \langle a_2,2\rangle\wedge \langle a_3,0\rangle$ & $\emptyset$ &$\emptyset$ \\
$p_{19}$ &$\langle a_1,1\rangle\wedge \langle a_3,1\rangle$ &  $\{x_5,x_6\}$& $\{x_5\}$
&$p_{43}$ & $\langle a_1,1\rangle\wedge \langle a_2,2\rangle\wedge \langle a_3,1\rangle $ & $\emptyset$ &$\emptyset$\\
$p_{20}$ & $\langle a_1,1\rangle\wedge \langle a_3,3\rangle$ & $\{x_1,x_2,x_3,x_6\}$&$\{x_1,x_2,x_3\}$
&$p_{44}$ & $\langle a_1,1\rangle\wedge \langle a_2,2\rangle\wedge \langle a_3,3\rangle $ & $\{x_1,x_2\}$ &$\{x_1,x_2\}$\\
$p_{21}$ &  $\langle a_2,1\rangle\wedge \langle a_3,0\rangle$ &$\{x_7\}$& $\{x_7\}$&
$p_{45}$ &  $\langle a_1,1\rangle\wedge \langle a_2,3\rangle\wedge \langle a_3,0\rangle $ & $\emptyset$ &$\emptyset$\\
$p_{22}$ &  $\langle a_2,1\rangle\wedge \langle a_3,1\rangle$ &
$\emptyset$ &$\emptyset$&
$p_{46}$ & $\langle a_1,1\rangle\wedge \langle a_2,3\rangle\wedge \langle a_3,1\rangle $ & $\{x_5,x_6\}$& $\{x_5\}$\\
$p_{23}$ &  $\langle a_2,1\rangle\wedge \langle a_3,3\rangle$ &$\{x_2,x_3\}$& $\{x_2,x_3\}$&
$p_{47}$ &  $\langle a_1,1\rangle\wedge \langle a_2,3\rangle\wedge \langle a_3,3\rangle $ & $\{x_6\}$ & $\emptyset$\\
$p_{24}$ &  $\langle a_2,2\rangle\wedge \langle a_3,0\rangle$ &$\emptyset$& $\emptyset$
&&& &\\
\hline
\end{tabular}
}
\end{table}

For a given class $X = \{x_1,x_2,x_3,x_4\}$, the positive and negative description regions can be constructed through the $0.5$-meaning sets computed by either ${\rm T}^{\min}$ or ${\rm T}^{\prod}$. A consideration of ${\rm T}^{\min}$ leads to the following regions:
\begin{eqnarray}
{\rm DPOS}_{(AT,{0.5})}^{3~\min}(X)&=&\{p \in {\rm CDL}_{AT}\mid m_{0.5}^{\min}(p)\neq\emptyset, m_{0.5}^{\min}(p)\subseteq X \}\nonumber\\
&=&\{p_4,p_{15},p_{16},p_{23},p_{26},p_{41},p_{44}\},\nonumber\\
{\rm DNEG}_{(AT,{0.5})}^{3~\min}(X)&=&\{p \in {\rm CDL}_{AT}\mid m_{0.5}^{\min}(p)\neq\emptyset, m_{0.5}^{\min}(p)\subseteq X^c\}\nonumber\\
&=&\{p_6,p_7,p_{13},p_{14},p_{17},p_{19},p_{21},p_{28},p_{29},p_{37},p_{38},p_{46},p_{47}\}.
\end{eqnarray}
If we consider ${\rm T}^{\prod}$, we can get the following two regions:
\begin{eqnarray}
{\rm DPOS}_{(AT,{0.5})}^{3~\prod}(X)&=&\{p \in {\rm CDL}_{AT}\mid m_{0.5}^{\prod}(p)\neq\emptyset, m_{0.5}^{\prod}(p)\subseteq X \}\nonumber\\
&=&\{p_4,p_{15},p_{16},p_{20},p_{23},p_{26},p_{41},p_{44}\},\nonumber\\
{\rm DNEG}_{(AT,{0.5})}^{3~\prod}(X)&=&\{p \in {\rm CDL}_{AT}\mid m_{0.5}^{\prod}(p)\neq\emptyset, m_{0.5}^{\prod}(p)\subseteq X^c\}\nonumber\\
&=&\{p_6,p_7,p_{13},p_{17},p_{19},p_{21},p_{28},p_{29},p_{37},p_{46}\}.
\end{eqnarray}
Since the rule induction from the above description regions is very straightforward, we omit the three-way decision rules for brevity.
\end{example}

\subsection{Three-way decision based on confidence of formulas}

In this section, we measure the degree to which a formula can be used to form an acceptance or a rejection rule based on the satisfiability degree. The two degrees are defined as the acceptance and rejection confidence of formulas. The description regions can be accordingly constructed by applying a threshold.

For a given class $X \subseteq O\!B$, there are three possibilities of a three-way decision rule that can be formed by a particular formula $p \in {\rm CDL}_A$ where $A \subseteq AT$, that is, an acceptance rule $p \longrightarrow_{\rm A} X$, a rejection rule $p \longrightarrow_{\rm R} X$, or a non-commitment rule $p \longrightarrow_{\rm N} X$. To make clearer interpretations of three-way decision rules, we adopt the work by Deng, Yao, and Yao~\cite{deng2014on} that interprets a three-way decision model by two two-way decision models. A first two-way decision model $({\rm A}^\prime, {\rm \bar A}^\prime)$ makes either an acceptance $({\rm A}^\prime)$ or a non-acceptance $({\rm \bar A}^\prime)$ decision regarding a given class $X \subseteq O\!B$ by rules in the following format: for any $y \in O\!B$,
\begin{eqnarray}
({\rm A}^\prime) && p_a \longrightarrow_{{\rm A}^\prime} X: ~{\rm if}~y\models p_{a}, ~{\rm then~accept~}y,\nonumber\\
({\rm \bar A}^\prime) && p_{na} \longrightarrow_{{\rm \bar A}^\prime} X: ~{\rm if}~y\models p_{na}, ~{\rm then~do~not~accept~}y.
\end{eqnarray}
The two decisions are dual, that is, a non-acceptance rule $p \longrightarrow_{{\rm \bar A}^\prime} X$ can be equivalently expressed as $\neg (p \longrightarrow_{{\rm A}^\prime} X)$ and vice versa. Similarly, a second two-way decision model $({\rm R}^\prime, {\rm \bar R}^\prime)$ makes either a rejection $({\rm R}^\prime)$ or a non-rejection $({\rm \bar R}^\prime)$ decision by rules in the following format: for any $y \in O\!B$,
\begin{eqnarray}
({\rm R}^\prime) && p_r \longrightarrow_{{\rm R}^\prime} X: ~{\rm if}~y\models p_r, ~{\rm then~reject~}y,\nonumber\\
({\rm \bar R}^\prime) && p_{nr} \longrightarrow_{{\rm \bar R}^\prime} X: ~{\rm if}~y\models p_{nr}, ~{\rm then~do~not~reject~}y,
\end{eqnarray}
where a non-rejection rule $p \longrightarrow_{{\rm \bar R}^\prime} X$ can be equivalently expressed as $\neg (p \longrightarrow_{{\rm R}^\prime} X)$ and vice versa. The three-way decisions of acceptance, rejection, and non-commitment are made as combinations of two decisions from the two two-way decision models, as given in Table \ref{tab:two-way_three-way}.

\begin{table}[!ht]
\begin{center}
\caption{Interpreting three-way decisions through two two-way decisions}
\label{tab:two-way_three-way}
\renewcommand{\arraystretch}{2}
\begin{tabular}{|c|c|c|}
\hline
\backslashbox{$({\rm A}^\prime, {\rm \bar A}^\prime)$-model}{$({\rm R}^\prime, {\rm \bar R}^\prime)$-model} & Rejection $({\rm R}^\prime)$& Non-rejection $({\rm \bar R}^\prime)$\\
\hline
Acceptance $({\rm A}^\prime)$ & Non-commitment (N) & Acceptance (A) \\
Non-acceptance $({\rm \bar A}^\prime)$ & Rejection (R) & Non-commitment (N) \\
\hline
\end{tabular}
\end{center}
\end{table}

An acceptance rule $p \longrightarrow_{\rm A} X$ in the three-way decision model is interpreted as a conjunction of an acceptance rule $({\rm A}^\prime)$ and a non-rejection rule $({\rm \bar R}^\prime)$, that is:
\begin{eqnarray}
p \longrightarrow_{\rm A} X & \Longleftrightarrow & (p \longrightarrow_{{\rm A}^\prime} X) \wedge (p \longrightarrow_{{\rm \bar R}^\prime} X)\nonumber\\
&\Longleftrightarrow & (p \longrightarrow_{{\rm A}^\prime} X) \wedge \neg (p \longrightarrow_{{\rm R}^\prime} X).
\end{eqnarray}
In the context of complete information, it follows that:
\begin{eqnarray}
\label{equa:acceptance_non-rejection}
p \longrightarrow_{\rm A} X &\Longleftrightarrow & (\forall\, x \in O\!B, x \models p \longrightarrow x \in X) \wedge \neg (\forall\, x \in O\!B,  x \models p \longrightarrow x \in X^c)\nonumber\\
&\Longleftrightarrow & \big(\bigwedge_{x \in O\!B} (x \models p \longrightarrow x \in X)\big) \wedge \neg \big(\bigwedge_{x \in O\!B}  (x \models p \longrightarrow x \in X^c)\big).
\end{eqnarray}
Following our discussion in Section \ref{sec:approximability}, the above conditions can be generalized into the context of incomplete information by:
\begin{enumerate}[label=(\arabic*)]
\item using the satisfiability degree $D(x \models p)$ to generalize the satisfiability $x \models p$;
\item using the indicator functions $1_X$ and $1_{X^c}$ to generalize the inclusion relationships $x \in X$ and $x \in X^c$;
\item using a T-norm to generalize the classical logic conjunction;
\item using a fuzzy implication ${\rm I}$ to generalize the classical logic implication;
\item using a fuzzy negation ${\rm N}$ to generalize the classical logic negation.
\end{enumerate}

A fuzzy negation, also called a negator, is defined as a decreasing mapping ${\rm N}:[0,1]\longrightarrow [0,1]$ satisfying the two conditions ${\rm N}(0)=1$ and ${\rm N}(1)=0$. A commonly used negator is ${\rm N}(x)=1-x$, which is usually referred to as the standard negator. As a result, the condition $(\bigwedge_{x \in O\!B} x \models p \longrightarrow x \in X)$ in Equation (\ref{equa:acceptance_non-rejection}) is generalized with respect to incomplete information into:
\begin{eqnarray}
{\rm T}_{x \in O\!B}\big({\rm I}(D(x \models p), 1_X(x))\big).
\end{eqnarray}
Similarly, the condition $\neg (\bigwedge_{x \in O\!B}  x \models p \longrightarrow x \in X^c)$ is generalized into:
\begin{eqnarray}
{\rm N}\big({\rm T}_{x \in O\!B}({\rm I}(D(x \models p), 1_{X^c}(x)))\big).
\end{eqnarray}
The conjunction of these two conditions is generalized by a T-norm. Thus, we finally generalize the entire condition in Equation (\ref{equa:acceptance_non-rejection}) into:
\begin{eqnarray}
{\rm T}\big( {\rm T}_{x \in O\!B}\big({\rm I}(D(x \models p), 1_X(x))\big), {\rm N}\big({\rm T}_{x \in O\!B}({\rm I}(D(x \models p), 1_{X^c}(x)))\big) \big),
\end{eqnarray}
which is considered as the acceptance confidence of a formula $p \in {\rm CDL}_A$, that is, the confidence of $p$ inducing an acceptance rule $p \longrightarrow_{\rm A} X$. One can apply the same analysis on rejection rule and compute the rejection confidence of $p$.

\begin{definition}
In a set-valued table $S\!T=(O\!B,AT,V\cup \{{\rm NA}\},S)$, given a set of attributes $A\subseteq AT$ and a class $X \subseteq O\!B$, the acceptance confidence ${\rm AC}(p)$ and rejection confidence ${\rm RC}(p)$ of a formula $p \in {\rm CDL}_A$ are defined as:
\begin{eqnarray}
{\rm AC}(p) &=& {\rm T}\big( {\rm T}_{x \in O\!B}\big({\rm I}(D(x \models p), 1_X(x))\big), {\rm N}\big({\rm T}_{x \in O\!B}({\rm I}(D(x \models p), 1_{X^c}(x)))\big) \big),\nonumber\\
{\rm RC}(p) &=& {\rm T}\big( {\rm T}_{x \in O\!B}\big({\rm I}(D(x \models p), 1_{X^c}(x))\big), {\rm N}\big({\rm T}_{x \in O\!B}({\rm I}(D(x \models p), 1_X(x)))\big)\big).
\end{eqnarray}
\end{definition}

The positive and negative description regions can be constructed by applying a threshold on the confidence of formulas.

\begin{definition}
In a set-valued table $S\!T=(O\!B,AT,V\cup \{{\rm NA}\},S)$, given a set of attributes $A\subseteq AT$ and a class $X \subseteq O\!B$, the positive and negative description regions are defined as:
\begin{eqnarray}
{\rm DPOS}_{(A,\alpha)}^4(X)&=&\{p\in {\rm CDL}_A\mid {\rm AC}(p)\geq \alpha \},\nonumber\\
{\rm DNEG}_{(A,\alpha)}^4(X)&=&\{p\in {\rm CDL}_A\mid {\rm RC}(p)\geq\alpha\},
\end{eqnarray}
where $\alpha\in [0,1]$ is a threshold.
\end{definition}

Depending on the choice of the threshold $\alpha$, the two regions ${\rm DPOS}_{(A,\alpha)}^4(X)$ and ${\rm DNEG}_{(A,\alpha)}^4(X)$ may have nonempty overlap. In this situation, we induce non-commitment rules.

\begin{definition}
In a set-valued table $S\!T=(O\!B,AT,V\cup \{{\rm NA}\},S)$, given a set of attributes $A\subseteq AT$ and a threshold $\alpha \in [0,1]$, the acceptance, rejection, and non-commitment rules for a given class $X \subseteq O\!B$ are as follows: for any $y\in O\!B$,
\begin{eqnarray*}
{\rm (A)}  & &p_a\longrightarrow_{\rm A} X:~\textit{ if}~y\models p_a~\textit{ for}~p_a\in {\rm DPOS}_{(A,{\alpha})}^4(X) - {\rm DNEG}_{(A,{\alpha})}^4(X), ~\textit{ then~accept~}y,\\
{\rm (R)}  & &p_r\longrightarrow_{\rm R} X:~\textit{ if}~y\models p_r~\textit{ for}~p_r\in {\rm DNEG}_{(A,{\alpha})}^4(X) - {\rm DPOS}_{(A,\alpha)}^4(X),~\textit{ then~reject~}y,\\
{\rm (N)}  & &\textit{ If~neither~{\rm(A)}~nor~{\rm(R)}~applies,~then~neither~accept~nor~reject~}y.
\end{eqnarray*}
\end{definition}

To apply the above approach, one need to specify the T-norm, negator, and fuzzy implication in order to calculate ${\rm AC}(p)$ and ${\rm RC}(p)$. Suppose we use the standard negator ${\rm N}(x)=1-x$ and follow the correspondences between T-norms and fuzzy implications discussed after Definition \ref{def:approximability}. If ${\rm T}^{\min}$ and ${\rm I_{KD}}$ are used, we have:
\begin{eqnarray}
\label{equa:ACRC_min}
{\rm AC}^{\min}(p)&=&{\rm T}^{\min}\big({\rm T}_{x\in O\!B}^{\min}\big({\rm I_{KD}}(D(x\models p),1_X(x))\big),{\rm N}\big({\rm T}_{x\in O\!B}^{\min}({\rm I_{KD}}(D(x\models p),1_{X^c}(x)))\big)\big)\nonumber\\
&=&\min\big(\min_{x\in O\!B}\big(\max(1-D(x\models p),1_X(x))\big), 1-\min_{x\in O\!B}\big(\max(1-D(x\models p),1_{X^c}(x))\big)\big)\nonumber\\
&=&\min\big(\min\big(\min_{x \in X}(1), \min_{x \in X^c}(1 - D(x \models p))\big),  1 - \min \big(\min_{x \in X}(1 - D(x \models p)), \min_{x \in X^c}(1)\big) \big) \nonumber\\
&=&\min\big(\min_{x\in {X^c}}(1-D(x\models p)),
1-\min_{x\in X}(1-D(x\models p))\big)\nonumber\\
&=&\min\big(1-\max_{x\in {X^c}}D(x\models p),
\max_{x\in X}D(x\models p)\big),\nonumber\\
{\rm RC}^{\min}(p)&=&{\rm T}^{\min}\big({\rm T}_{x\in O\!B}^{\min}\big({\rm I_{KD}}(D(x\models p),1_{X^c}(x))\big),{\rm N}\big({\rm T}_{x\in O\!B}^{\min}({\rm I_{KD}}(D(x\models p),1_{X}(x)))\big)\big)\nonumber\\
&=&\min\big(\min_{x\in O\!B}\big(\max(1-D(x\models p),1_{X^c}(x))\big), 1-\min_{x\in O\!B}\big(\max(1-D(x\models p),1_{X}(x))\big)\big)\nonumber\\
&=&\min\big(\min\big(\min_{x \in X}(1 - D(x \models p)), \min_{x \in X^c}(1)\big),  1 - \min \big(\min_{x \in X}(1), \min_{x \in X^c}(1 - D(x \models p))\big) \big) \nonumber\\
&=&\min\big(\min_{x\in {X}}(1-D(x\models p)),
1-\min_{x\in {X^c}}(1-D(x\models p))\big)\nonumber\\
&=&\min\big(1-\max_{x\in{X}}D(x\models p),
\max_{x\in{X^c}}D(x\models p)\big).
\end{eqnarray}
If ${\rm T}^{\prod}$ and ${\rm I_{RC}}$ are used, we have:
\begin{eqnarray}
\label{equa:ACRC_prod}
{\rm AC}^{\prod}(p)&=&{\rm T}^{\prod}\big({\rm T}_{x\in O\!B}^{\prod}({\rm I_{RC}}(D(x\models p),1_X(x))),{\rm N}({\rm T}_{x\in O\!B}^{\prod}({\rm I_{RC}}(D(x\models p),1_{X^c}(x))))\big)\nonumber\\
&=&\big(\prod_{x\in O\!B}(1-D(x\models p)+D(x\models p)\cdot1_X(x))\big)\cdot \big(1-\prod_{x\in O\!B}(1-D(x\models p)+D(x\models p)\cdot1_{X^c}(x))\big)\nonumber\\
&=&\big(\prod_{x\in {X^c}}(1-D(x\models p))\big)\cdot
\big(1-\prod_{x\in X}(1-D(x\models p))\big), \nonumber\\
{\rm RC}^{\prod}(p)&=&{\rm T}^{\prod}\big({\rm T}_{x\in O\!B}^{\prod}({\rm I_{RC}}(D(x\models p),1_{X^c}(x))),{\rm N}({\rm T}_{x\in O\!B}^{\prod}({\rm I_{RC}}(D(x\models p),1_{X}(x))))\big)\nonumber\\
&=&\big(\prod_{x\in O\!B}(1-D(x\models p)+D(x\models p)\cdot1_{X^c}(x))\big)\cdot \big(1-\prod_{x\in O\!B}(1-D(x\models p)+D(x\models p)\cdot1_{X}(x))\big)\nonumber\\
&=&\big(\prod_{x\in {X}}(1-D(x\models p))\big)\cdot
\big(1-\prod_{x\in {X^c}}(1-D(x\models p))\big).
\end{eqnarray}

\begin{example}
We illustrate the above approach to three-way decision by using the satisfiability degree calculated in Example~\ref{example:incomplete_satisfiability}. Suppose the given class is $X = \{x_1,x_2,x_3,x_4\}$. By using Equations (\ref{equa:ACRC_min}) and (\ref{equa:ACRC_prod}), we calculate the acceptance and rejection confidence of formulas in ${\rm CDL}_{AT}$, which is given in Table~\ref{tab:example_incomplete_rule_confidence}.

\begin{table}[!ht]
\centering
\caption{Confidence of formulas in ${\rm CDL}_{A}$ for Table \ref{tab:example_set-valued_table_duplicate}}
\label{tab:example_incomplete_rule_confidence}
\scalebox{0.8}{
\renewcommand{\arraystretch}{1.2}
\begin{tabular}{|c|c|c|c|c||c|c|c|c|c||c|c|c|c|c|}
\hline
Formula & ${\rm AC}^{\min}$& ${\rm RC}^{\min}$  &${\rm AC}^{\prod}$& ${\rm RC}^{\prod}$& Formula & ${\rm AC}^{\min}$&${\rm RC}^{\min}$ & ${\rm AC}^{\prod}$ & ${\rm RC}^{\prod}$& Formula & ${\rm AC}^{\min}$&${\rm RC}^{\min}$ & ${\rm AC}^{\prod}$ & ${\rm RC}^{\prod}$\\
\hline
$p_1$&0.5&0&0.25&0          		&
$p_{17}$&0&0.5&0&0.75        		&
$p_{33}$&0&0&0&0					\\
$p_2$&0.5&0&0.25&0            		&
$p_{18}$&0&0&0&0             		&
$p_{34}$&0&0&0&0					\\
$p_3$&0&0&0&0                		&
$p_{19}$&0&0.5&0&0.625 		&
$p_{35}$&0&0&0&0					\\
$p_4$&0.667&0  &0.667&0      		&
$p_{20}$&0.5&0&0.75&0        		&
$p_{36}$&0.333&0&0.333&0\\	
$p_5$&0&0&0&0                		&
$p_{21}$&0&1&0&1					&
$p_{37}$&0.333&0.5&0.125&0.417\\
$p_6$&0&0.667&0&0.667        		&
$p_{22}$&0&0&0&0 					&
$p_{38}$&0.333&0.5&0.250&0.167\\
$p_7$&0&0.667&0&0.667    		& 
$p_{23}$&1&0&1&0 					&
$p_{39}$&0&0&0&0\\
$p_8$&0.5&0&0.5&0            		&
$p_{24}$&0&0.333&0&0.333 				&
$p_{40}$&0&0&0&0\\
$p_{9}$&0&0&0&0             		&
$p_{25}$&0&0&0&0					& 
$p_{41}$&1&0&1&0\\
$p_{10}$&0&0&0&0             		&
$p_{26}$&1&0&1&0  					&
$p_{42}$&0&0&0&0\\
$p_{11}$&0.5&0&0.25&0         		& 
$p_{27}$&0.333&0.333&0.222&0.222    &
$p_{43}$&0&0&0&0\\
$p_{12}$&0.333&0&0.333&0 		    &
$p_{28}$&0&0.667&0&0.667			&
$p_{44}$&1&0&1&0\\
$p_{13}$&0.333&0.5&0.125& 0.417	    &
$p_{29}$&0.333&0.5&0.167&0.333		&
$p_{45}$&0&0&0&0\\
$p_{14}$&0.333&0.5&0.250& 0.167	&
$p_{30}$&0&0&0&0					&
$p_{46}$&0&0.5&0&0.625\\
$p_{15}$&1&0&1&0             		&
$p_{31}$&0&0&0&0				 	&
$p_{47}$&0&0.5&0&0.25\\
$p_{16}$&1&0&1&0            		&
$p_{32}$&0&0&0&0	                &
&&&&\\
\hline
\end{tabular}
}
\end{table}

Suppose we apply a threshold $\alpha=0.6$, the positive and negative description regions based on ${\rm T}^{\min}$ are:
\begin{eqnarray}
{\rm DPOS}_{(AT,0.6)}^{4~\min}(X)&=&\{p\in {\rm CDL}_{AT}\mid {\rm AC}^{\min}(p)\geq 0.6\}\nonumber\\
&=&\{p_4,p_{15},p_{16},p_{23},p_{26},p_{41},p_{44}\},\nonumber\\
{\rm DNEG}_{(AT,0.6)}^{4~\min}(X)&=&\{p\in {\rm CDL}_{AT}\mid {\rm RC}^{\min}(p)\geq 0.6\}\nonumber\\
&=&\{p_6,p_7,p_{21},p_{28}\}.
\end{eqnarray}
And the two description regions based on ${\rm T}^{\prod}$ are:
\begin{eqnarray}
{\rm DPOS}_{(AT,0.6)}^{4~\prod}(X)&=&\{p\in {\rm CDL}_{AT}\mid {\rm AC}^{\prod}(p)\geq 0.6\}\nonumber\\
&=&\{p_4,p_{15},p_{16},p_{20},p_{23},p_{26},p_{41},p_{44}\},\nonumber\\
{\rm DNEG}_{(AT,0.6)}^{4~\prod}(X)&=&\{p\in {\rm CDL}_{AT}\mid {\rm RC}^{\prod}(p)\geq 0.6\}\nonumber\\
&=&\{p_6,p_7,p_{17},p_{19},p_{21},p_{28},p_{46}\}.
\end{eqnarray}
Since the rule induction from the above description regions is very straightforward, we omit the three-way decision rules for brevity. If a threshold $\alpha=0.3$ is applied, then the formulas $p_{13}, p_{14}, p_{27}, p_{29}, p_{37}$, and $p_{38}$ will be included in both the positive description region ${\rm DPOS}_{(AT,0.6)}^{4~\min}(X)$ and the negative description region ${\rm DNEG}_{(AT,0.6)}^{4~\min}(X)$, which results in non-commitment decisions due to the conflict.
\end{example}

It should be noted that in the conceptual formulation, the value ${\rm NA}$ is not considered and a pair $\langle a,{\rm NA}\rangle$ where $a \in AT$ is not included in any decision rule derived. This is because we consider the description space to be the conjunctive description language ${\rm CDL}_A$ defined in Definition~\ref{def:CDL} where a condition $v \in V_a$ is used (${\rm NA} \not\in V_a$). As we discussed after Example \ref{example:incomplete_approximability}, whether a pair $\langle a,{\rm NA}\rangle$ should be included in a decision rule or not depends on the semantics of the specific attribute $a$. To accommodate a meaningful ${\rm NA}$ in the conceptual formulation, one may need to generalize the definition of the conjunctive description language, which will be one of the directions for the future work.

Based on the satisfiability degree of formulas, we discussed two approaches to the conceptual formulation of three-way decision with incomplete information. One approach applies a threshold $\alpha$ to get $\alpha$-meaning sets of formulas, and another approach calculates the acceptance and rejection confidence of formulas. Compared with the computational formulation discussed in Section \ref{sec:computational_incomplete}, the conceptual formulation explicitly analyzes the descriptions (i.e., the conjunctive formulas in our approaches). Thus, the conceptual formulation provides clearer interpretation and better understanding of decision rules. However, the computation is not as efficient as the computational formulation.

The approaches discussed in the computational and conceptual formulations are closely related. The approach based on $\alpha$-similarity classes in the computational formulation and the approach based on $\alpha$-meaning sets of formulas take similar strategies. They both apply a threshold to convert either a quantitative similarity degree or a quantitative satisfiability degree into a qualitative inclusion relation of an object to a set (i.e., either an $\alpha$-similarity class or an $\alpha$-meaning set). The approach based on approximability of objects in the computational formulation and the approached based on confidence of formulas in the conceptual formulation also follow very similar ideas. They both apply fuzzy logic operators (i.e., T-norm, fuzzy implication, and fuzzy negation) to estimate the degree to which an acceptance or a rejection decision can be induced. While the two approaches based on $\alpha$-similarity classes and $\alpha$-meaning sets apply a threshold when constructing definable building blocks, the two approaches based on approximability and confidence use a threshold when constructing approximations (i.e., description regions). More detailed theoretical and experimental comparison of these approaches and investigation of their relationships will be studied in our future work.

\section{Conclusions and future work}
\label{sec:conclusions}

The computational and conceptual formulations are two complementary sides of three-way decision in rough set theory. A review of these two formulations with complete information shows two basic concepts, that is, equivalence relations in the computational formulation and satisfiability of formulas in the conceptual formulation. By considering generalizations of these two basic concepts, we discuss the corresponding two formulations of three-way decision with incomplete information. A computational formulation generalizes equivalence relations into similarity degree of objects, and a conceptual formulation generalizes satisfiability of formulas into quantitative satisfiability degree. The similarity degree leads to two approaches of the computational formulation, including one approach using $\alpha$-similarity classes and another using approximability of objects. The quantitative satisfiability degree leads to two approaches of the conceptual formulation, including one approach using $\alpha$-meaning sets and another using acceptance and rejection confidence. In all these approaches, three-way decision rules can be easily derived from the description regions.

Several directions of future work are already pointed out along our discussion, including a first direction of removing redundant attributes and decision rules, a second direction of determining the thresholds, a third direction of using subsethood measures to generalize the subset relationships in defining description regions, a fourth direction of experimental comparison and analysis of the proposed approaches with incomplete information, and a fifth direction of studying the meaningfulness of the special value ${\rm NA}$. In addition, one may consider further generalizations of the proposed approaches. For instance, our work assumes that each value in the set $s_a(x)$ is evenly likely to be the actual value, which can be generalized into a probability distribution. This will lead to new definitions of similarity degree of objects and satisfiability degree of formulas. Correspondingly, one may investigate generalized approaches in the computational and conceptual formulations.

\section*{Acknowledgement}
The authors thank reviewers and editors for their constructive comments. This work is supported by the National Natural Science Foundation of China (Grant Nos. 61473239, 61976130) and the China Scholarship Council (Grant No. 201707000052). The authors are grateful to Dr. Yiyu Yao for his insightful suggestions.

\appendix
\gdef\thesection{Appendix}
\section{Proof of Theorem \ref{theo:equivalent_Boolean_algebras}}
\label{appendix:equivalent_Boolean_algebras}

\begin{proof}
We prove the following two subset relationships: (1) $B(O\!B/E_A) \subseteq B({\rm CDEF}_A(T))$; (2) $B(O\!B/E_A) \supseteq B({\rm CDEF}_A(T))$.
\begin{enumerate}[label=(\arabic*)]
\item Prove $B(O\!B/E_A) \subseteq B({\rm CDEF}_A(T))$.\\
This can be immediately proved by Theorem \ref{theo:subset_relationship} which states $O\!B/E_A \subseteq {\rm CDEF}_A(T)$. A union of an arbitrary number of sets from $O\!B/E_A$ must also be a union of the same group of sets from ${\rm CDEF}_A(T)$, that is, $B(O\!B/E_A) \subseteq B({\rm CDEF}_A(T))$.
\item Prove $B(O\!B/E_A) \supseteq B({\rm CDEF}_A(T))$.\\
We first prove that any conjunctively definable set $S \in {\rm CDEF}_A(T)$ can be represented as a union of some equivalence classes in $O\!B/E_A$. Let $A = \{a_1, a_2, ..., a_m\}$. Without loss of generality,  suppose $S$ can be described by a conjunctive formula $p_s \in {\rm CDL}_A$ that involves the first $k$ attributes in $A$, that is, $S = m(p_s)$ and
\begin{equation}
p_s = \langle a_1, v_1 \rangle \wedge \langle a_2, v_2 \rangle \wedge \cdots \wedge \langle a_k, v_k \rangle
\end{equation}
where $1 \le k \le m, v_i \in V_{a_i}$, and $a_i \ne a_j$ for $i \ne j (1 \le i, j \le k)$. $p_s$ can be easily transformed into a full disjunctive normal form as follows:
\begin{eqnarray}
p_s &=& \langle a_1, v_1 \rangle \wedge \langle a_2, v_2 \rangle \wedge \cdots \wedge \langle a_k, v_k \rangle \wedge T \nonumber\\
&=& \langle a_1, v_1 \rangle \wedge \langle a_2, v_2 \rangle \wedge \cdots \wedge \langle a_k, v_k \rangle  \wedge (\bigvee_{\substack{(v_{k+1}, \cdots, v_m)  \in \\ V_{a_{k+1}} \times \cdots \times V_{a_m}}} ( \langle a_{k+1}, v_{k+1} \rangle \wedge \cdots \wedge \langle a_m, v_m \rangle ) ) \nonumber\\
&=& \bigvee_{\substack{(v_{k+1}, \cdots, v_m)  \in \\ V_{a_{k+1}} \times \cdots \times V_{a_m}}} \langle a_1, v_1 \rangle \wedge \langle a_2, v_2 \rangle \wedge \cdots \wedge \langle a_m, v_m \rangle.
\end{eqnarray}
As a generalization of Equation (\ref{equa:conjunctive_intersection}), one can easily verify that the disjunction in a formula can be interpreted by set-union~\cite{yao2015two}, that is, $m(p \vee q) = m(p) \cup m(q)$. Thus, we have:
\begin{eqnarray}
S &=& m(\bigvee_{\substack{(v_{k+1}, \cdots, v_m)  \in \\ V_{a_{k+1}} \times \cdots \times V_{a_m}}} \langle a_1, v_1 \rangle \wedge \langle a_2, v_2 \rangle \wedge \cdots \wedge \langle a_m, v_m \rangle) \nonumber\\
&=& \bigcup_{\substack{(v_{k+1}, \cdots, v_m)  \in \\ V_{a_{k+1}} \times \cdots \times V_{a_m}}} m(\langle a_1, v_1 \rangle \wedge \langle a_2, v_2 \rangle \wedge \cdots \wedge \langle a_m, v_m \rangle),
\end{eqnarray}
where each meaning set $m(\langle a_1, v_1 \rangle \wedge \langle a_2, v_2 \rangle \wedge \cdots \wedge \langle a_m, v_m \rangle)$ is either an equivalence class in $O\!B/E_A$ or the empty set which can be simply removed from the union. As a result, a conjunctively definable set $S \in {\rm CDEF}_A(T)$ is represented as a union of some equivalence classes in $O\!B/E_A$. Consequently, a union of an arbitrary number of conjunctively definable sets from ${\rm CDEF}_A(T)$ can also be represented as a union of some equivalence classes in $O\!B/E_A$, which indicates that $B(O\!B/E_A) \supseteq B({\rm CDEF}_A(T))$.
\end{enumerate}
From the above two subset relationships, we can conclude that $B(O\!B/E_A) = B({\rm CDEF}_A(T))$.
\end{proof}

\section*{References}

\end{document}